\documentclass[twocolumn,switch,times]{article}
\usepackage{preprint}
\usepackage{hyperref}
\usepackage[authoryear,round]{natbib}

\hypersetup{
    colorlinks=true,
    linkcolor=black,
    urlcolor=cyan,
    citecolor=cyan,
    anchorcolor=black
}

\usepackage[utf8]{inputenc}	
\usepackage[T1]{fontenc}
\usepackage{xcolor}
\usepackage{lineno}					
\usepackage{tikz} 					
\usepackage{fancyhdr}
\usepackage{geometry}
\usepackage{lipsum}
\usepackage{scalerel}
\usepackage{sidecap}
\usepackage{times}
\usepackage{doi}
\usepackage{amssymb}
\usepackage{amsmath}
\usepackage{tabularx}
\usepackage{makecell} 
\usepackage{dsfont}
\usepackage{bbm}
\usepackage{bm}
\usepackage{array,booktabs}
\usepackage{upgreek}
\usepackage{algorithm}
\usepackage{algpseudocodex}
\usepackage[toc,title]{appendix}
\usepackage{authblk}
\usepackage{placeins}

\usepackage{newfloat}
\DeclareFloatingEnvironment[name={Supplementary Figure}]{suppfigure}
\usepackage{sidecap}
\sidecaptionvpos{figure}{c}
\usepackage[font=small]{caption}
\usepackage{etoolbox}

\usepackage{titlesec}
\titlespacing\section{0pt}{12pt plus 3pt minus 3pt}{1pt plus 1pt minus 1pt}
\titlespacing\subsection{0pt}{10pt plus 3pt minus 3pt}{1pt plus 1pt minus 1pt}
\titlespacing\subsubsection{0pt}{8pt plus 3pt minus 3pt}{1pt plus 1pt minus 1pt}

\definecolor{lime}{HTML}{A6CE39}

\newcolumntype{P}[1]{>{\centering\arraybackslash}p{#1}}

\newcommand{\appendixref}[1]{%
  \hyperref[#1]{Appendix~\ref*{#1}}\unskip
}

\title{Multispectral airborne laser scanning for tree species classification: a benchmark of machine learning and deep learning algorithms}

\usepackage{authblk}

\author[1,\small\textdagger,\small\ensuremath{\ast}]{Josef Taher}

\author[1,\small\textdagger]{Eric Hyyppä}

\author[1,\small\textdagger]{Matti Hyyppä}

\author[1,\small\textdagger]{Klaara Salolahti}

\author[1,\small\textdagger]{Xiaowei Yu}

\author[1,\small\textdagger]{Leena Matikainen}

\author[1]{Antero Kukko}

\author[1]{Matti Lehtomäki}

\author[1]{Harri Kaartinen}

\author[1]{Sopitta Thurachen}

\author[1]{Paula Litkey}

\author[2]{Ville Luoma}

\author[2]{Markus Holopainen}

\author[3]{Gefei Kong}

\author[3]{Hongchao Fan}

\author[4]{Petri Rönnholm}

\author[4]{Matti Vaaja}

\author[5]{Antti Polvivaara}

\author[5]{Samuli Junttila}

\author[5]{Mikko Vastaranta}

\author[6]{Stefano Puliti}

\author[6]{Rasmus Astrup}

\author[7]{Joel Kostensalo}

\author[8]{Mari Myllymäki}

\author[9,10]{Maksymilian Kulicki}

\author[9,11]{Krzysztof Stereńczak}

\author[12]{Raul de Paula Pires}

\author[12]{Ruben Valbuena}

\author[13]{Juan Pedro Carbonell-Rivera}

\author[13]{Jesús Torralba}

\author[14]{Yi-Chen Chen}

\author[14, 15]{Lukas Winiwarter}

\author[14]{Markus Hollaus}

\author[14]{Gottfried Mandlburger}

\author[16,14]{Narges Takhtkeshha}

\author[16]{Fabio Remondino}

\author[11]{Maciej Lisiewicz}

\author[11]{Bartłomiej Kraszewski}

\author[17]{Xinlian Liang}

\author[17]{Jianchang Chen}

\author[1]{Eero Ahokas}

\author[1]{Kirsi Karila}

\author[1]{Eugeniu Vezeteu}

\author[1]{Petri Manninen}

\author[1]{Roope Näsi}

\author[1]{Heikki Hyyti}

\author[1]{Siiri Pyykkönen}

\author[1]{Peilun Hu}

\author[1]{Juha Hyyppä}

\affil[1]{Department of Remote Sensing and Photogrammetry, Finnish Geospatial Research Institute FGI, Vuorimiehentie 5, Espoo, FI-02150, Finland}

\affil[2]{Department of Forest Sciences, University of Helsinki, Latokartanonkaari 7, Helsinki, FI-00790, Finland}

\affil[3]{Department of Civil and Environmental Engineering, Norwegian University of Science and Technology, Høgskoleringen 7A, Trondheim, NO-7491, Norway}

\affil[4]{Department of Built Environment, Aalto University, PO Box 14100, Espoo, FI-00076 AALTO, Finland}

\affil[5]{University of Eastern Finland, Faculty of Science, Forestry, and Technology, School of Forest Sciences, P.O. Box 111, Joensuu, FI-80101, Finland}

\affil[6]{Division of Forest and Forest Resources, National Forest Inventory, Norwegian Institute for Bioeconomy Research (NIBIO), Høgskoleveien 8, Ås, NO-1433, Norway}

\affil[7]{Natural Resources, Natural Resources Institute Finland, Yliopistokatu 6B, Joensuu, FI-80100, Finland}

\affil[8]{Bioeconomy and Environment, Natural Resources Institute Finland, Latokartanonkaari 9, Helsinki, FI-00790, Finland}

\affil[9]{IDEAS NCBR, ul. Chmielna 69, Warsaw, 00–801, Poland}

\affil[10]{Institute of Fundamental Technological Research, Polish Academy of Science, ul. Pawińskiego 5B, Warsaw, 02–106, Poland}

\affil[11]{Department of Geomatics, Forest Research Institute IBL, ul. Braci Leśnej 3, Sekocin Stary, 05–090, Poland}

\affil[12]{Dept. of Forest Resource Management. Swedish University of Agricultural Sciences, Umeå, Sweden}

\affil[13]{Geo-Environmental Cartography and Remote Sensing Group (CGAT), Universitat Politècnica de València, Camí de Vera s/n, Valencia, 46022, Spain}

\affil[14]{Department of Geodesy and Geoinformation, Technische Universität Wien, Wiedner Hauptstraße 8, Vienna, 1040, Austria}

\affil[15]{Faculty of Engineering Sciences, Universität Innsbruck, Technikerstraße 13, Innsbruck, 6020, Austria}

\affil[16]{3D Optical Metrology (3DOM) Unit, Bruno Kessler Foundation (FBK), Via Sommarive, 18, Trento, 38123, Italy}

\affil[17]{The State Key Laboratory of Information Engineering in Surveying, Mapping and Remote Sensing, Wuhan University, Wuhan, 430070, China}

\begin{document}

\twocolumn[\begin{@twocolumnfalse}

\maketitle

\begin{abstract}

Climate-smart and biodiversity-preserving forestry demands precise information on forest resources, extending to the individual tree level. Multispectral airborne laser scanning (ALS) has shown promise in automated point cloud processing, but challenges remain in leveraging deep learning techniques and identifying rare tree species in class-imbalanced datasets.
This study addresses these gaps by conducting a comprehensive benchmark of deep learning and traditional shallow machine learning methods for tree species classification. For the study, we collected high-density multispectral ALS data ($>1000$ $\mathrm{pts}/\mathrm{m}^2$) at three wavelengths using the FGI-developed HeliALS system, complemented by existing Optech Titan data (35 $\mathrm{pts}/\mathrm{m}^2$), to evaluate the species classification accuracy of various algorithms in a peri-urban study area located in southern Finland. We established a field reference dataset of 6326 segments across nine species using a newly developed browser-based crowdsourcing tool, which facilitated efficient data annotation. The ALS data, including a training dataset of 1065 segments, was shared with the scientific community to foster collaborative research and diverse algorithmic contributions. Based on 5261 test segments, our findings demonstrate that point-based deep learning methods, particularly a point transformer model, outperformed traditional machine learning and image-based deep learning approaches on high-density multispectral point clouds. For the high-density ALS dataset, a point transformer model provided the best performance reaching an overall (macro-average) accuracy of 87.9\%  (74.5\%) with a training set of 1065 segments and 92.0\% (85.1\%) with a larger training set of 5000 segments. With 1065 training segments, the best image-based deep learning method, DetailView, reached an overall (macro-average) accuracy of 84.3\% (63.9\%), whereas a shallow random forest (RF) classifier achieved an overall (macro-average) accuracy of 83.2\% (61.3\%). For the sparser ALS dataset, an RF model topped the list with an overall (macro-average) accuracy of 79.9\% (57.6\%), closely followed by the point transformer at 79.6\% (56.0\%).
Importantly, the overall classification accuracy of the point transformer model on the HeliALS data increased from 73.0\% with no spectral information to 84.7\% with single-channel reflectance, and to 87.9\% with spectral information of all the three channels. Furthermore, we studied the scaling of the classification accuracy as a function of point density and training set size using 5-fold cross-validation of our dataset. Based on our findings, multispectral information is especially beneficial for sparse point clouds with 1--50 $\mathrm{pts}/\mathrm{m}^2$. Furthermore, we observed that the classification error follows a power law $\varepsilon(m) \propto m^{-\alpha}$ as a function of the training set size $m$, and the classification error of the point transformer reduced significantly faster with increasing training set size compared to RF.

\end{abstract}


\vspace{0.5cm}

\vfill\hrule\vspace{2pt}
\enlargethispage{2\baselineskip}
\footnotesize Published in ISPRS Journal of Photogrammetry and Remote Sensing, 2026. Final version available at: \url{https://doi.org/10.1016/j.isprsjprs.2026.01.031} .
\copyright \ 2026. This manuscript version is made available under the CC-BY-NC-ND 4.0 license \url{https://creativecommons.org/licenses/by-nc-nd/4.0/}

\end{@twocolumnfalse}]

\renewcommand{\thefootnote}{\small\ensuremath{\ast}}
\footnotetext[1]{Corresponding author}
\renewcommand{\thefootnote}{}
\footnotetext[2]{\textit{Email addresses}: \newline josef.taher@nls.fi (Josef Taher), eric.hyyppa@nls.fi (Eric Hyyppä), matti.hyyppa@nls.fi (Matti Hyyppä), klaara.salolahti@nls.fi (Klaara Salolahti), xiaowei.yu@nls.fi (Xiaowei Yu), leena.matikainen@nls.fi (Leena Matikainen), juha.coelasr@gmail.com (Juha Hyyppä)}
\renewcommand{\thefootnote}{\small\textsuperscript{\textdagger}}
\footnotetext[3]{These authors contributed equally.}
\setcounter{footnote}{0}

\section{Introduction}
\label{sec: introduction}

Knowledge of the species-specific size distribution of forest stands is crucial for effective forest management planning and  optimization of the wood supply chain. This information allows for decisions on tailored silvicultural treatments, such as harvesting schedules, thinning operations, and regeneration strategies. Reliable information on tree species also contributes to sustainable forest management, ensuring a continuous supply of high-quality timber products while maintaining ecological integrity. Namely, the diversity of tree species plays an important role in shaping forest ecosystems, and it affects factors such as productivity, resilience to disturbances, recovery rates, internal competition among trees, ecosystem health, economic potential and biodiversity. For example in the boreal forest zone, aspen (\textit{Populus tremula} L.) is widely recognized as a keystone species, having an important role in supporting biodiversity \citep{kouki2004long, kivinen2020keystone}, since aspen stands often harbor a rich diversity of fungi, lichens, insects, birds, and mammals \citep{kuusinen1994epiphytic, angelstam1994woodpecker,tikkanen2006red, remm2017multilevel}. Additionally, accurate tree species identification in municipalities is required for achieving the targeted restoration and effective enhancement of biodiversity and canopy cover within urban ecosystems \citep{EU_nature, GER_nature, FIN_nature}.

Although remote sensing techniques, such as satellite imagery and airborne laser scanning (ALS), have shown great promise in forest monitoring \citep{hyyppa2000accuracy, naesset2004laser, holmgren2004identifying, orka2009classifying}, accurate species classification at the individual tree level remains a challenging task. For example, previous ALS studies in the boreal forest zone have demonstrated a high classification accuracy for the dominant pine and spruce trees \citep{holmgren2004identifying, orka2012simultaneously, yu2017single, hakula2023individual}, but the classification accuracy of deciduous trees has remained markedly lower especially for under-represented species \citep{orka2012simultaneously, axelsson2018exploring, yu2017single}.  In general, the classification accuracy degrades if the number of species is high \citep{heinzel2011exploring, prieur2021comparison} or if there are closely related species, particularly those with similar spectral and structural characteristics. Therefore, further efforts are needed to improve species classification using remote sensing techniques since accurate species classification will be required to enable multifunctional forestry serving multiple ecosystem services optimally at the same time.  

Today, airborne laser scanning is the most prevalent technique for forest inventory in the boreal forest zone, with a strong emphasis on economically efficient forest resource utilization. The state of the art in  species classification using ALS is summarized in several review papers \citep{chen2024tree, michalowska2021review, wang2018review, koenig2016full, fassnacht2016review} and a recent benchmarking study using the FOR-species20K open dataset by \citet{puliti2025benchmarking}. In general, the accuracy of species classification is affected by several factors, including the point cloud quality (e.g. point density, noise, geometric accuracy affected by beam footprint), additional features available for classification (intensity at one or multiple wavelengths, echo types, waveform information), the chosen classification method (machine learning or deep learning), and the difficulty of the studied forest area (number and similarity of species, tree density, size distribution of trees). Until recently, ALS-based species classification has been implemented using hand-crafted features and traditional shallow supervised machine learning (ML) classifiers, such as random forests (RF) \citep{orka2012simultaneously, dalponte2012tree,yu2017single}, support vector machines (SVM) \citep{orka2012simultaneously, dalponte2012tree, dalponte2014tree} or discriminant analysis \citep{holmgren2004identifying, orka2009classifying, heinzel2011exploring}.  Typical features for species classification have included point cloud metrics representing the geometry of a tree (e.g. crown shape, tree height, density and height distributions of points, trunk hits, echo ranks, point counts), backscatter and intensity statistics (e.g. intensity/amplitude, reflectance and waveform distributions), and temporal changes in the canopy (e.g. the difference of intensity or average canopy height between leaf-on and leaf-off seasons) \citep{holmgren2004identifying, orka2009classifying, fassnacht2016review, koenig2016full, lin2016comprehensive, yu2017single}. Throughout the manuscript, we  refer to traditional shallow machine learning classifiers as \textit{machine learning} approaches following the convention in several recent papers in tree species classification~\citep{xi2020see, mayra2021tree, hell2022classification}. In contrast, we use the term \textit{deep learning} method to refer to deep neural networks that learn their features.

During the past few years, deep learning (DL) approaches have been shown to outperform traditional shallow machine learning methods in species classification  \citep{xi2020see, liu2021tree}. As a result, several deep learning methods have been recently proposed for species classification both using ALS data \citep{ hell2022classification, marinelli2022approach, fan2023tree, lin2024pctrees} and terrestrial laser scanning (TLS) data \citep{seidel2021predicting, liu2022tree}. Deep learning approaches can be divided into image-based 2D methods \citep{seidel2021predicting, marinelli2022approach} and point-based 3D methods \citep{xi2020see,liu2021tree, fan2023tree,lin2024pctrees}. In image-based DL methods, the point cloud is projected into multiple views from different perspectives, which are given as an input for a 2D convolutional neural network (CNN), such as ResNet \citep{he2016deep}, DenseNet \citep{huang2017densely}, or YOLO \citep{Yolov82023}. On the other hand, the input of point-based methods is the point cloud itself with potential additional features. Typical neural net architectures of point-based methods include, e.g., PointNet \citep{qi2017pointnet}, PointNet++ \citep{qi2017pointnet++}, dynamic graph CNN (DGCNN) \citep{wang2019dynamic}, and point transformer \citep{zhao2021point}. Deep learning methods require large amounts of training data to reach their potential, which has resulted in a need for large open datasets enabling the development and benchmarking of new deep-learning-based classification techniques.
To this end, \citet{puliti2025benchmarking} recently presented FOR-species20K, a 33 species open dataset consisting of 20 000 trees collected using terrestrial, mobile and unmanned aerial vehicle based laser scanning with the goal to facilitate the development of sensor-agnostic classification approaches using only the point cloud geometry. Furthermore, \citet{puliti2025benchmarking} presented a comparison of seven sensor-agnostic deep learning methods concluding that the studied image-based deep learning models marginally outperformed the point-based methods.

Backscatter intensity (or reflectance), even though hard to standardize across different sensors, has been successfully applied for species classification from the beginning of ALS research. \citet{holmgren2004identifying} showed that a combination of geometric and intensity features led to an overall accuracy of 95\% for the discrimination of spruce and pine. Since then intensity and geometric features have been used in many studies on tree species classification. \citet{orka2009classifying} demonstrated a classification accuracy of 88\%  between Norway spruce and birch for about 350 trees, and \citet{korpela2010tree} achieved an accuracy of 88--90\% for Scots pine, Norway spruce, and birch across 13000 trees. Importantly, multiple studies have demonstrated that the classification accuracy obtained with geometric features can be significantly improved with the addition of intensity features \citep{axelsson2018exploring, yu2017single, lin2016comprehensive, suratno2009tree, shi2018tree, hakula2023individual, orka2009classifying}. For example, \citet{orka2009classifying}, \citet{yu2017single}, and \citet{hakula2023individual} reported 11, 9.9 and 13.5 percentage point improvements, respectively, to the overall classification accuracy when introducing single-channel intensity features to the classification. Motivated by the success of single-channel intensity features, the integration of ALS data with passive multispectral or even hyperspectral data has also been studied to further increase the classification accuracy \citep{holmgren2008species, orka2012simultaneously, dalponte2014tree, deng2016comparison, kaminska2018species, kaminska2021single, zhong2022identification, quan2023tree, lisiewicz2025comprehensive}.

Multispectral ALS using multiple laser wavelengths provides a further alternative to enhance the accuracy of species classification with potential to meet the increasing requirements for biodiversity reporting, climate change adaptation, and sustainable resource management, for example.
Originally, multispectral laser scanning was developed to increase the capacity of ALS point clouds for automated object recognition \citep{kaasalainen2007toward, hyyppa2013unconventional}. Multispectral data also provides additional information for tree species classification since different tree species have a different reflectance response as a function of wavelength \citep{hovi2017spectral}. During the past decade, multispectral ALS has been shown to improve species classification accuracy in multiple studies~\citep{yu2017single, axelsson2018exploring, amiri2019classification, prieur2021comparison, rana2022effect, hakula2023individual, wang2024individual}, with some studies observing a reduction of the classification error by up to $\sim30$\% compared to the use of single-channel features \citep{axelsson2018exploring, amiri2019classification, hakula2023individual}. Despite the promise of deep learning  for semantic segmentation on multispectral data \citep{reichler2024semantic, oinonen2024unsupervised, ruoppa2025unsupervised},  deep learning approaches for species classification have been studied to date in only a single paper focusing on multispectral ALS data with a point density of 30 pts/$\mathrm{m}^2$  \citep{wang2024individual}. Furthermore, the paper by \citet{hakula2023individual} is the only study utilizing high-density ($>1000$ pts/$\mathrm{m}^2$) multispectral ALS data for species classification. In \citet{hakula2023individual}, the use of multispectral ALS data reduced the overall classification error by 65\% compared to the use of single-channel geometric features and by 31\% compared to single-channel geometric and intensity features for the main tree species, including spruce, pine and birch. The use of deep learning methods for species classification on high-density multispectral ALS data has not been studied previously.

Consequently, there is a need for more in-depth research on multispectral airborne laser scanning for tree species classification, especially in the presence of a wide variety of species and with high-density point clouds. Thus, the primary goal of our study is to carry out an international benchmark of tree species classification methods on multispectral ALS data. To facilitate this advancement, we undertook the following actions:
\begin{itemize}
     \item We collected high-density ALS data (1300 pts/$\mathrm{m}^2$) using a novel multispectral scanner system in a study area located in southern Finland, where we have previously collected multispectral ALS data with a  point density of 35 pts/$\mathrm{m}^2$ using the Optech Titan system \citep{fernandez2016capability}. In the following, the two datasets are labeled as dense data and sparse data, respectively. We selected a peri-urban study area since it offers a sufficient variety of tree species in a single study area for our benchmarking study. 
    \item We established a high-quality field reference dataset comprising 6326 segments and nine species, including aspen, that is vital for biodiversity in the boreal forests. The field reference dataset was carefully acquired using a crowdsourcing tool facilitating the collection of large datasets required by deep learning models. 
    \item We shared a training dataset of the dense and sparse multispectral ALS data with the international scientific community to organize a benchmarking study on tree species classification using multispectral ALS data. Thirteen teams participated in the benchmarking study submitting in total  nine machine learning methods, 13 point-based deep learning methods (including their variants), and four image-based deep learning methods.
    \item  We further investigated the decisive factors affecting the classification accuracy, including the classification method, availability of spectral information (no intensity vs single-channel intensity vs multi-channel intensity), size of the training dataset, point density, segmentation quality, crown class, and similarity between the species. 
\end{itemize}

\section{Materials and Methods}
\label{sec: methods}

\subsection{Study area}
\label{sec: test site}

\begin{figure*}[t]
    \centering
    \includegraphics[width=0.85\textwidth]{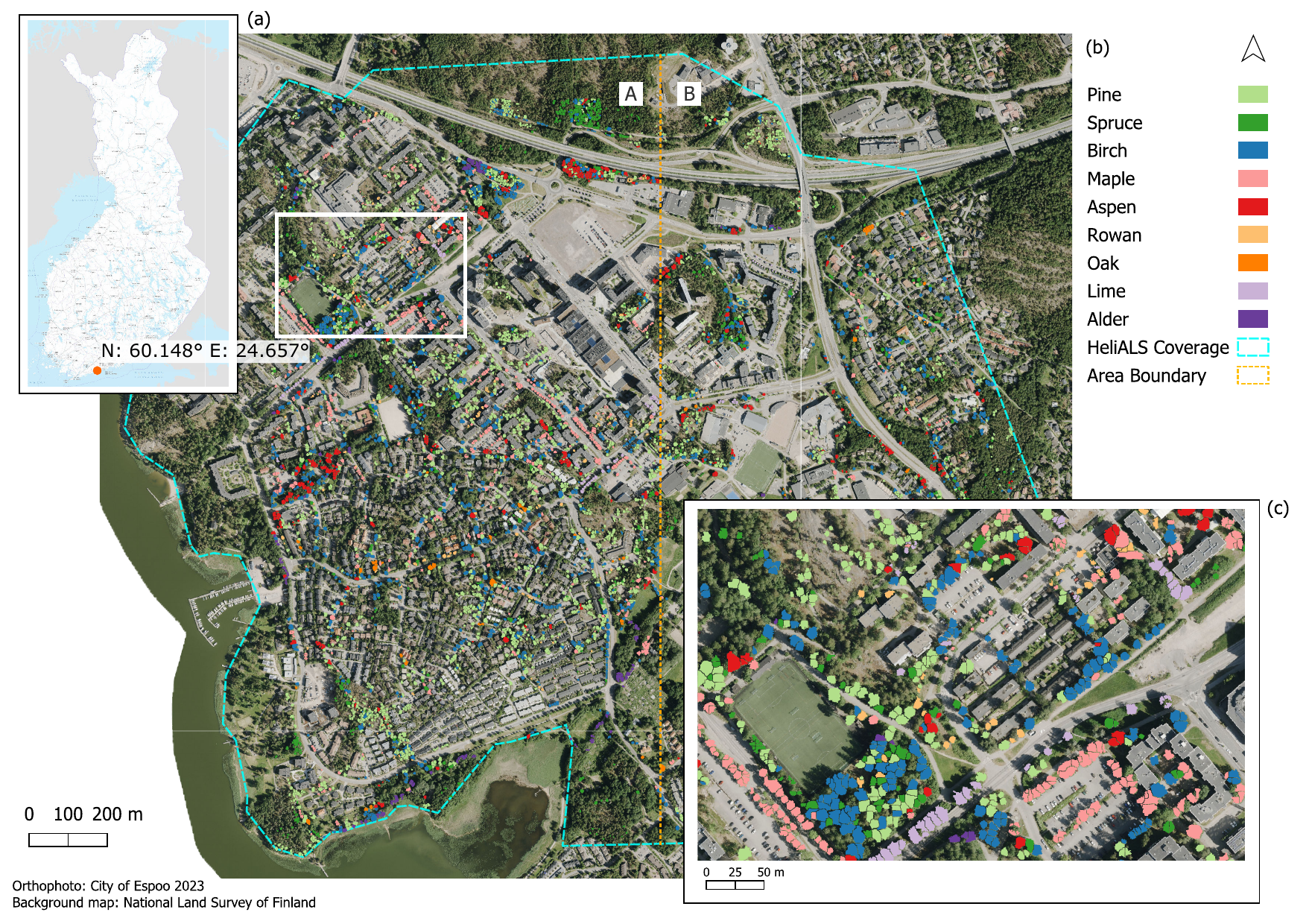}
    \caption{
    (a) The Espoonlahti study area is located in southern Finland. (b) Orthophoto of the Espoonlahti study area together with collected tree segments colored by the species in the reference dataset. The coverage of the HeliALS data is shown with the light blue line. Areas A and B are separated by the orange line. (c) A close-up of the orthophoto shown in (b). }
    \label{fig: test}
\end{figure*}

The study area (center point approximately 60.1462°N 24.6587°E) is located in Espoonlahti in the city of Espoo, directly 20 km west of the center of Helsinki, on the southern coast of Finland. The study area shown in Fig.~\ref{fig: test} consists of two sub-regions, A and B, both of which were used for the current study. The study area can be characterized as a peri-urban site since it is partially covered by forests and partially by suburban neighborhoods.  The area was selected due to the large diversity of tree species growing in the area compared to typical managed Finnish boreal forests with only a handful of species. On the study area, a total of 20--30 different tree species can be found, including natural and planted trees both in suburban environment and in forests, as the area comprises different residential areas, public buildings, a sports park and some recreational and non-recreational, unmanaged forest areas. The forests are natural forests and typically either pine-dominated rocky forests or mixed forests with coniferous and deciduous species. The mixed forests are relatively dense, while the rocky forests are sparser. Tree species classified in the study included pine (\textit{Pinus sylvestris}), spruce (\textit{Picea sp.}), birch (\textit{Betula sp.}), maple (\textit{Acer platanoides}), aspen (\textit{Populus tremula}), rowan (\textit{Sorbus sp.}), oak (\textit{Quercus robur}), linden (\textit{Tilia sp.}) and alder (\textit{Alnus sp.}), see more information in Section \ref{sec: dataset}. Pine, spruce, birch, aspen, rowan and alder are typically occurring natural tree species in the area, while most maples, oaks and lindens are planted trees in the environment.

\subsection{Measurement systems and acquisition of laser scanning data}
\label{sec: measurement systems}

\begin{figure}[t!]
    \centering
    \includegraphics[width=\linewidth]{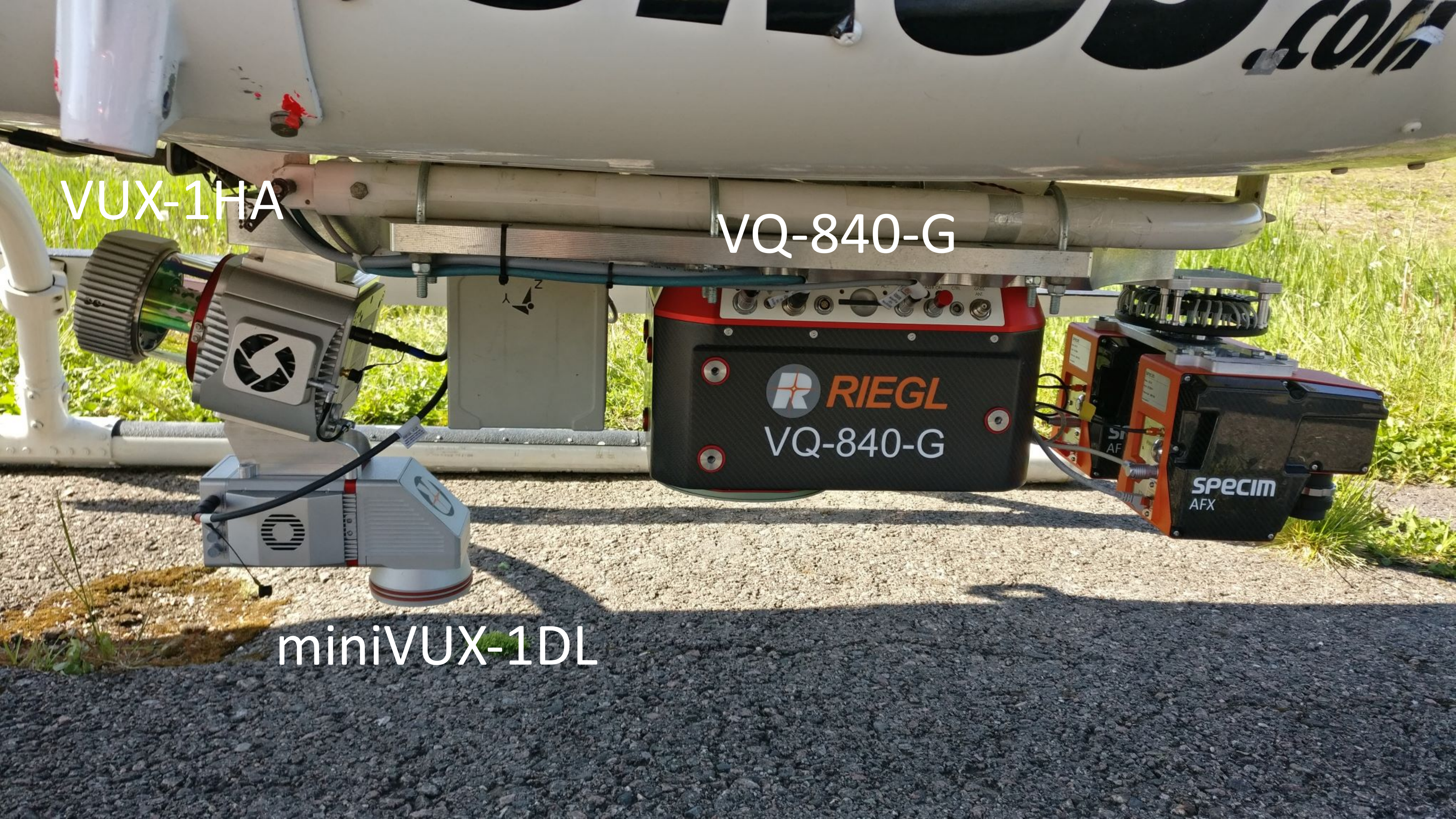}
    \caption{Photograph of the scanner arrangement of the HeliALS system, showing the VUX-1HA scanner at $\lambda=1550$ nm, miniVUX-1DL scanner at $\lambda=905$ nm, and VQ-840-G scanner at $\lambda=532$ nm.}
    \label{fig: HeliALS system}
\end{figure}

\begin{table}[ht!]
    \centering
    \caption{Specifications of the HeliALS measurement system.} 
    \label{tab: HeliALS specs}
   \resizebox{0.5\textwidth}{!}{
   \begin{tabular}{llll}
        \toprule
        Characteristics & Value \\
        \hline
         Altitude AGL  &  100 m \\
         Flight speed&  14 m/s\\
         \toprule
        Channel-specific characteristics & Channel 1 & Channel 2 & Channel 3 \\
         \hline
         Scanner name & VUX-1HA & miniVUX-1DL & VQ-840-G \\
         Laser wavelength & 1550 nm & 905 nm & 532 nm \\
         Beam divergence & 0.5 mrad & 0.5x1.6 mrad & 1 mrad \\
         
         Beam diameter at ground & 5 cm  & 5x16 cm &  10 cm\\
         Range accuracy & 5 mm &  15 mm &  20 mm\\
         Direction & 15 deg & Nadir &  Nadir\\
         Field of view & 180$^\circ$ & 46$^\circ$ (conical) & 40$^\circ$ (conical) \\
         Pulse repetition rate & 1017kHz & 100kHz & 200kHz \\
         Resulting point density & 581~pts/$\mathrm{m}^2$ & 175~pts/$\mathrm{m}^2$ & 519~pts/$\mathrm{m}^2$ \\
         \bottomrule
    \end{tabular}}
\end{table}

\begin{table}[ht!]
    \centering
    \caption{Specifications of the Optech Titan measurement system.} 
    \label{tab: Optech Titan specs}
    \resizebox{0.5\textwidth}{!}{\begin{tabular}{llll}
        \toprule
        Characteristics & Value \\
        \hline
         Altitude AGL  &  700 m\\
         Flight speed  & $\sim$70 m/s \\
         Strip width  &  510 m \\
         Lateral overlap & 30\%  \\
         \toprule
        Channel-specific Characteristics & Channel 1 & Channel 2 & Channel 3 \\
         \hline
         Laser wavelength & 1550 nm & 1064 nm & 532 nm \\
         Beam divergence & 0.35 mrad & 0.35 mrad & 0.7 mrad \\
         Beam diameter at ground & 25 cm & 25 cm & 50 cm \\
         Direction & 3.5° forward & nadir & 7° forward \\
         Field of view & $40^\circ$ & $40^\circ$& $40^\circ$  \\
         Pulse repetition rate &  300 kHz & 300 kHz & 300 kHz  \\
         Resulting point density$^*$ & 11~pts/$\mathrm{m}^2$ &  13~pts/$\mathrm{m}^2$ &  11~pts/$\mathrm{m}^2$\\
         \bottomrule
         \multicolumn{4}{l}{\footnotesize{$^*$ After cutting points of overlapping flight lines.}}
    \end{tabular}}
\end{table}

Measurement systems used for ALS data acquisition included HeliALS and Optech Titan. The HeliALS system has been developed at the Finnish Geospatial Research Institute (FGI), while Optech Titan was the first operational multispectral ALS system and launched by Teledyne Optech (Ontario, Canada) in 2014. 

The HeliALS dataset was acquired on July 20 and 28, 2023, from an altitude of 100 m with leaf-on conditions.  A helicopter, flying at 14 m/s, carried a system of three Riegl drone laser scanners operating   in infrared, near-infrared and green wavelengths, namely a VUX-1HA ($\lambda=1550$ nm), a miniVUX-1DL ($\lambda=905$ nm), and a VQ-840-G ($\lambda=532$ nm). Positioning was achieved using a NovAtel (LITEF) ISA-100C inertial measurement unit (IMU), a NovAtel PwrPak7 Global Navigation Satellite System (GNSS) receiver, and a NovAtel (Vexxis) GNSS-850 antenna.  The initial performance of the HeliALS system was presented in \citet{hakula2023individual}, which was the first study focusing on tree species classification from high-density multispectral ALS data.  After the publication, the green channel data have been optimized by increasing the receiver aperture of the VQ-840-G scanner from 3 mrad to 6 mrad.
The data were processed to point clouds at the FGI. Further specifications of the HeliALS system and data acquisition are provided in Table~\ref{tab: HeliALS specs}.

The Optech Titan system operates in three channels with infrared ($\lambda=1550$ nm), near-infrared ($\lambda=1064$ nm) and green wavelengths ($\lambda=532$ nm), respectively. The data were acquired on June 14, 2016, when trees were in full leaf \citep{karila2019effect}. The acquisition was carried out in cooperation with TerraTec Oy (Helsinki, Finland). A fixed-wing aircraft and a flying altitude of 700 m were used. Data from the three channels were obtained as three separate point clouds. After basic processing by TerraTec Oy, the data were delivered to the FGI in the Finnish ETRS-TM35FIN coordinate system with height values in the Finnish N2000 height system. Further specifications of the  Optech Titan system and data acquisition  are presented in Table~\ref{tab: Optech Titan specs}.

\subsection{Pre-processing of laser scanning data}
\label{sec: pre-processing}

The HeliALS triple-wavelength  data were processed into point clouds using RiProcess software (version 1.9.0, Riegl GmbH, Austria). GNSS-IMU trajectories for the two flights were computed in Waypoint Inertial Explorer (8.90, NovAtel Inc., Canada) using a virtual base station from the Trimnet service (Geotrim Oy, Finland) and tightly coupled kinematic post-processing. Boresight calibrations between the IMU and each scanner were solved using RiProcess functionalities that use planar surfaces for solving the rotation angle estimates.

The data from each scanner were loaded to TerraScan (Terrasolid Ltd., Finland), and for each point the closest points within 20 cm were sought from the  two other channels to form the reflectance triplets that were then stored in the RGB color fields in the LAZ file structure. Thus, points that do not have close-by counterparts on either one or two channels may exhibit only NA (coded as value `1') on the respective RGB fields. The reflectance values in the LAZ were stored as decibel values and exported with RIEGL extrabytes (echo deviation, reflectance, and pulse width). The definition of the reflectance, amplitude and echo deviation attributes for the Riegl scanners are described in \citet{pfennigbauer2010improving}. Eventually, the full point cloud dataset was split into 200 m square tiles and delivered for the benchmark in 1.2 LAZ format with ETRS-TM35FIN coordinates and ellipsoidal elevation values. 

Preprocessing steps for the Optech Titan data included some basic processing steps carried out by TerraTec Oy, intensity range ($R$) correction according to $R^2$ \citep{matikainen2017object}, cutting of overlap points and removal of an additional cross line (using TerraScan). Multispectral information for this dataset is available in the form of three range-corrected intensity channels. Reflectance information is not available. The point densities measured after preprocessing were 11 points/$\mathrm{m}^2$, 13 points/$\mathrm{m}^2$ and 11 points/$\mathrm{m}^2$ for the Channels 1, 2 and 3, respectively. The preprocessed set of data was provided as ‘original’ data for the benchmarking.

The original point clouds from the Optech Titan data were also further classified into ground, vegetation, building and noise by using LAStools software (rapidlasso GmbH, Germany). Ground points were used to create a digital terrain model and for normalizing the point clouds by subtracting the ground elevation from the height of the laser points. 

Tree segmentation was carried out based on the Optech Titan data. The resulting segments were also applied for tree species classification with the HeliALS data to eliminate the variability of segment boundaries from the comparison. Given the fact that plenty of deciduous trees are growing in the study area, we used a varying window watershed-based method for the tree detection first depicted in \citep{kaartinen2012international}. First, we created a canopy height model (CHM) from the first returns of the vegetation points of Channels 1 and 2 with 0.5 m pixel size. The CHM was then smoothed by a Gaussian filter with a varying window size, followed by a local maximum filtering to identify the tops of trees. The window size varied according to the CHM height of the target as follows: 3 pixels for 0--7 m height, 5 pixels for 7--20 m height, 7 pixels for 20--30 m height, and 9 pixels for greater than 30 m height. Finally, the marker-controlled watershed algorithm was applied to delineate the crowns of trees using detected tree tops as seeds \citep{kaartinen2012international}. Output of the tree segmentation is a raster image with a unique label for each segment, i.e. the pixels with the same label form one segment. These images together with the original and normalized point clouds were delivered to the partners for species classification.

\subsection{Benchmark dataset}
\label{sec: dataset}

\begin{figure*}[h]
    \centering
    \includegraphics[width=0.85\textwidth]{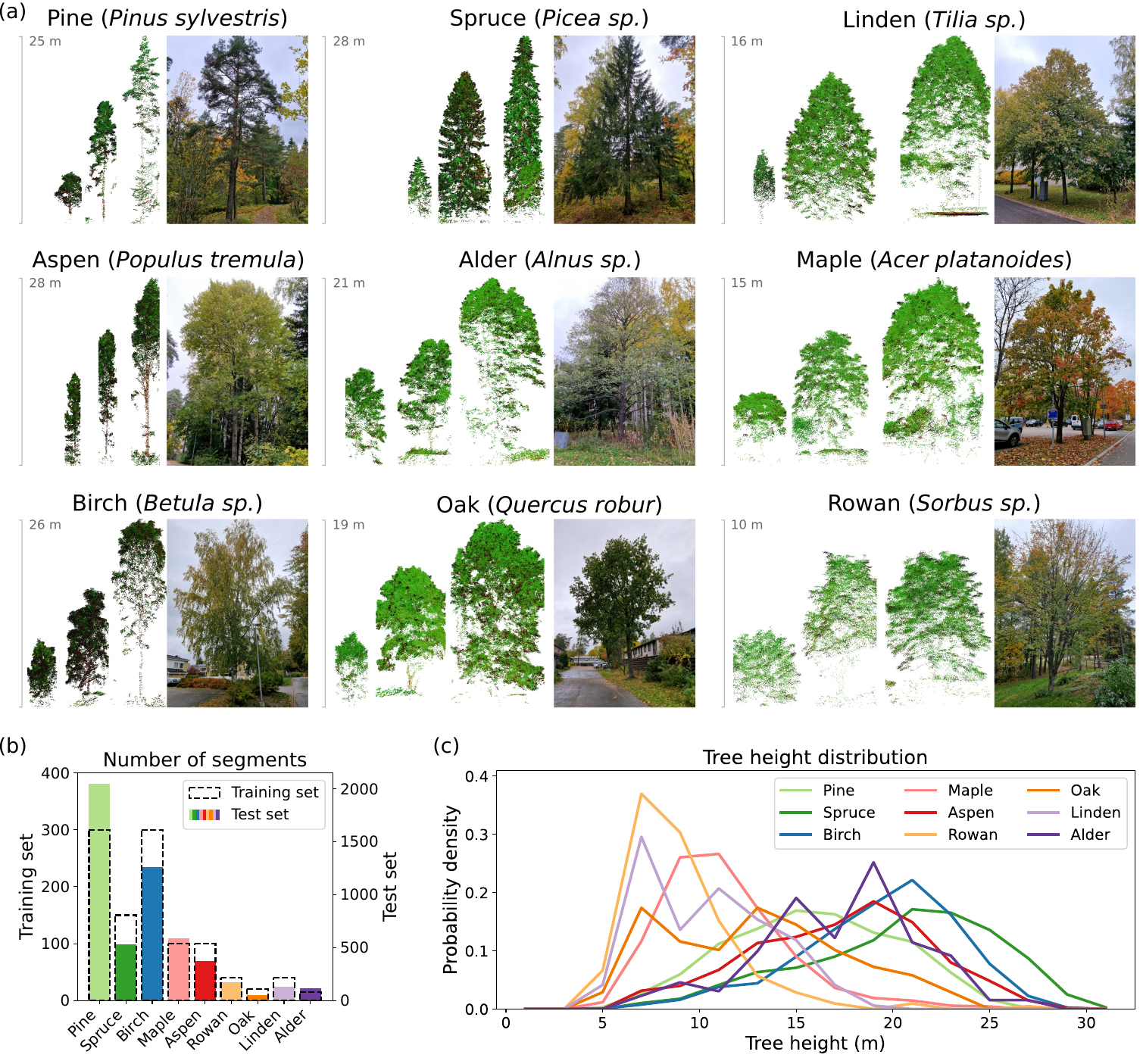}
    \caption{(a) Example multispectral point clouds and photographs of the nine tree species found in the dataset. The red, green and blue color channels of the point clouds represent the laser pulse return intensities at the wavelengths $\lambda_1 = \text{532 nm}, \lambda_2 = \text{905 nm}$ and $\lambda_3 = \text{1550 nm}$, respectively. The height scale varies between the visualized point clouds. Photographs of the representative species instances were taken during the autumn of 2024. (b) Number of segments per species in the training and test sets. The dashed lines represent the training set (left y-axis) and the solid bars correspond to the test set (right y-axis). (c) Species-wise tree height distributions for the entire dataset.}
    \label{fig: species visualization and distribution}
\end{figure*}

The contestants of the benchmark competition received a training set containing segmented point clouds, the segment labels and the ground-truth species classes. The training set consisted of 1065 segments. The participating methods were benchmarked with a test set containing 5261 segments. The dataset was split into training and test sets using random sampling for a given number of segments per species. The full dataset was not available at the time of constructing the training set because the collection of the field reference data was still ongoing. Therefore, no detailed stratification approaches were used for splitting the dataset into training and test sets. Figure \ref{fig: species visualization and distribution} visualizes the nine tree species included in the dataset, the number of segments per species in the training and test sets, and also the species-wise tree height distributions for the entire dataset. See also the distribution of mean intensity for the studied tree species at the three wavelengths of the HeliALS and Optech Titan systems in \ref{ap: distribution of mean intensity for different species}. The distributions showcase the advantage of multispectral data in providing spectral information at multiple wavelengths.

\subsection{Reference data collection}
\label{sec: reference data}

\begin{figure}[ht]
    \centering
    \includegraphics[width=0.48\textwidth]{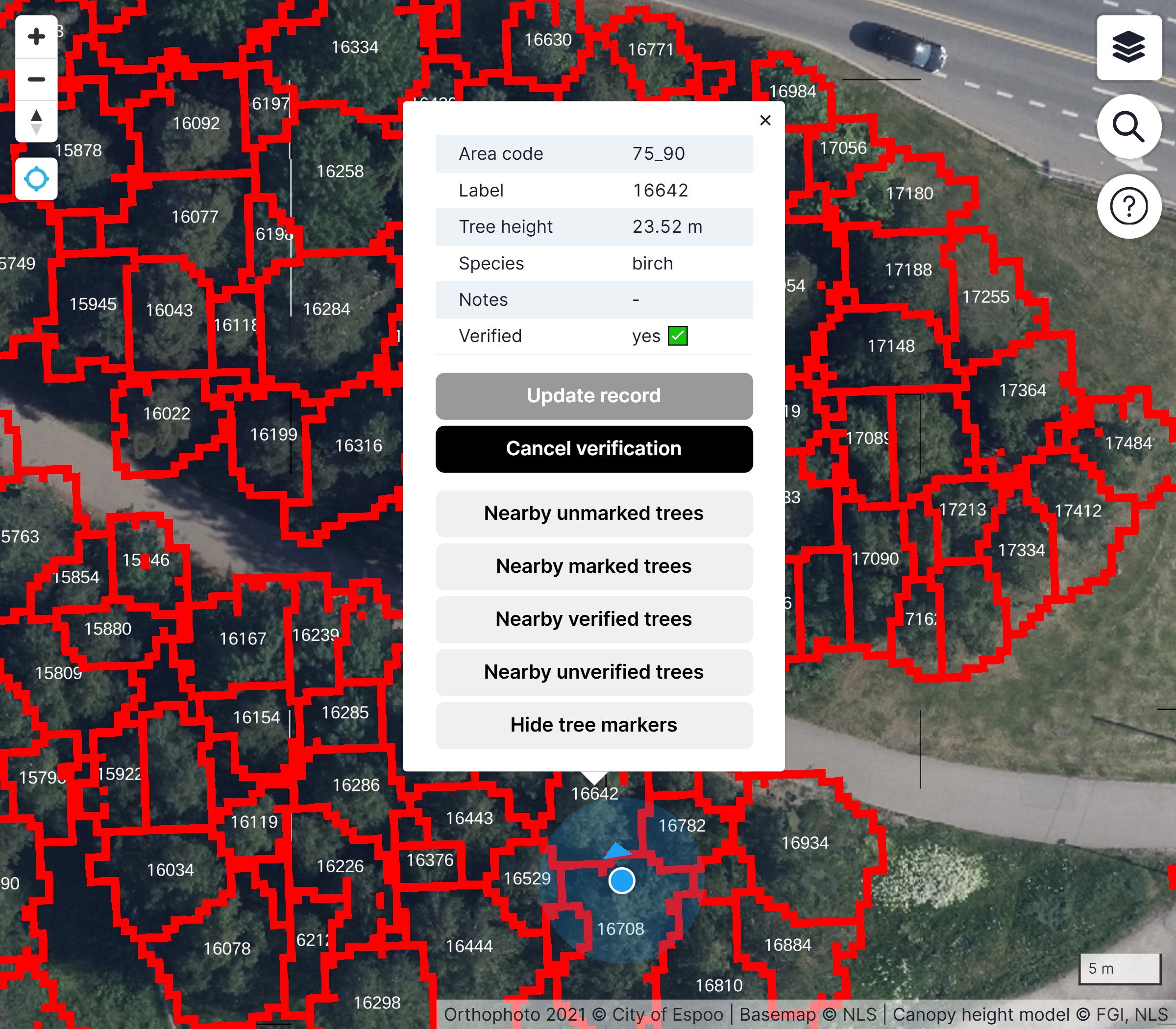}
    \caption{Screenshot of the crowdsourcing application used to collect ground-truth species data. In this example, the application shows a true orthophoto map, segment boundaries (red lines), the surveyor's location (blue circle) based on GNSS positioning, and the information of the nearest segment together with possible actions for the surveyor. The true orthophoto map in the background was obtained from the \cite{espoo_open_data}.}
    \label{fig: crowdsourcing tool}
\end{figure}

A browser-based crowdsourcing tool was developed to help collecting the ground-truth species data in the field. The crowdsourcing tool is a web application that visualizes the study area on a 2D background map as shown in Fig.~\ref{fig: crowdsourcing tool}. The web application shows the pre-processed segment boundaries on the map and surveyors can annotate segments with the correct species and also add notes if needed. The crowdsourcing application enables surveyors to locate themselves on the study area with GNSS. To aid positioning when GNSS accuracy is low, a true orthophoto map from the \cite{espoo_open_data} was visualized as the background map. The 5-cm-resolution orthophoto map had been photographed in the summer of 2021. To help identifying the correct trees on the study area, the canopy height model of the study area was also added to the map as an additional map layer.

Reference data was collected in the field using the crowdsourcing application by the employees of the FGI in 2023 and 2024. The process was guided by several key principles.
First, participants were introduced to the data collection tools before starting the field work to establish a baseline for consistency and reduce initial misclassification errors. Second, we made improvements to the user interface of the crowdsourcing application during the field campaign based on our learnings in the field to further ease the collection of the ground-truth data.
Third, we implemented an expert verification process to ensure data quality and reduce inherent subjectivity: The initial species classifications made by the crowd were subsequently revisited on the field and verified by experts using the crowdsourcing tool. To help determine which segments needed more attention in the verification phase, we computed for each segment the proportion of classification methods that predicted an incorrect species and prepared an associated map. In the case of minority species, segments with multiple incorrect predictions were typically correctly labeled in the field. For the more common species, some segments with multiple incorrect predictions had been incorrectly labeled on the first round of field measurements, and they were subsequently corrected during the on-site verification. This two-step approach---using the crowd for broad coverage and experts for final validation---helped to reduce subjectivity. 

In addition to field data collection using the crowdsourcing application, some  planted deciduous roadside trees were annotated using the open tree database of \citet{espoo_open_data}, and some clear instances of spruces were labeled using the 5-cm-resolution true orthophoto due to the unique shape of the spruces.

\subsection{Categories of tree segments}
\label{sec: segment categories}

\begin{figure*}[h]
    \centering
    \includegraphics[width=0.83\textwidth]{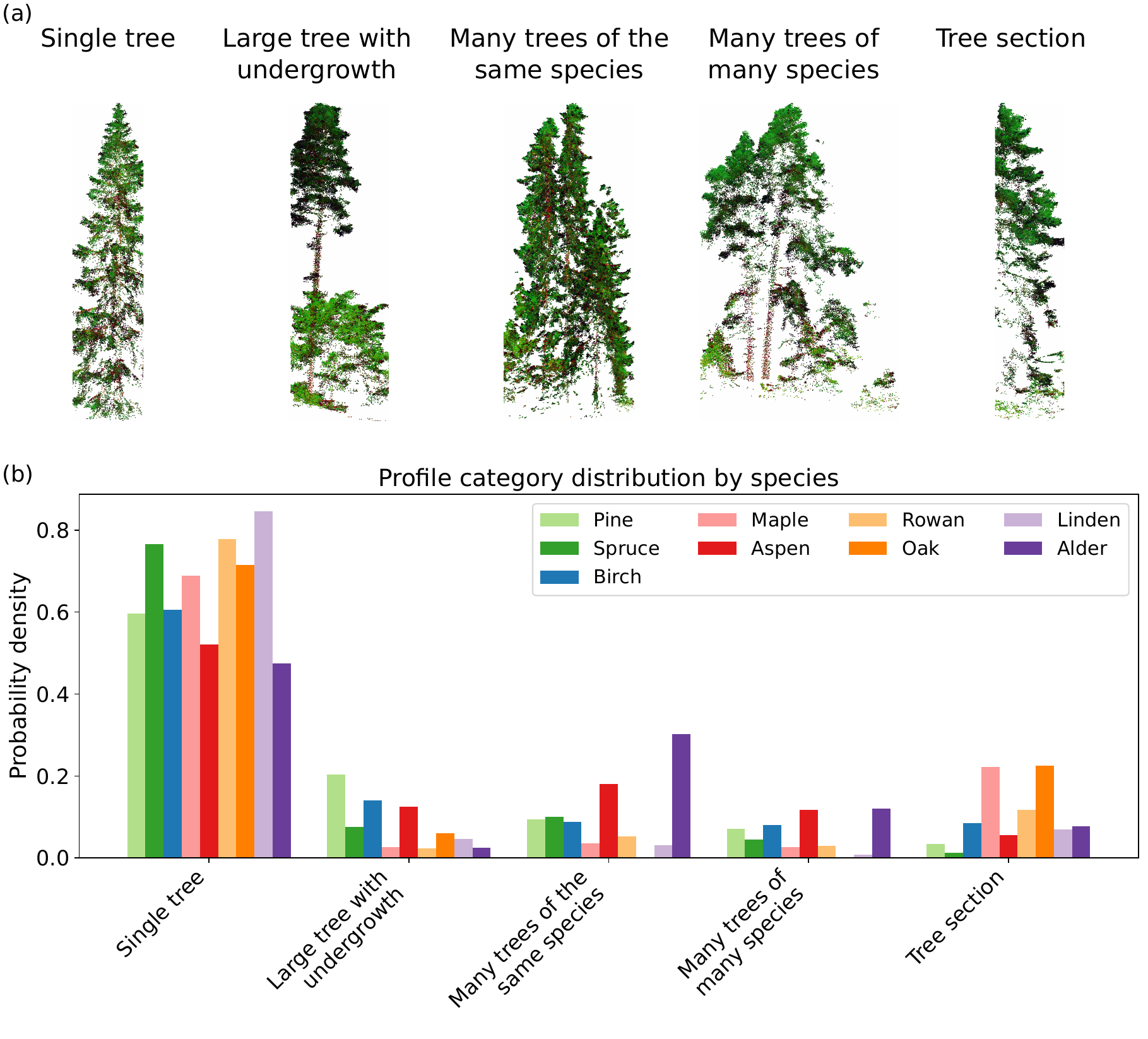}
    \caption{(a) Point cloud projections of segment instances belonging to the different profile categories. (b) Profile category distribution by species for the test set.}
    \label{fig: profile category distributions}
\end{figure*}

After labeling the segments and verifying the labels, the segments were categorized by manually inspecting profiles generated from the HeliALS segments. The applied categories were `Single tree', `Large tree with undergrowth', `Many trees of the same species', `Many trees of many species', and `Tree section'. In the case of `Single trees', there was only one tree visible in the profile image of the segment. A segment belongs to the `Large tree with undergrowth' category if there was one taller tree and possibly some significantly smaller trees or bushes at most one third of the height of the larger tree in the segment. `Many trees of the same species' category corresponds to segments which contained multiple trees, each having the same species. If there were multiple trees with different species in the same segment, the segment was categorized into the `Many trees of many species' category and the ground-truth species was recorded according to the species of the tallest tree within the segment. Segments with many trees where a single prominent tree species could not be identified were not included in the analysis. `Tree section' corresponds to segments that contained only a part of a tree.

Figure \ref{fig: profile category distributions} presents the profile category distribution by species for the test set. For each species class, the majority of the segments belong to the `Single tree' category. However, it is worth mentioning that in the study area, alders typically grow in places with lots of other alders and, therefore, 30\% of segments labeled as alder belong to the `Many trees of the same species' category. Furthermore, 22\% of maples and oaks belong to the `Tree section' category because of over-segmentation of their wide canopies.

The segments were also categorized to different crown classes using alternative criteria based on tree heights and distances to neighboring trees as inspired by \citet{wang2019field}. Segments were considered `Isolated' if within $8~\mathrm{m}$ radius, there were only segments with a height of at most half of the height of the `Isolated' segment. Segments were considered `Dominant' if the segment was not `Isolated' and it was the tallest segment within $8~\mathrm{m}$ radius. If there was at least one segment at least $5~\mathrm{m}$ taller than a given segment within $6~\mathrm{m}$ radius, the segment was assigned to the `Smaller tree next to larger tree' category. `Roadside' segments were detected using the open tree database of the \cite{espoo_open_data}. All other segments were assigned to the `Co-dominant' category.

\section{Benchmark contest and species classification algorithms}
\label{sec: algorithms}

\subsection{Overview of the benchmark contest}

The benchmark dataset was released at the Second ISPRS GEOBENCH workshop, ``Evaluation and Benchmarking of Sensors, Systems and Geospatial Data in Photogrammetry and Remote Sensing,'' held in Krakow, Poland, on October 23--24, 2023. The initial submission deadline at the end of January 2024 was extended to the summer of 2024. This extension was necessary to allow for thorough checking of the reference data and to ensure accurate and consistent contributions.

Participating partners were asked to submit the following: 1) a description of their applied methods, including relevant citations if previously published; 2) an Excel spreadsheet containing the tree segment number, classified species number, and sub-region code (A or B) for each segment (study area was divided into two sub-regions); and 3) if modified segments (i.e., divided or merged) were used, a raster image of the final segment numbers, matching the original segment number raster in cell size and location. 

FGI also developed classification methods that were submitted for benchmarking. Therefore, the analysis of the classification results was performed by separate people at the FGI.

\subsection{Participating methods}
\label{sec: participating methods}

Submissions to the benchmarking competition included point-based (3D) deep learning methods, profile-based (2D) deep learning methods and shallow machine learning methods. The following sections describe the participating methods in detail. We adopt the following naming convention for the participating methods: \textlangle organization name\textrangle-\textlangle method details\textrangle-\textlangle classifier category\textrangle-\textlangle2D or 3D-based method\textrangle. As an exception, the variants of the DetailView method do not fit this naming convention as discussed in Sec.~\ref{sec: profile-based deep learning methods}.

\subsubsection{Point-based (3D) deep learning methods}
\label{sec: point-based deep learning methods}

\begin{table*}[ht!]
\centering
\caption{\label{tab: methods DL 3D} Summary of point-based (3D) deep learning methods for species classification. For each method, the table briefly describes point cloud pre-processing steps, input data format of the deep learning model, architecture of the model, training strategies, and inference strategies.   }
\resizebox{\textwidth}{!}{
\begin{tabularx}{1.45\textwidth} {
>{\hsize=.7\hsize\linewidth=\hsize}X
>{\hsize=1.1\hsize\linewidth=\hsize}X
>{\hsize=1.2\hsize\linewidth=\hsize}X
>{\hsize=1.3\hsize\linewidth=\hsize}X
>{\hsize=.7\hsize\linewidth=\hsize}X
  }
 \specialrule{1.0pt}{0em}{0em}
  \textbf{Method and data} & \textbf{Pre-processing} & \textbf{Input data} & \textbf{Architecture and Training} & \textbf{Inference} \\
 \specialrule{1.0pt}{0em}{0em}
FGI-PointTransformer-DL-3D, \mbox{HeliALS}  & Removal of noise and ground points. Voxelization with a side length of 5 cm to obtain a more balanced point distribution. Point clouds of the three channels merged and each point assigned radiometric information from the two other channels based on nearest neighbor interpolation. Imputation strategies used if point attributes are missing for some of the channels.  & (3+13)D point cloud with 8192 points per segment. Besides coordinates, attributes include intensities, amplitudes, reflectances, and echo deviations for the 3 scanners, and an echo return number. Coordinates normalized to unit sphere, other attributes normalized by scaling (except for the echo return number).   & Five randomly initialized Point Transformer models \citep{zhao2021point} trained from scratch with 95-5 train-validation split. Both ordinary cross-entropy loss (FGI-PointTransformer-DL-3D) and weighted cross-entropy loss (FGI-PointTransformerWeighted-DL-3D) were tested. Augmentations applied at random (small random translation, scaling, and jittering of coordinates, and small random scaling and jittering of other attributes).  &  Majority voting based on the five trained Point Transformer models. \\
\specialrule{1.0pt}{0em}{0em}
FGI-PointTransformer-VUX-DL-3D, \mbox{HeliALS}
 & Same as FGI-PointTransformer-DL-3D but using radiometric and echo information only from the VUX scanner while ignoring radiometric information of the two other scanners. & (3+6)D point cloud with 8192 points per segment. Besides point coordinates, attributes include the intensity, echo deviation, amplitude, reflectance, number of returns, and return number of the VUX scanner. Normalization as for FGI-PointTransformer-DL-3D. & Same as FGI-PointTransformer-DL-3D. Both ordinary and weighted cross-entropy loss were tested.& Same as FGI-PointTransformer-DL-3D. \\
\specialrule{1.0pt}{0em}{0em}
FGI-PointTransformer-VUXReflectance-DL-3D,  \mbox{HeliALS}
 & Same as FGI-PointTransformer-DL-3D but using only the reflectance from the VUX scanner and ignoring any other radiometric or echo information. & (3+1)D point cloud with 8192 points per segment. Besides coordinates, attributes include the reflectance of the VUX scanner. Normalization as for FGI-PointTransformer-DL-3D. & Same as FGI-PointTransformer-DL-3D. Both ordinary and weighted cross-entropy loss were tested.& Same as FGI-PointTransformer-DL-3D. \\
 \specialrule{1.0pt}{0em}{0em} 
FGI-PointTransformer-GeometryOnly-DL-3D, \mbox{HeliALS}
 & Same as FGI-PointTransformer-DL-3D but using only the geometry of the combined point cloud. & 3D point cloud with 8192 points per segment. Coordinates normalized to unit sphere. & Same as FGI-PointTransformer-DL-3D with ordinary cross-entropy loss. & Same as FGI-PointTransformer-DL-3D. \\
 \specialrule{1.0pt}{0em}{0em} 
FGI-DGCNN-DL-3D, \mbox{HeliALS} & Same as FGI-PointTransformer-DL-3D.
& Same as FGI-PointTransformer-DL-3D. & Five randomly initialized dynamic graph convolutional neural net (DGCNN) models \citep{wang2019dynamic} trained from scratch with 90-10 train-validation split and cross-entropy loss. Same augmentations as FGI-PointTransformer-DL-3D. &  Majority voting based on the five trained DGCNN models. \\
\specialrule{1.0pt}{0em}{0em}
FGI-PointNet-DL-3D, \mbox{HeliALS}& 
Same as FGI-PointTransformer-DL-3D. & Same as FGI-PointTransformer-DL-3D. & Five randomly initialized PointNet models \citep{qi2017pointnet} trained from scratch with 90-10 train-validation split and cross-entropy loss. 
Same augmentations as FGI-PointTransformer-DL-3D. &  Majority voting based on the five trained PointNet models. \\
 \specialrule{1.0pt}{0em}{0em}
FGI-Point2Vec-DL-3D, \mbox{HeliALS} & Same as FGI-PointTransformer-DL-3D, \mbox{HeliALS}. & (3+13)D point cloud with 4096 points per segment. Besides coordinates, attributes include intensities, amplitudes, reflectances, and echo deviations for the 3 scanners, and an echo return number. Coordinates normalized to unit sphere, other attributes z-score normalized.  &  Point2Vec model \citep{zeid2023point2vec} in fully-supervised classification setup. Trained from scratch using 80-20 train-validation split. Geometric augmentations applied at random (scaling, translation and rotation in $30^\circ$ degree steps around the z-axis). Weighted cross-entropy loss with label smoothing. & Majority voting based on five inference runs with a single model. \\
 \specialrule{1.0pt}{0em}{0em}
TUW-PointNet++-DL-3D, \mbox{HeliALS}  & Point clouds of the three channels combined by introducing three flag variables indicating the source channel of each point. & (3+4)D point cloud with 8192 points per segment. Besides coordinates, attributes include the three flag variables (one of which is set to one), and the intensity value. Coordinates centered, but not scaled to preserve height. & PointNet++ model \citep{qi2017pointnet++} with two PointNet layers and a final multilayer perceptron layer trained from scratch with 67-33 train-validation split. Geometric augmentations by generating 6 rotations around the z-axis for each segment. Weighted loss function to mitigate class imbalance. The model extends Ensemble PointNet++ \citep{TUWien_github2024,puliti2025benchmarking} to multispectral data. 
& Prediction of the trained model. \\
 \specialrule{1.0pt}{0em}{0em}
FBK-PointNet++-DL-3D, \mbox{HeliALS} and  \mbox{Optech Titan} & Removal of noise and ground points. Point clouds of the three channels were merged and each point was assigned radiometric information from the three channels based on nearest neighbor interpolation. & (3+3)D point cloud with 500 points per segment. Besides coordinates, attributes include the reflectance for each channel. Coordinates centered, but not scaled. Reflectances normalized. & PointNet++ model \citep{qi2017pointnet++} trained from scratch with 80-20 train-validation split. Geometric and spectral augmentations for alder and oak to triple the number of examples. Geometric augmentations for rowan and linden to double the number of examples. & Prediction of the trained model. \\
\specialrule{1.0pt}{0em}{0em}
FGI-PointTransformer-DL-3D, \mbox{Optech Titan}  & Removal of noise and ground points. Point clouds of the three channels merged and each point assigned intensity information from the two other channels based on nearest neighbors. & (3+4)D point cloud with 2048 points per segment. Besides coordinates, attributes include intensities for the 3 scanners, and an echo return number. Normalization as for
the HeliALS dataset.   & 
Same as FGI-PointTransformer-DL-3D for the HeliALS dataset using ordinary cross-entropy loss.
& 
Same as FGI-PointTransformer-DL-3D for the HeliALS dataset.\\
 \specialrule{1.0pt}{0em}{0em}
FGI-PointNet-DL-3D, \mbox{Optech Titan} & Same as FGI-PointTransformer-DL-3D, \mbox{Optech Titan}. & Same as FGI-PointTransformer-DL-3D, \mbox{Optech Titan}. & Same as FGI-PointNet-DL-3D, \mbox{HeliALS}. & Majority voting based on the five trained PointNet models. \\

 \specialrule{1.0pt}{0em}{0em}
FGI-DGCNN-DL-3D, \mbox{Optech Titan} & Same as FGI-PointTransformer-DL-3D, \mbox{Optech Titan}. & Same as FGI-PointTransformer-DL-3D, \mbox{Optech Titan}. &  Same as FGI-DGCNN-DL-3D, \mbox{HeliALS}. & Majority voting based on the five trained DGCNN models. \\

\specialrule{1.0pt}{0em}{0em}
\end{tabularx}}
\end{table*}

We summarize the studied point-based deep learning methods in Table~\ref{tab: methods DL 3D}, where we present relevant details for each of the methods, including information on point cloud pre-processing, format of the input data, model architecture, training, and inference. We refer to the point-based methods also as 3D deep learning methods.

In this work, our comparison includes the following 3D deep learning methods: FGI-PointTransformer-DL-3D and its variants, FGI-DGCNN-DL-3D, FGI-PointNet-DL-3D, FGI-Point2Vec-DL-3D, TUW-PointNet++-DL-3D, and FBK-PointNet++-DL-3D. The studied methods utilize various point-based deep learning models, including Point Transformer \citep{zhao2021point} for FGI-PointTransformer-DL-3D, dynamic graph convolutional net \citep{wang2019dynamic} for FGI-DGCNN-DL-3D, PointNet \citep{qi2017pointnet} for FGI-PointNet-DL-3D, Point2Vec \citep{zeid2023point2vec} for FGI-Point2Vec-DL-3D, and PointNet++ \citep{qi2017pointnet++} for TUW-PointNet++-DL-3D and FBK-PointNet++-DL-3D.
For most of the methods, nearest neighbor interpolation is used to represent the multispectral point cloud in a format, where each point contains the radiometric information from all of the three channels. As an exception, TUW-PointNet++-DL-3D does not perform nearest neighbor interpolation. Instead, each point is assigned a single intensity value, and three flag variables indicate the source channel of the intensity. 

There were differences in the used  radiometric features and echo properties between the different models in the case of the HeliALS dataset that included the intensity, amplitude, reflectance, echo deviation and an echo return number for each point.  FGI-PointTransformer-DL-3D, FGI-DGCNN-DL-3D, FGI-PointNet-DL-3D, and FGI-Point2Vec-DL-3D utilized all of the radiometric features and echo properties, implying that each 3D point was assigned  13 additional radiometric and echo features. TUW-PointNet++-DL-3D only utilized the intensities and FBK-PointNet++-DL-3D only considered the reflectances in the model input. For FGI-PointTransformer-DL-3D, we also studied different combinations of the included radiometric and echo features in order to evaluate their effect on the classification accuracy. In FGI-PointTransformer-VUX-DL-3D, we utilized the geometric information from the three scanners and radiometric and echo features only from the VUX scanner. As a further variant, FGI-PointTransformer-VUXReflectance-DL-3D only utilized the reflectance of the VUX scanner while disregarding all other radiometric and echo features. Finally, FGI-PointTransformer-GeometryOnly-DL-3D utilized no radiometric or echo information from any of the scanners, thus serving as a baseline when no radiometric information is available. 

When it comes to the format of the input data, most of the software implementations of the studied architectures, such as DGCNN, PointNet++ and PointNet, required a fixed point count per segment.  The input point count ranges from 500 points per segment for FBK-PointNet++-DL-3D to 8192 points per segment for FGI-PointTransformer-DL-3D, FGI-DGCNN-DL-3D, FGI-PointNet-DL-3D, and TUW-PointNet++-DL-3D. Importantly, the software implementation of the Point Transformer architecture does not set constraints to the point count, but we still used a fixed point count to obtain a fair comparison between the different architectures. Before feeding the points to the model, FGI-PointTransformer-DL-3D, FGI-DGCNN-DL-3D, FGI-PointNet-DL-3D normalize the coordinates of the input point cloud into a unit sphere, whereas TUW-PointNet++-DL-3D and FBK-PointNet++-DL-3D center the coordinates, hence preserving the scale. 

Some of the studied methods attempt to mitigate the class imbalance of the training set. FGI-PointTransformerWeighted-DL-3D, FGI-Point2Vec-DL-3D and TUW-PointNet++-DL-3D utilize a weighted loss function to provide a higher weight to minority species, such as oak, alder and rowan. FBK-PointNet++-DL-3D instead augments the training dataset by generating geometric augmentations for rowan and linden, and geometric and spectral augmentations for alder and oak. On the other hand, TUW-PointNet++-DL-3D and the methods by FGI use augmentations to all segments during training without any specific focus on the minority species. The methods by FGI apply small random translation, scaling, and jittering of coordinates, as well as small random scaling and jittering of other attributes. TUW-PointNet++-DL-3D generates six rotational augmentations around the z-axis for each segment.

\begin{table*}[ht!]
\centering
\caption{\label{tab: methods DL 2D} Summary of profile-based (2D) deep learning methods for species classification. For each method, the table briefly describes point cloud pre-processing steps, input data format of the deep learning model, architecture of the model, training strategies, and inference strategies. }
\vspace{0.2 cm}
\resizebox{\textwidth}{!}{
\begin{tabularx}{1.45\textwidth} {
>{\hsize=.7\hsize\linewidth=\hsize}X
>{\hsize=1.05\hsize\linewidth=\hsize}X
>{\hsize=1.05\hsize\linewidth=\hsize}X
>{\hsize=1.2\hsize\linewidth=\hsize}X
>{\hsize=1\hsize\linewidth=\hsize}X
  }
 \specialrule{1.0pt}{0em}{0em}
  \textbf{Method and data} & \textbf{Pre-processing} & \textbf{Input data} & \textbf{Architecture and Training} & \textbf{Inference} \\
 \specialrule{1.0pt}{0em}{0em}
SLU-YOLOv8-DL-2D, \mbox{HeliALS}  &  For each segment, points from VUX and mini-VUX were merged and the point cloud was projected to four 2D RGB images corresponding to different rotations around the z-axis (0$^\circ$, 45$^\circ$, 90$^\circ$, or 135$^\circ$).  Green channel was not used. & $4\times \mathrm{2D}$ RGB images ($160 \times 320$ pixels) per segment.  For each pixel, the RGB values correspond to a mean mini-VUX intensity (NIR), mean VUX intensity (SWIR), and a combination of the two, respectively.  & YOLOv8 \citep{Yolov82023} pretrained on the ImageNet dataset \citep{deng2009imagenet} was finetuned over 15 training epochs. The RGB images for rotation angles 0$^\circ$, 45$^\circ$, and 90$^\circ$ were used in the training set, while the RGB images for the angle 135$^\circ$ formed the validation set. &  Trained YOLOv8 model predicted the species for each of the four RGB images generated from a single segment, and the results were aggregated to obtain the final prediction.  \\
\specialrule{1.0pt}{0em}{0em}
DetailView-DL-2D, \mbox{HeliALS}  & Removal of noise and ground points. Voxelization with a side length of 5 cm to obtain a more balanced point distribution. Point clouds of the three channels merged. Intensity and reflectance ignored. &  $(4+1+1+1)\times \mathrm{2D}$ depth images ($256 \times 256$ pixels) per segment. Intensity information was not used. Out of the seven depth images, four were from the sides, one from the top, one from the bottom, and one representing a close-up between $z=1.0$ m and $z=1.5$ m. & DetailView neural net model \citep{puliti2025benchmarking, Detailview2024} based on	DenseNet-201-instances \citep{huang2017densely} trained from scratch over 24 epochs using 90-10 train-validation split. The batch size was set twice as large and the learning rate a factor of 10 smaller compared to \citet{puliti2025benchmarking}. Augmentations included point cloud random subsampling and rotations around the z-axis. & Predicted class probabilities averaged across 50 augmentations of each segment. \\
\specialrule{1.0pt}{0em}{0em}
SA-Detailview-DL-2D, \mbox{HeliALS}  &  Point clouds of mini-VUX (NIR) and VUX (SWIR)  merged. Removal of ground points.  The green channel was ignored.
& Same as DetailView-DL-2D. &  DetailView neural net model \citep{puliti2025benchmarking, Detailview2024} based on	DenseNet-201-instances \citep{huang2017densely}. The model had been previously trained on the FOR-species dataset with 33 different species and 98-2 training-validation split using same augmentations as DetailView-DL-2D. \textbf{The model was NOT re-trained using the provided training set.} &  Predicted class probabilities averaged across 50 augmentations of the segment. If predicted tree species was not present in the current dataset, it was reclassified to a species in the dataset based on the largest relative accuracy in the confusion matrix.  \\
\specialrule{1.0pt}{0em}{0em}
NTNU-ConvNeXt-T-DL-2D, \mbox{Optech Titan}  &  Point clouds of the three scanners  merged. For each segment, the point cloud was converted to 6 depth images corresponding to 4 different rotations around the z-axis and 2 projections along the negative and positive z-axis following \citet{goyal2021revisiting}. Furthermore, a front-view depth image was obtained between $z=1.0$ m and $z=2.5$ m.& $(6+1)\times \mathrm{2D}$ depth images ($128 \times 128$ pixels that was rescaled before input layer to $256 \times 256$ pixels) per segment. Furthermore, 20 additional global features were computed for each segment based on the tree height and intensities of the 3 channels following \citet{yu2017single}. Single-channel intensity features included, e.g., minimum, maximum, and moments.  &  Neural net with 3 input branches corresponding to the 6 depth images from the sides, the front-view depth image, and the global features. ConvNeXt-T \citep{liu2022convnet} provided the backbone for the two image branches, and a multi-layer perceptron block for the global features. Training-validation split of 80-20 with data augmentations consisting of random rotations around the z-axis, translations and scaling. &  Prediction of the trained model.  \\
\specialrule{1.0pt}{0em}{0em}
\end{tabularx}}
\end{table*}

At inference time, the methods by FGI utilize an ensemble voting strategy based on five trained models obtained with different random initializations. TUW-PointNet++-DL-3D and FBK-PointNet++-DL-3D directly utilize the prediction of the trained model without any ensemble voting strategies.

\subsubsection{Profile-based (2D) deep learning methods}
\label{sec: profile-based deep learning methods}

We summarize the profile-based, i.e., 2D deep learning methods in Table~\ref{tab: methods DL 2D}, where we present relevant details for each of the methods, including information on point cloud pre-processing, format of the input data, model architecture, training, and inference. 

Our comparison of 2D deep learning approaches includes four methods: SLU-YOLOv8-DL-2D, DetailView-DL-2D, SA-Detailview-DL-2D, and NTNU-ConvNeXt-T-DL-2D.  We note that SA-Detailview-DL-2D represents a sensor agnostic (SA) deep learning model that has been previously trained on the FOR-species20K dataset \citep{puliti_2024_13255198} without any re-training on the provided training dataset. All the other models, including Detailview-DL-2D, have been trained on the training set provided as a part of the benchmarking competition. When it comes to the used spectral features, NTNU-ConvNeXt-T-DL-2D utilized radiometric information from all of the three channels. SLU-YOLOv8-DL-2D took into account both geometry and spectral information from VUX and mini-VUX scanners, while ignoring the data of the green channel. In contrast, DetailView utilized only the geometry of the multispectral point cloud without any radiometric information. 

The studied 2D deep learning methods are based on various convolutional neural net backbones, including YOLOv8 \citep{Yolov82023} for SLU-YOLOv8-DL-2D, DenseNet-201 \citep{huang2017densely} for DetailView, and ConvNeXt-T \citep{liu2022convnet} for NTNU-ConvNeXt-T-DL-2D. DetailView and NTNU-ConvNeXt-T-DL-2D convert the segment-wise point clouds into multiple depth image projections from different angles. On the other hand, SLU-YOLOv8-DL-2D projects the point clouds into multiple RGB images, where the RGB values of each pixel are based on the mean intensities of VUX, mini-VUX and their combination. As their input, all the models take between four and seven 2D projections from different viewing angles. The resolutions of the projections range from $128 \times 128$ pixels for NTNU-ConvNeXt-T-DL-2D to $256 \times 256$ pixels for DetailView and $160 \times 320$ pixels for SLU-YOLOv8-DL-2D. NTNU-ConvNeXt-T-DL-2D rescales the point cloud projection to $256 \times 256$ pixels before the input layer of the model. All of the methods use side-view projections corresponding to different rotations around the z-axis. Furthermore, DetailView utilizes a top-view projection, a bottom-view projection, and a projection to a close-up of the trunk. In addition to the side-, top- and bottom-view projections, NTNU-ConvNeXt-T-DL-2D includes a front-view depth image and 20 additional global features mostly based on the intensities of the three channels.
While most of the methods utilize orthographic projections, NTNU-ConvNeXt-T-DL-2D follows the approach by \cite{goyal2021revisiting} and performs the projection using a pinhole camera model (perspective projection).

\begin{table*}[ht!]
\centering
\caption{\label{tab: methods ML} Summary of machine learning methods for species classification.  For each method, the table briefly describes point cloud pre-processing steps, features of the classifier, type of the classifier, and training strategies. }
\vspace{0.2 cm}
\resizebox{\textwidth}{!}{
\begin{tabularx}{1.45\textwidth} {>{\hsize=.6\hsize\linewidth=\hsize}X
>{\hsize=1.4\hsize\linewidth=\hsize}X
>{\hsize=.8\hsize\linewidth=\hsize}X
>{\hsize=1.2\hsize\linewidth=\hsize}X
  }
 \specialrule{1.0pt}{0em}{0em}
  \textbf{Method and data} & \textbf{Pre-processing and features} & \textbf{Input data} & \textbf{Classifier and Training} \\ 
 \specialrule{1.0pt}{0em}{0em}
 IBL-BalancedRF-ML, \mbox{Optech Titan}  & Segments with under- or over-segmentation errors were identified and removed, which resulted in 660 training segments. Following \citet{kaminska2021single} and \citet{lisiewicz2025comprehensive}, features were computed, including geometric features based on the merged point cloud, and intensity features based on individual channels and their combinations. For most features, calculations were performed for points situated above half the height of each segment.  & A feature vector of 30 elements after dimensionality reduction. Variables based on combinations of different channels had the highest importance. & Random forest model was trained with a balanced RF approach \citep{chen2004using}, up-sampling minority species. To ensure stable results and mitigate overfitting, 5-fold cross-validation was repeated 20 times. \\
 \specialrule{1.0pt}{0em}{0em}
UEF-RF/LGBM/SVM-ML, \mbox{Optech Titan}  & Following  \citet{holmgren2004identifying}, \citet{orka2009classifying}, \citet{orka2012simultaneously}, and \citet{hovi2016lidar}, intensity features were computed, including single-channel moments as a function of height for each echo type, and two multi-channel features known as the normalized difference water index and conifer index \citep{trier2018tree}. The only geometric feature was the 95th percentile of the $z$ coordinate.  & A feature vector of 143 elements after dimensionality reduction and 259 elements prior to dimensionality reduction. In the dimensionality reduction, correlation coefficient was evaluated for each pairs of features, and one of the features in the pair was removed if the correlation coefficient exceeded 0.9. & Three ML methods were tested for the classification, including random forest, light gradient boosting machine (LGBM), and support vector machine (SVM) using radial basis function (RBF) kernel. 80-20 train-test split used. Grid search and cross validation were utilized to optimize hyperparameters. No augmentation or species imbalance mitigation. \\ 
\specialrule{1.0pt}{0em}{0em}
IDEAS-RF-ML, \mbox{Optech Titan} & Features were extracted separately from the full tree point cloud (with the bottom 2 m excluded) and crown points (isolated with a 1 m buffer around CHM). In total, 122 features were computed across both point clouds, including point count, height distribution metrics, intensity statistics per each channel, and features based on correlations between height and intensity.
& A feature vector of 122 elements before dimensionality reduction and 90 elements after dimensionality reduction. & Random forest classifier trained using 80-20 train-test split. Optimal number of features and hyperparameters were selected using a grid search by optimizing the \(F_1\) score. No augmentation or species imbalance mitigation. \\
\specialrule{1.0pt}{0em}{0em}
FGI-RF-ML, \mbox{HeliALS} and Optech Titan  & 92 features computed for each segment before dimensionality reduction. Following \citet{lehtomaki2015object}, \citet{yu2017single}, and \citet{hakula2023individual}, features included geometric features using the channel with most points, intensity statistics including, e.g., normalized intensity histograms for each channel, proportions of laser pulse return types (e.g. first of many), and histogram of local descriptors.   & A feature vector of 92 elements before dimensionality reduction and 20--40 elements after dimensionality reduction. & Random forest classifier trained based on out-of-bag prediction. During training, feature selection and model optimization were performed using a grid search based on the out-of-bag prediction. No augmentation or species imbalance mitigation. \\
\specialrule{1.0pt}{0em}{0em}
LUKE-MultiRF-ML, \mbox{Optech Titan}  & 36 features computed for each segment, including, e.g., moments and quantiles of the $z$-coordinate and the radial distance from the center of the segment, average intensities of the channels, average intensities of the first and other echoes for each channel, and features combining intensities of two channels.  
& A feature vector of 36 elements. & 10 random forest models trained using 5-fold cross validation, including a default model classifying segments into the nine species and nine single-species models. Synthetic minority over-sampling technique \citep{chawla2002smote} used to balance the training set with the oversampling parameter set to 10 for pine, spruce, and birch, and to 15 for the rest. If exactly one of the single-species classifiers predicted the segment to belong to its species at inference time, the segment was assigned to this species. Otherwise, the species of the segment was assigned based on the default model.  \\ 
\specialrule{1.0pt}{0em}{0em}
UPV-GB-ML, \mbox{Optech Titan}  &  Point clouds were filtered to remove ground points and outliers. In total, 384 geometric features and 48 single-channel intensity features (16 for each channel) were computed. Single-channel intensity features included, e.g., min, mean, standard deviation and different percentiles. Similar statistical metrics were evaluated for geometrical neighborhood features, all included in the Class3Dp software \citep{carbonell2024class3dp}, such as, number of returns, eigenvalues of the covariance matrix,  and features based on \citet{west2004context}, \citet{rusu2010semantic}, and \citet{demantke2012streamed}. & A feature vector of 18 elements after dimensionality reduction. The dimensionality reduction was based on ranking the features based on their permutation importance and then evaluating the learning curve. The learning curve plateaued after 18 features.   & Gradient boosting (GB) machine used for classification. The training was performed using 5-fold cross validation. No augmentation or species imbalance mitigation.  \\
\specialrule{1.0pt}{0em}{0em}
Aalto-RF-ML, \mbox{Optech Titan}  & 34 features computed for each segment, including e.g., intensity statistics for each channel based on points within the top 3 m of the segment, crown diameter within and below the top 3 m of the segment, the apex angle of the tree, and features combining intensities of two channels following \citet{hakula2023individual}.   & A feature vector of 34 elements. & Random forest classifier trained using 80-20 training validation split. To mitigate class imbalance, new feature vectors were simulated for each species with fewer than 125 examples using the class-wise mean and standard deviation of the feature vector \citep{ronnholm2022utilising}. After augmentation, each class had at least 125 training examples. \\
\specialrule{1.0pt}{0em}{0em}
\end{tabularx}}
\end{table*}

To generate more training examples, DetailView utilizes point cloud random subsampling and rotations around the z-axis as augmentations. SLU-YOLOv8-DL-2D utilizes four rotational augmentations around the z-axis per segment, while NTNU-ConvNeXt-T-DL-2D applies random translation and scaling in addition to random rotations around the z-axis. At inference time, SLU-YOLOv8-DL-2D and DetailView use an ensemble voting strategy to obtain the final prediction, whereas NTNU-ConvNeXt-T-DL-2D directly uses the output of the model for prediction. DetailView performs the prediction for 50 different augmentations of the segment increasing the required amount of computation.

\subsubsection{Machine learning methods}
\label{sec: machine learning methods}

We summarize the studied machine learning classifiers in Table~\ref{tab: methods ML}, where we present relevant details for each of the methods, including information on point cloud pre-processing, feature extraction, classifier type, model training, and inference. 

Our comparison includes the following methods: IBL-BalancedRF-ML, UEF-RF/LGBM/SVM-ML, IDEAS-RF-ML, FGI-RF-ML, LUKE-MultiRF-ML, UPV-GB-ML, and Aalto-RF-ML. 
Most of the studied methods, such as IBL-BalancedRF-ML, UEF-RF-ML, IDEAS-RF-ML, FGI-RF-ML, LUKE-MultiRF-ML, and Aalto-RF-ML,  utilize a random forest classifier or its variant for the species prediction. IBL-BalancedRF-ML applies a balanced RF procedure \citep{chen2004using} to up-sample the number of training examples for minority species, whereas LUKE-MultiRF-ML utilizes multiple RF models for the species prediction.  In addition to the RF classifiers, our comparison includes a support vector machine (SVM) used by UEF-SVM-ML, and variants of gradient boosting (GB) techniques used by UEF-LGBM-ML and UPV-GB-ML.

All of the studied methods utilize hand-crafted geometric and radiometric features from the three channels, though the number of features and details vary from one method to another. For example, UEF-RF/LGBM/SVM-ML utilizes only a single geometric feature, whereas UPV-GB-ML evaluates 384 geometric features before dimensionality reduction. Apart from IDEAS-RF-ML and UPV-GB-ML, all the  methods use features combining intensities from multiple channels. Furthermore, several methods, such as UEF-RF/LGBM/SVM-ML, IDEAS-RF-ML, and FGI-RF-ML, include intensity histograms or intensity statistics as a function of height as their features. Laser echo properties are considered in the features by UEF-RF-ML, FGI-RF-ML, and LUKE-MultiRF-ML. Apart from LUKE-MultiRF-ML and Aalto-RF-ML, the studied methods apply some dimensionality reduction strategy to determine the most important features. After dimensionality reduction, the number of features varies between 18 and 143 for the different methods.

The majority of the methods do not utilize any strategy for species imbalance mitigation or training data augmentation. IBL-BalancedRF-ML, LUKE-MultiRF-ML, and Aalto-RF-ML use over-sampling strategies for the minority species in an attempt to improve their classification accuracy. At inference time, most of the methods directly use the trained classifier to obtain the prediction. As an exception, LUKE-MultiRF-ML utilizes nine single-species classifiers and one default classifier, which is used for the classification task if the results of the single-species classifiers are not consistent. 

\subsection{Accuracy analysis}
\label{sec: accuracy analysis}

The benchmarking results were evaluated with five main evaluation metrics: overall accuracy, precision, recall, \(F_1\) score and macro-average accuracy.
Additionally, confusion matrices were computed for a subset of the competition results.

The overall accuracy (OA) is defined as the proportion of correct classifications among all samples
\begin{equation}
    \text{OA} (y, \hat{y}) = \frac{1}{n_{\text{samples}}} \sum_{i = 1}^{n_{\text{samples}}} \mathds{1}(\hat{y}_i = y_i),
\end{equation}
where $\hat{y}_i$ is the predicted species of the $i$th segment, $y_i$ is the corresponding ground-truth species, $\mathds{1}(\cdot)$ is an indicator function and $n_{\text{samples}}$ is the total number of ground-truth tree segments.

The species-wise precision, i.e., user's accuracy, and recall, i.e., producer's accuracy, are given as
\begin{equation}
    \text{Precision}_s = \frac{\text{TP}_s}{\text{TP}_s + \text{FP}_s},
\end{equation}
\begin{equation}
    \text{Recall}_s = \frac{\text{TP}_s}{\text{TP}_s + \text{FN}_s},
\end{equation}
where the true positives $\text{TP}_s$, false positives $\text{FP}_s$ and false negatives $\text{FN}_s$ can be obtained from the confusion matrix for each tree species class $s$. In this work, the recall metric is also referred to as the species-wise accuracy.

The $F_1$ score for a single species class is defined as
\begin{equation}
    F_{1}^{(s)} = 2 \cdot \frac{\text{Precision}_s \cdot \text{Recall}_s}{\text{Precision}_s + \text{Recall}_s}.
\end{equation}
Furthermore, macro $F_1$ is the arithmetic mean over all the species-wise $F_1$ scores.

In an attempt to evaluate the performance of the studied methods in the presence of substantially unbalanced dataset, we used the macro-average accuracy.
The macro-average accuracy (also known as macro-average recall) is given as:
\begin{equation}
    \text{Macro-average accuracy} = \frac{1}{|S|} \sum_{s \in S} \frac{\text{TP}_s}{\text{TP}_s + \text{FN}_s}
\end{equation}
where $|S|$ denotes the total number of tree species in the benchmarking dataset.

Moreover, we used the classification error as a performance metric when appropriate: the overall classification error refers to the complement of the overall accuracy, i.e.,
\begin{equation}
    \varepsilon_\mathrm{OA} = 1 - \mathrm{OA},
\end{equation}
where $\varepsilon_\mathrm{OA}$ denotes the overall classification error. Similarly, the macro-average classification error is the complement of the macro-average accuracy.

We further estimated the uncertainty of overall accuracy and macro-average accuracy for each of the classification methods using a bootstrapping approach. Based on the test set of our dataset, we generated 2000 bootstrapping datasets of 5261 tree segments using random sampling with replacement. For each classification method, we recalculated the accuracy metrics for each bootstrapping dataset, resulting in a bootstrapping distribution of 2000 accuracy values. We derived a 95\% confidence interval (CI) by taking the 2.5th and 97.5th percentiles of the distribution. The estimated confidence intervals reflect the uncertainty arising from the finite test set, but do not include, e.g., randomness in the model training.

\subsection{Scaling analysis of classification error}
\label{sec: scaling law methods}

In addition to benchmarking the classification methods on the fixed training and test set introduced in Sec.~\ref{sec: dataset}, we also studied the scaling of the classification error as a function of training set size and point density. We conducted these experiments using the deep learning and machine learning models achieving the highest overall accuracy on the HeliALS dataset in the benchmark contest. In the scaling studies, we excluded 88 segments from the original training set of the benchmarking competition due to non-optimal profile images, sparse point clouds, or building intersection within the segment boundaries. In addition, the species label of 18 segments from the original training set was corrected due to errors detected during the on-site verification. Compared to the test set of the benchmark contest, the dataset for scaling studies included 19 additional segments. In total, the dataset for scaling law studies consisted of 6257 tree segments. Below, this dataset is called the scaling analysis dataset. Importantly, these modifications  to the dataset did not significantly affect the classification results. For example, FGI-PointTransformer-DL-3D obtained an overall accuracy of 87.9\% when trained on the original training set of 1065 segments in the benchmark competition. The same model obtained an overall accuracy of 88.5\% when trained on 1000 segments from the scaling analysis dataset.

In the scaling studies, we estimated the overall and macro-average classification errors of the selected classification models using 5-fold cross-validation. The folds were selected using stratified random sampling to ensure that each fold follows the same species distribution. For the scaling experiments with varying training set size, we utilized 5-fold cross-validation on subsets of the scaling analysis dataset, thus effectively creating training sets ranging from approximately 250 segments to approximately 5000 segments. Importantly, each fold in the subsets had a similar species distribution as the original dataset as a result of another round of stratified random sub-sampling from the folds previously generated from the full dataset. 

For the scaling experiments as a function of point density, we randomly sub-sampled the HeliALS tree segments to a desired point density. We varied the point density from $10^{-1}$ $\mathrm{pts}/\mathrm{m}^2$ representing an extremely sparse point cloud to around 1000 $\mathrm{pts}/\mathrm{m}^2$ representing a dense ALS point cloud. We evaluated the classification performance using 5-fold cross-validation on the entire sub-sampled scaling analysis dataset.

\section{Results and Discussion}
\label{sec: results}

In this section, we evaluate the performance of the studied species classification methods on the multispectral data collected with the HeliALS and Optech Titan systems. First in Sec.~\ref{sec: benchmark of species classification methods}, we present the results of the benchmarking contest and discuss key factors affecting the classification accuracy, such as the use of a deep learning or a machine learning model, model architecture and training strategy. In Sec.~\ref{sec: effect of multispectral}, we analyze the effect of multispectral information on classification accuracy. Sec.~\ref{sec: species wise analysis} discusses the relation between classification accuracy and species.  In Sec.~\ref{sec: scaling laws}, we analyze the scaling of classification accuracy for selected deep learning and machine learning models as a function of training set size and point density to study the impact of these factors in more detail and to highlight the differences of deep learning and machine learning methods.
Subsequently, we investigate  the effect of segmentation quality and crown class on species classification in Sec.~\ref{sec: effect of segmentation quality}. Finally, we provide further discussion and compare our results to previous studies in Sec.~\ref{sec: further discussion}.

\subsection{International benchmarking of species classification methods}
\label{sec: benchmark of species classification methods}

 \begin{figure*}[ht]
    \centering
    \includegraphics[width=\linewidth]{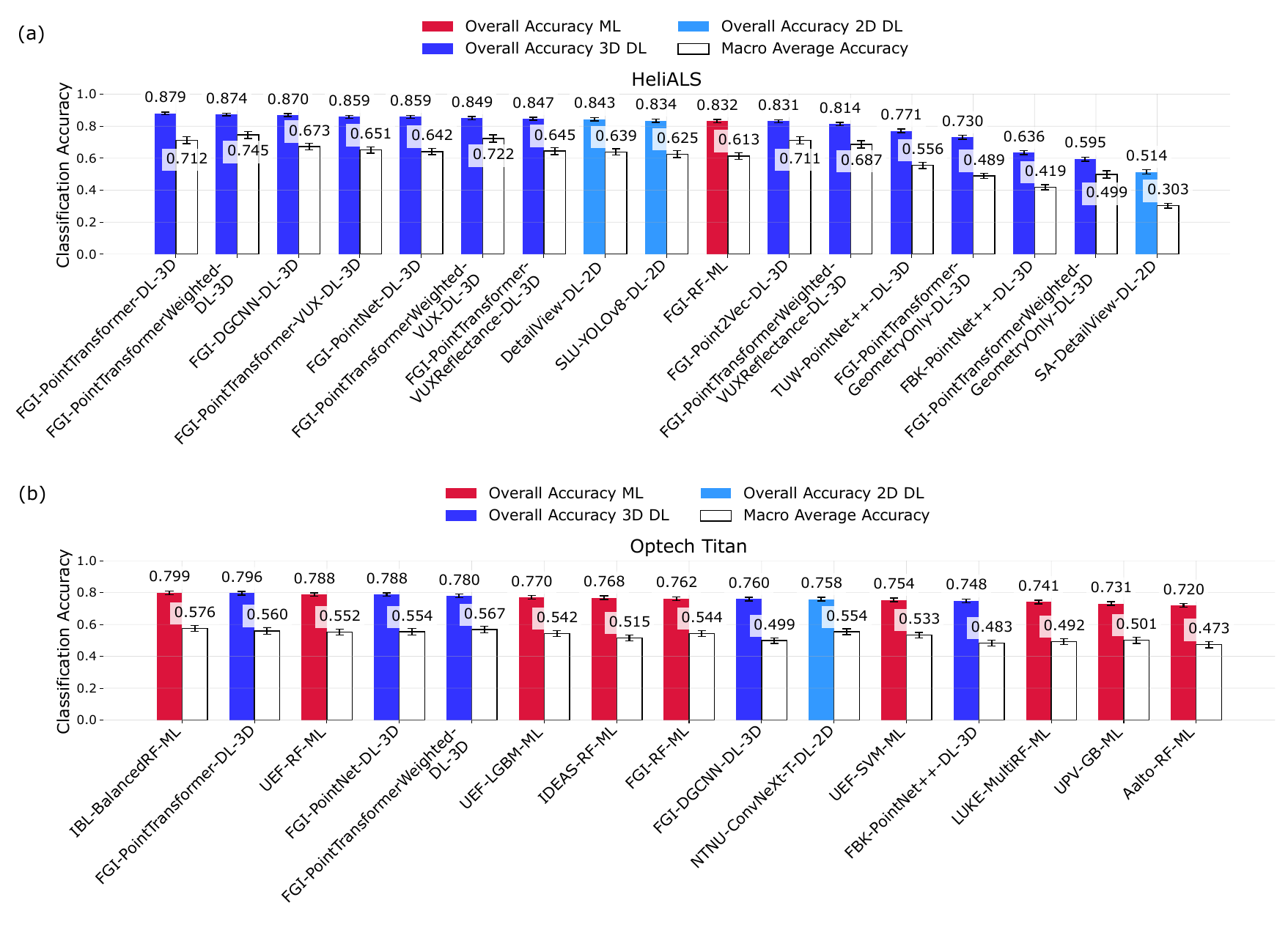}
    \caption{(a) Comparison of overall (filled bar) and macro-average (outline) accuracy for methods using HeliALS data, with point-based DL methods shown with dark blue color, image-based DL methods shown with light blue color, and machine learning methods shown with red color. The error bars represent 95\% confidence intervals based on a bootstrapping approach, see Sec.~\ref{sec: accuracy analysis}. SA-DetailView-DL-2D was not retrained on the training set of the current study, and the model parameters were directly transferred from \citet{puliti2025benchmarking}. When neglecting aspens, rowans and alders not present in the FOR-species20K dataset \citep{puliti2025benchmarking}, SA-Detailview-DL-2D reached an overall accuracy of 58.8\% and a macro-average accuracy of 45.4\%.  (b) Same as (a) but for the methods using  the Optech Titan dataset.}
    \label{fig:comparison}
\end{figure*}

 The species classification methods introduced in Sec.~\ref{sec: participating methods} were benchmarked  using overall accuracy, macro-average accuracy, species-wise precision and recall as the key metrics. We compare the overall and macro-average accuracy for the methods using the HeliALS data in Fig.~\ref{fig:comparison}(a) and for the methods using the Optech Titan data in Fig.~\ref{fig:comparison}(b).  In general, the methods using the HeliALS data reached a higher overall and  macro-average accuracy compared to methods using the Optech Titan data, which can be attributed to  higher point density and geometric accuracy of the point cloud provided by the HeliALS system, see Sec.~\ref{sec: measurement systems}. 
 
 For the HeliALS dataset, FGI-PointTransformer-DL-3D reached the highest overall accuracy of 87.9\%, and the corresponding method with a weighted loss function, FGI-PointTransformerWeighted-DL-3D, achieved the highest macro-average accuracy of 74.5\%. In addition to the point transformer model, the other best performing methods on the HeliALS dataset were also point-based deep learning methods, with FGI-DGCNN-DL-3D and FGI-PointNet-DL-3D reaching an overall (macro-average) accuracy of 87.0\% (67.3\%) and 85.9\% (64.2\%), respectively. The best image-based deep learning method, DetailView-DL-2D, reached an overall (macro-average) accuracy of 84.3\% (63.9\%) and the only machine learning model, FGI-RF-ML, had an overall (macro-average) accuracy of 83.2\% (61.3\%).  Based on the confidence intervals from bootstrapping, the overall accuracy of FGI-PointTransformer-DL-3D was within the confidence intervals of FGI-PointTransformerWeighted-DL-3D and FGI-DGCNN-DL-3D, suggesting that the difference between these methods is not statistically significant. The total width of the 95\% confidence intervals for the overall accuracy ranged between 1.7\% and 2.7\% across the methods using the HeliALS dataset.
 
 For most of the methods, the macro-average accuracy was significantly lower than the overall accuracy due to the inherent difficulty of predicting the minority species, such as alder or oak. The point transformer model clearly outperformed the other methods in the prediction of the minority species, which resulted in a significant improvement of the macro-average accuracy, especially when using the weighted loss function. FGI-PointTransformerWeighted-DL-3D achieved the highest macro-average accuracy even when taking the finite confidence intervals into account. In general, the total width of the 95\% confidence intervals for the macro-average accuracy was 3.1-4.7\% for the methods using the HeliALS data. Since the macro-average accuracy weighs each class equally, rare classes contribute a higher variance, leading to a wider confidence interval compared to overall accuracy.

\begin{figure*}[h!]
    \centering
    \includegraphics[width=\textwidth]{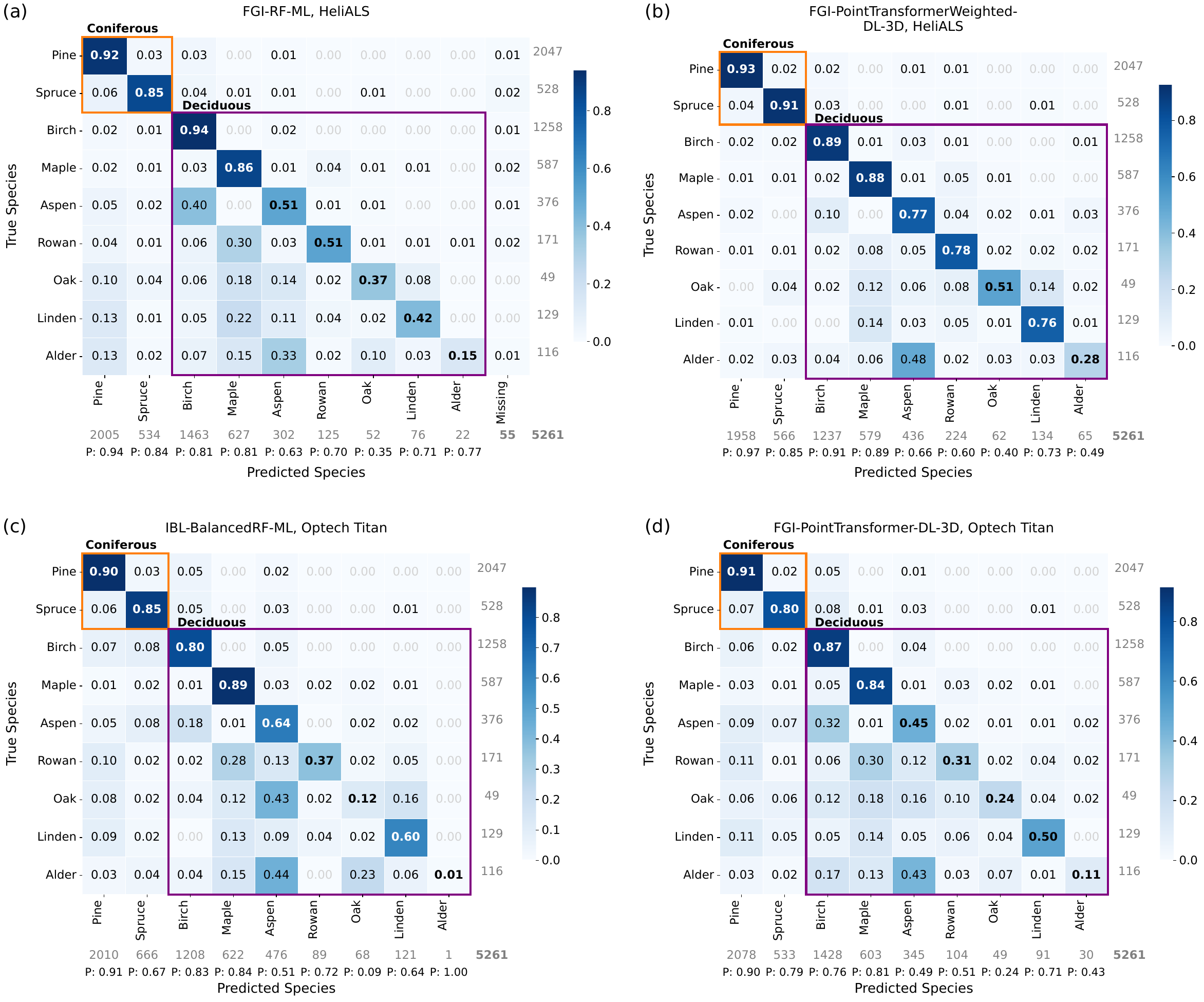}
    \caption{(a) Normalized confusion matrix of FGI-RF-ML obtaining the highest macro $F_1$ score (0.64) among the maching learning models using HeliALS data. FGI-RF-ML produced 55 missing predictions. (b) Normalized confusion matrix of FGI-PointTransformerWeighted-DL-3D obtaining the highest macro $F_1$ score (0.73) among the deep learning models using HeliALS data. (c) Normalized confusion matrix of IBL-BalancedRF-ML obtaining the highest macro $F_1$ score (0.57) among the machine learning models using Optech Titan data. (d) Normalized confusion matrix of FGI-PointTransformer-DL-3D obtaining the highest macro $F_1$ score (0.58) among the deep learning models using Optech Titan data. All confusion matrices were row-normalized to enable comparison across species regardless of the class size. Each cell count was divided by the sum of its true-class row, and thus, each row sums to 1.  Diagonal values represent species-wise recalls. Precisions are displayed below the matrix. Row and column sums are shown to the right and bottom of the matrix in gray.}
    \label{fig:confusion-matrices}
\end{figure*}

 For the Optech Titan dataset, the highest classification accuracy was reached by a random forest classifier, IBL-BalancedRF-ML, which achieved an overall accuracy of 79.9\% and a macro-average accuracy of 57.6\%. The point transformer method, FGI-PointTransformer-DL-3D, ranked second with an overall accuracy of 79.6\% and a macro-average accuracy of 56.0\%.  For a larger training set size, the deep learning method is expected to outperform the machine learning model as discussed in Sec.~\ref{sec: scaling laws}.   Based on the confidence intervals from bootstrapping, the overall accuracy of IBL-BalancedRF-ML was within the confidence intervals of FGI-PointTransformer-DL-3D, UEF-RF-ML, and FGI-PointNet-DL-3D, implying that the difference between these methods is not statistically significant. The total width of the 95\% confidence intervals ranged between 2.1-2.4\% for the overall accuracy and between 3.6-4.2\% for the macro-average accuracy across the methods using the Optech Titan dataset.
 
 On the Optech Titan dataset, most of the participating methods were RF classifiers or their variants. The best and worst-performing RF models had a difference of 7.9 (10.3) percentage points in terms of the overall (macro-average) accuracy implying that the performance of the various RF approaches was relatively consistent. This suggests that RF is a robust and reliable method, providing reasonable accuracy across different implementations. Based on the submissions by UEF, a RF classifier also outperformed both LGBM and SVM classifiers, though the difference was only a few percentage points. 
 
 It is difficult to draw exhaustive conclusions on the factors causing the performance differences between the studied RF classifiers. However, the method descriptions in Sec.~\ref{sec: machine learning methods} suggest that the best-performing RF classifiers often utilized  multi-channel intensity and echo features as well as various histogram features, such as intensity histograms. Interestingly, intensity features appear to be more important than geometric features since the second best RF method, UEF-RF-ML, only utilized a single geometric feature, but still managed to reach a good performance thanks to various intensity and echo features. When it comes to training strategies, the best performing model, IBL-BalancedRF-ML, cleaned the training dataset by removing under- or over-segmented trees, and applied a balanced RF procedure  \citep{chen2004using} to up-sample the number of training segments for minority species. However, LUKE-MultiRF-ML and Aalto-RF-ML also applied strategies to mitigate class imbalance but they were outperformed by several methods that did not use any strategy for class imbalance mitigation.
 
 Some RF methods, such as FGI-RF-ML, had trouble evaluating all of their features on certain segments due to missing point attributes for some of the channels, thus hampering their prediction. On sparse tree segments, the computation of complicated geometric features may also not always be possible, which may lead to further missing feature values. Thus, sufficiently simple geometric features should be preferred on sparse ALS data. Imputation strategies may also be useful if the point attributes are missing for some of the channels on certain tree segments due to differences in the sensing directions of the different channels.

\begin{figure*}[h!]
    \centering
        \includegraphics[width=0.9\linewidth]{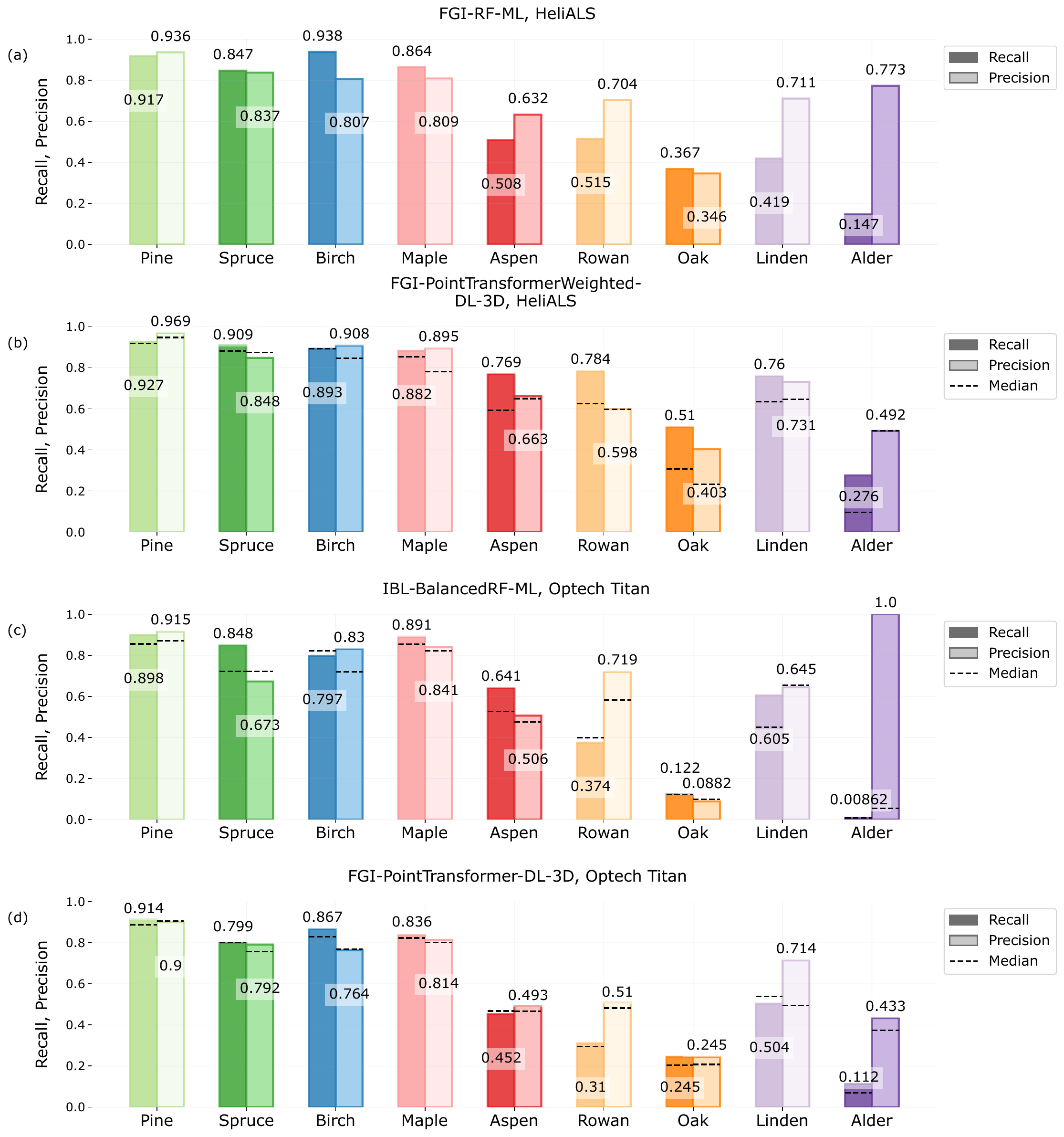}

    \caption{ Precision and recall for the best machine learning and deep learning methods for HeliALS (panels (a) and (b)) 
 and Optech Titan (panels (c) and (d)) datasets. Best methods were chosen based on the macro-average \(F_1\) score. Median values  for precision and recall (dashed horizontal lines) were calculated across the methods utilizing the same dataset and approach (DL/ML). (a) Precision and recall  for FGI-RF-ML utilizing HeliALS data (Macro \(F_1=  0.64\)) (colored bars). (b) Precision and recall for the best DL method, FGI-PointTransformerWeighted-DL-3D, (Macro \(F_1= 0.73\)) (colored bars) and the median across other DL methods (dashed lines) on the HeliALS dataset. (c) Precision and recall for the best ML method (Macro \(F_1= 0.57\)) (colored bars) and the median across other ML methods (dashed lines) on the Optech Titan data. (d) Precision and recall for the best DL method (Macro \(F_1= 0.58 \)) (colored bars) and the median across other DL methods on the Optech Titan data (dashed lines).}
    \label{fig:specieswise-recalls}
\end{figure*}

The spread of the overall and macro-average accuracy metrics was lower on the Optech Titan dataset compared to HeliALS dataset. Perhaps surprisingly, the worst-performing methods using the Optech Titan data were better than the worst-performing methods using the HeliALS data, for which most of the submitted methods were deep learning models. We suspect that the larger spread in the results for the HeliALS dataset is partly caused by the deep learning models having a higher risk of overfitting unless necessary precautions are implemented, such as using a sufficient number of (randomized) augmentations or early stopping. Importantly, the best performing deep learning methods, such as FGI-PointTransformer-DL-3D, FGI-DGCNN-DL-3D, FGI-PointNet-DL-3D, and DetailView-DL-2D applied a large number of different augmentations during training, including point cloud random subsampling, rotations around the z-axis, small random scaling and jittering of coordinates and other features. The worst-performing point-based deep learning method applied augmentations only for the minority species to mitigate class imbalance. Furthermore, the most successful methods typically applied some ensemble strategy at inference time, such as majority voting based on five separately trained classifiers (FGI-PointTransformer-DL-3D) or averaged predictions across a large number of rotational augmentations (DetailView-DL-2D). These observations suggest that deep learning methods have potential for high classification accuracy but care is needed in the choice of training and inference strategies.
 
 Using the HeliALS data, we also tested a sensor-agnostic method, SA-Detailview-DL-2D. The method had been previously trained on the FOR-species20K dataset ~\citep{puliti2025benchmarking}, and we did not re-train it on the current dataset as explained in Sec.~\ref{sec: profile-based deep learning methods}. The method did not utilize any spectral information either. On the full test set, SA-Detailview-DL-2D achieved an overall (macro-average) accuracy of 51.4\% (30.3\%) whereas the overall (macro-average) accuracy was 58.8\% (45.4\%) when neglecting aspen, rowan, and alder not present in the FOR-species20K dataset. However, the performance of the DetailView model was greatly improved by training the model from scratch using the training dataset of our study. This resulted in an overall (macro-average) accuracy of 84.3\% (63.9\%) for DetailView-DL-2D.  The finding suggests that the model architecture itself is promising for species classification, but the model trained on the global FOR-species20K dataset did not generalize well enough for our ALS dataset. Possible reasons behind the poor out-of-domain performance of SA-Detailview-DL-2D are  the relatively small proportion of high-density ALS data compared to TLS and MLS data in the FOR-species20K training dataset, the impact of segmentation artifacts, and the poor representativeness of the species of this study in the FOR-species20K training dataset.

\subsection{Effect of multispectral information}
\label{sec: effect of multispectral}

To investigate the importance of spectral information on species classification, we compared variants of the best-performing Point Transformer model in scenarios, where no intensity information was available (FGI-PointTransformer-GeometryOnly-DL-3D), reflectance information from a single channel was available (FGI-PointTransformer-VUXReflectance-DL-3D), all intensity and echo information from a single channel was available (FGI-PointTransformer-VUX-DL-3D), and all multispectral information from the three channels was available (FGI-PointTransformer-DL-3D). Using geometric information only, we obtained an overall (macro-average) accuracy of 73.0\% (48.9\%), which was further improved to 84.7\% (64.5\%) with a single-channel reflectance, to 85.9\% (65.1\%) with all single-channel information, and to 87.9\% (71.2\%) with spectral information from all the three channels.
Thus, single-channel spectral information reduced the overall error by 48\% and macro-average error by 32\%. The multispectral information further decreased the overall error by 14\% and macro-average error by 17\% compared to using all single-channel spectral information. Importantly, the macro-average classification error could be reduced by a further 11\% by switching to use a weighted loss function, which enabled a more accurate classification of the minority species without sacrificing the overall classification accuracy. 

\subsection{Species-wise analysis}
\label{sec: species wise analysis}

To understand differences in the classification accuracy between different species, we present confusion matrices of the best-performing deep learning and machine learning models for both the HeliALS and Optech Titan datasets in Figs.~\ref{fig:confusion-matrices}(a)--(d).  See Fig.~\ref{fig: confusion_matrices_raw} in \ref{ap: confusion matrices} for the corresponding unnormalized confusion matrices. Furthermore, Figures~\ref{fig:specieswise-recalls}(a)--(d) illustrate the species-wise precision and recall for the best methods on both datasets together with the median of the machine learning or deep learning approaches. Based on Figs.~\ref{fig:confusion-matrices}(a)--(d), we observe that pines, spruces, birches, and maples are reliably classified on both datasets across the models. The other deciduous trees representing minority species (see Fig.~\ref{fig: species visualization and distribution}) are associated with significantly lower species-wise recall and precision. Based on the confusion matrices, we observe  that aspens are often classified as birches, rowans as maples, lindens as maples, alders as aspens, and oaks as all other deciduous trees across the different models and datasets. The difficulty in predicting these species arises partly due to their low frequency in the benchmarking dataset and partly due to a rather similar visual appearance  as further discussed in Sec.~\ref{sec: further discussion}.  For the Optech Titan dataset, we observe no clear qualitative difference between the predictions of the best deep learning and the best machine learning model. 

The dense HeliALS dataset enables a more accurate classification of the minority species. Importantly, the best deep learning method, FGI-PointTransformerWeighted-DL-3D, clearly outperforms the best machine learning method FGI-RF-ML in the prediction of minority species on the HeliALS dataset, which highlights the promise of deep learning for species classification on dense ALS point clouds.  FGI-PointTransformerWeighted-DL-3D achieved a recall of 76.9\% and a precision of 66.3\% for aspen that has an important role in the biodiversity of boreal forests as discussed in Sec.~\ref{sec: introduction}. Using Optech Titan data, IBL-BalancedRF-ML achieved a  recall of 64.1\% and a precision of 50.6\% for aspen. See \ref{ap: confusion matrices} for a full list of normalized confusion matrices for the methods not included in Figs.~\ref{fig:confusion-matrices}(a)--(d).

\subsection{Scaling laws of classification error}
\label{sec: scaling laws}

A significant aspect of machine learning is the influence of training set size on the classification accuracy. Another major aspect is the influence of specific properties of the input data, such as point cloud density. In the following sections we provide detailed analysis of both aforementioned factors.

\subsubsection{Scaling laws with respect to training set size}
\label{sec: scaling law vs training set size}

\begin{figure*}[h]
    \centering
    \includegraphics[width=0.9\textwidth]{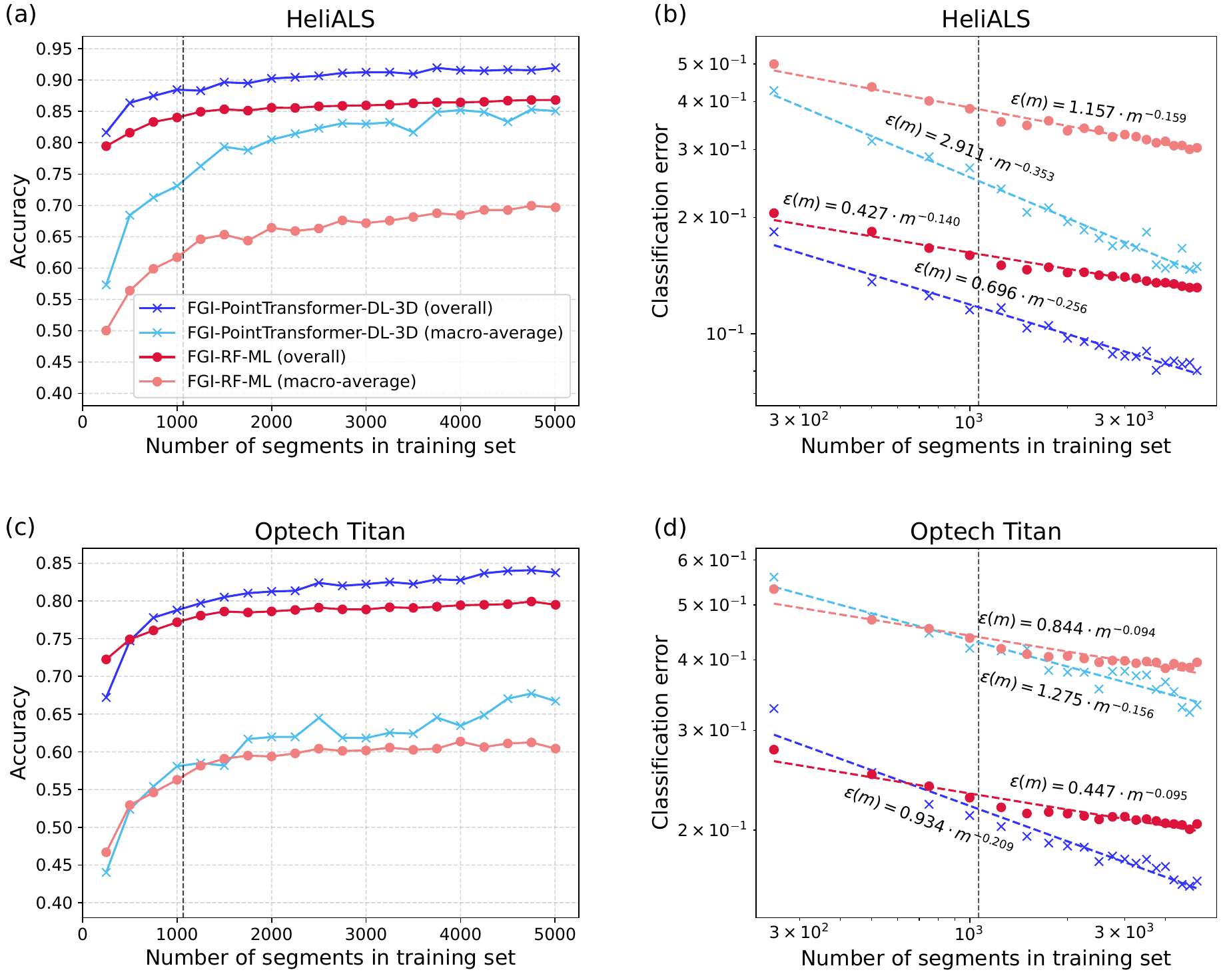}
    \caption{(a) Overall accuracy (dark color) and macro-average accuracy (light color) of FGI-PointTransformer-DL-3D (blue) and FGI-RF-ML (red) as a function of the number of training segments  for HeliALS data. (b) Classification error corresponding to (a) as a function of training set size in logarithmic scale, together with power law fits given by Eq.~\eqref{eq: power law}. (c) Same as (a) but using the Optech Titan dataset. (d) Same as (b) but using the Optech Titan dataset. In all panels, the grey vertical lines denote the size of the training set in the benchmark competition. For the DL method, the training set was further divided into training and validation sets for hyperparameter tuning. }
    \label{fig: scaling laws accuracy vs training set size}
\end{figure*}

\begin{figure*}[h]
    \centering
    \includegraphics[width=\textwidth]{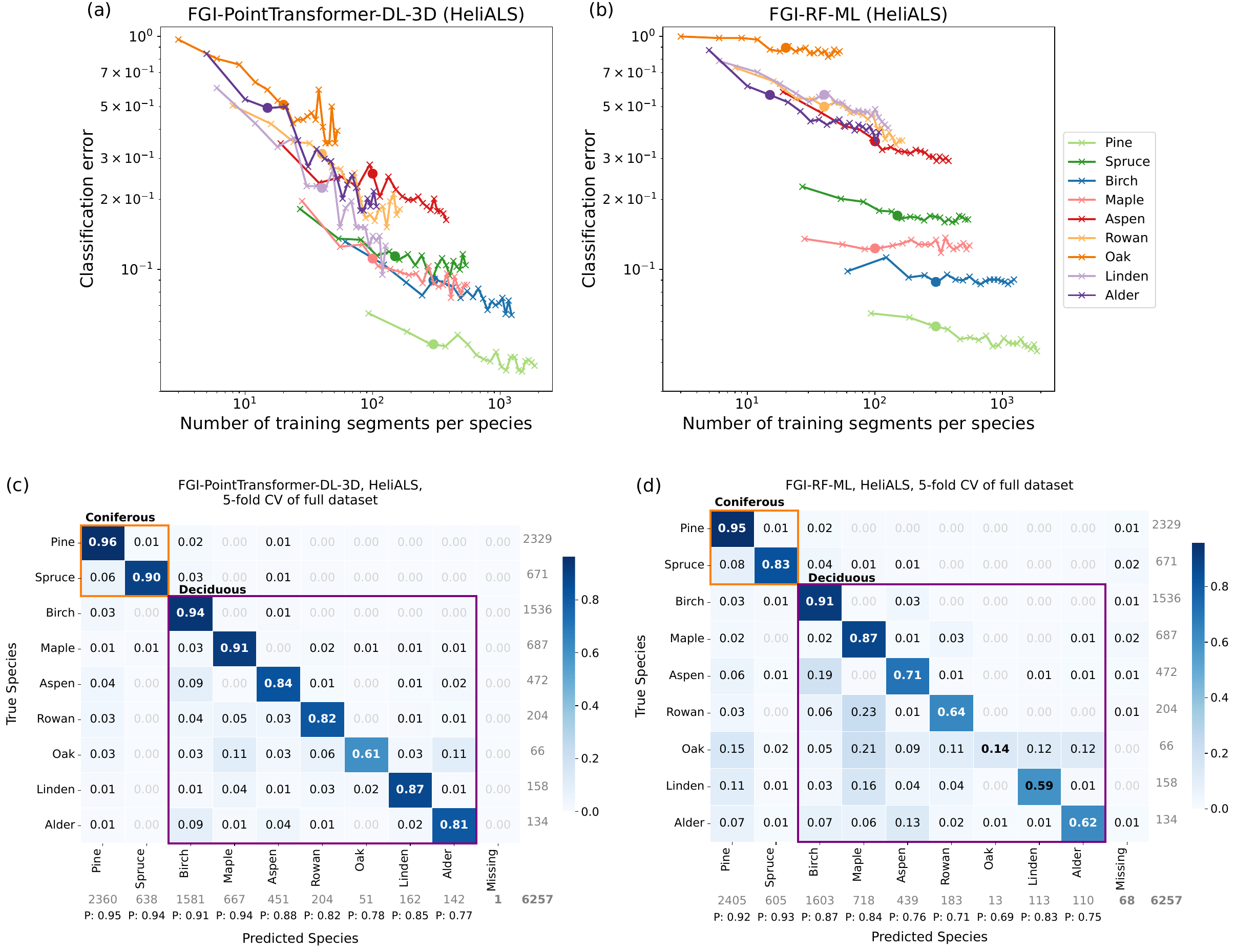}
    \caption{Species-wise classification errors of (a) FGI-PointTransformer-DL-3D and (b) FGI-RF-ML as a function of the number of training segments in logarithmic scale for HeliALS data. The circle markers denote the number of training segments per species in the benchmark competition. For the DL method, the training set was further divided into training and validation sets for hyperparameter tuning. (c) and (d) show the row-normalized confusion matrices of FGI-PointTransformer-DL-3D (macro $F_1 = 0.86$) and FGI-RF-ML (macro $F_1 = 0.73$) for HeliALS data when using 5-fold cross-validation on the full ground-truth species dataset (5000 training segments).}
    \label{fig: recall vs training set size}
\end{figure*}

In this section, we study the effect of the training set size on the species classification accuracy using FGI-PointTransformer-DL-3D as a representative deep learning method and  FGI-RF-ML as a representative machine learning method. See Sec.~\ref{sec: scaling law methods} for details on the implementation of the scaling analysis. We illustrate the overall and macro-average accuracy of the studied models as a function of the training set size for HeliALS and Optech Titan datasets in Figs.~\ref{fig: scaling laws accuracy vs training set size}(a) and (c), respectively. For the HeliALS data, we observe that the deep learning model outperforms the machine learning model already at the training set size of 250 segments. As the number of segments in the training set increases, the difference in the performance of the DL and ML models widens up. For the largest considered training set sizes, the macro-average accuracy of the deep learning method nearly catches up with the overall accuracy of the machine learning method. For a training set size of 5000 segments, FGI-PointTransformer-DL-3D reached an overall (macro-average) accuracy of 92.0\% (85.1\%), whereas FGI-RF-ML achieved an overall (macro-average) accuracy of 86.8\% (69.6\%). 

 For small training sets of the Optech Titan dataset, the ML method outperforms the DL method in terms of both overall and macro-average accuracy. Similarly to the case of the HeliALS data, the accuracy of the DL method improves more rapidly as the size of the training set is increased. Importantly, there is a crossover point at around 700 training segments, above which the overall and macro-average accuracy of the DL model outperform those of the ML model. For the largest considered training set size of 5000 segments, FGI-PointTransformer-DL-3D reached an overall (macro-average) accuracy of 83.7\% (66.7\%), whereas FGI-RF-ML achieved an overall (macro-average) accuracy of 79.5\% (60.4\%).

To gain more quantitative insight into the scaling behavior of species classification, we show the overall and macro-average classification error as a function of the training set size  in a logarithmic scale in Figs.~\ref{fig: scaling laws accuracy vs training set size}(b) and (d) for the HeliALS and Optech Titan data, respectively. Similarly to neural scaling laws for language and vision tasks \citep{hestness2017deep, kaplan2020scaling, bahri2024explaining}, we observe a power law scaling of the classification error $\varepsilon$ following the model
\begin{equation}
\varepsilon(m) = A m^{-\alpha}, \label{eq: power law}
\end{equation}
where $A$ is a multiplier, $m$ denotes the training set size, and $\alpha$ describes the speed of convergence. We observe that the power law scaling applies both for the point transformer model representing deep learning methods and the RF classifier representing machine learning methods. For the HeliALS data in Fig.~\ref{fig: scaling laws accuracy vs training set size}(b), we notice no clear deviation from the power law fit across the studied range of training set sizes spanning an order of magnitude. For the Optech Titan dataset, the macro-average classification error of the ML model appears to slowly deviate from the power law behavior at large training set sizes, though even larger training set sizes would be needed to verify this observation. Importantly, the classification error of the DL method converges approximately twice faster with increasing training set size compared to the ML method. For example on the HeliALS dataset, the DL method is characterized by $\alpha=0.256$ ($\alpha=0.353$) regarding the overall (macro-average) error, whereas the ML method reaches $\alpha=0.14$ ($\alpha=0.159$). Moreover, we observe that the DL method has a higher $\alpha$ and converges faster on the HeliALS dataset than on the Optech Titan dataset, implying that DL methods may provide a larger advantage on dense ALS point clouds with high geometric accuracy.

The power law behavior indicates that the classification accuracy may be further improved by increasing the training set size. According to our results, the benefits of deep learning methods become more apparent on large datasets consisting of thousands of trees, whereas machine learning methods may provide sufficient performance on small datasets of up to a few hundred trees. However, the improvements obtained by increasing the training set size obey a law of diminishing returns. Namely, the relative reduction of the error obtained by increasing the training set size from 1000 segments to 10000 segments is expected to be equivalent to the reduction obtained by increasing the training set size from 100 to 1000 segments. Extrapolating the fitted scaling laws on the HeliALS data, we estimate that reaching a macro-average accuracy above 90\% across the  nine species would require 14 000 training segments using the DL method and up to $4.9 \times 10^6$ segments using the ML approach, highlighting the advantages of deep learning on large datasets. For the Optech Titan data, we similarly estimate that $12 \times 10^6$ segments would be required to reach a macro-average accuracy above 90\% with the deep learning method. However, these estimates should be regarded as rough order of magnitude estimates, and we note that the transferability of the deep learning methods should be studied further as discussed in Sec.~\ref{sec: further discussion}. The good fit of the power law model also highlights the high quality of our field reference dataset, namely, any potential classification errors in our field reference dataset must be significantly rarer than the lowest achieved error of 8.0\% since we observe no clear deviation from the power law scaling. 

We present the species-wise classification error as a function of the number of training examples for the DL model in Fig.~\ref{fig: recall vs training set size}(a) and for the ML model in Fig.~\ref{fig: recall vs training set size}(b). Using the DL model, we observe a power law scaling of the classification error for each of the species. However, the minority species, such as oak, linden, and alder, have a significantly higher classification error compared to the majority species, such as pine, birch, spruce, and maple, for a given number of training segments.   Pine trees are  the easiest species for the classification, and a classification error of around 6\% is reached with only 100 pine trees in the training set. Using the RF model, the classification error of the majority species appears to stagnate at large training set sizes, whereas the classification error of the minority species continues to improve with increased training set size, with the exception of oak that is hardly learned by the RF model.

\subsubsection{Scaling laws with respect to point density}
\label{sec: scaling law vs point density}

 \begin{figure*}[ht!]
    \centering
    \includegraphics[width=0.92\textwidth]{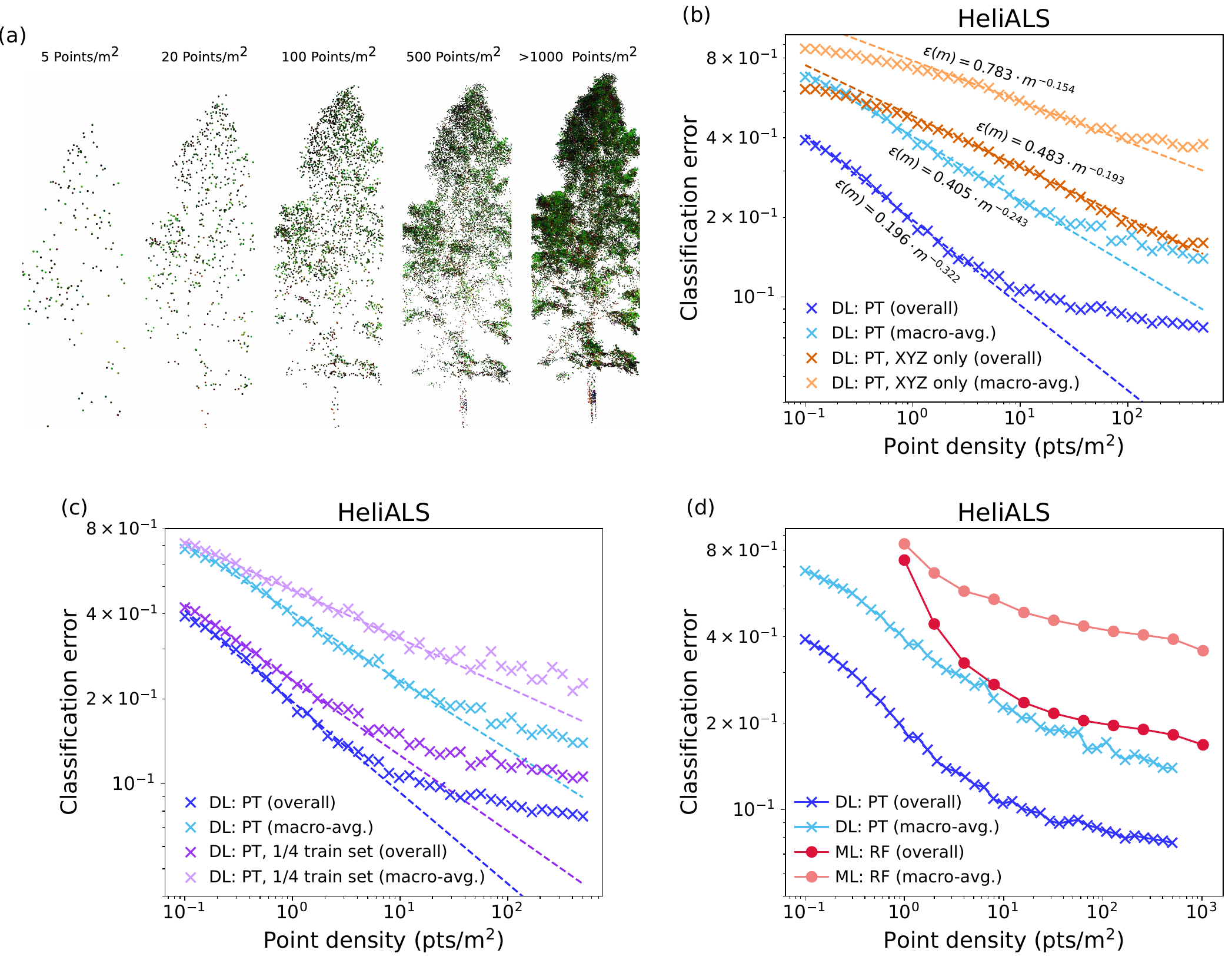}
    \caption{(a) Point cloud projections of an example segment with varying point density obtained as a result of sub-sampling of the HeliALS point cloud. The red, green and blue color channels represent the laser pulse return intensities at the wavelengths $\lambda_1 = \text{532 nm}, \lambda_2 = \text{905 nm}$ and $\lambda_3 = \text{1550 nm}$, respectively. (b) Overall (dark color) and macro-average (light color) classification errors as a function of  point density for FGI-PointTransformer-DL-3D (PT) when using all features (blue markers) and when using only geometric features (orange markers) on the HeliALS dataset. The dashed lines show fits of Eq.~\eqref{eq: point density scaling}. (c) Overall (dark color) and macro-average (light color) classification errors as a function of point density for FGI-PointTransformer-DL-3D when trained with the full HeliALS training set (blue markers) and when trained with 25\% of the training set (purple markers). (d) Overall (dark color) and macro-average (light color) classification errors as a function of point density for FGI-PointTransformer-DL-3D (blue markers) and FGI-RF-ML (RF, red markers) when trained with the full HeliALS training set.}
    \label{fig: classification error wrt point density}
\end{figure*}

 We also study the classification error as a function of point density by randomly sub-sampling the segments to a desired point density as illustrated in Fig.~\ref{fig: classification error wrt point density}(a) and explained in Sec.~\ref{sec: scaling law methods}. In Figs.~\ref{fig: classification error wrt point density}(b)-(d), we compare the classification error as a function of point density for FGI-PointTransformer-DL-3D representing a baseline DL model, FGI-PointTransformer-GeometryOnly-DL-3D using only geometric information, a variant of the baseline DL model using only 25\% of the training set, and FGI-RF-ML representing a ML model. We observe that the baseline DL model exhibits two regimes of convergence. At low point densities below 10~$\mathrm{pts}/\mathrm{m}^2$, the error of the baseline DL model exhibits rapid convergence following a power law 
\begin{equation}
    \varepsilon (\sigma) = A\sigma^{-\alpha}, \label{eq: point density scaling}
\end{equation}
where $\sigma$ denotes the point density per area. The overall (macro-average) classification error of the baseline DL model approaches 10.5\% (22.7\%) already at a point density of 10~$\mathrm{pts}/\mathrm{m}^2$. For point densities above 10~$\mathrm{pts}/\mathrm{m}^2$, both the overall and macro-average error continue to decrease with increasing point density but with a significantly lower rate. In this regime, the classification error potentially follows another power law, but with different parameters.

The geometry-only DL model exhibits 2--4 times higher overall and macro-average classification error across the studied point densities compared to the baseline DL model utilizing multi-spectral information as shown in Fig.~\ref{fig: classification error wrt point density}(b). Interestingly, the geometry-only DL model shows a slower convergence of the classification error at low point densities compared to the baseline DL model, but the power law scaling extends across a larger range of  point densities up to $\geq 100$ $\mathrm{pts}/\mathrm{m}^2$. Thus,  the geometry-only DL model begins to catch up with the baseline model for point densities above 10~$\mathrm{pts}/\mathrm{m}^2$. This also suggests that the improvement provided by the multi-spectral information is the largest for point densities around 10~$\mathrm{pts}/\mathrm{m}^2$, which highlights the advantage of multi-spectral information especially on sparse ALS datasets. For the largest considered point density of around 500~$\mathrm{pts}/\mathrm{m}^2$, the overall classification error of the geometry-only DL model still remains a factor of two higher than that of the baseline DL model using multispectral information. The observed results confirm similar findings from a multitude of applications relating to multispectral- and hyperspectral laser scanning \citep{takhtkeshha2024multispectral}.

\begin{figure*}[h!]
    \centering
    \includegraphics[width=\linewidth]{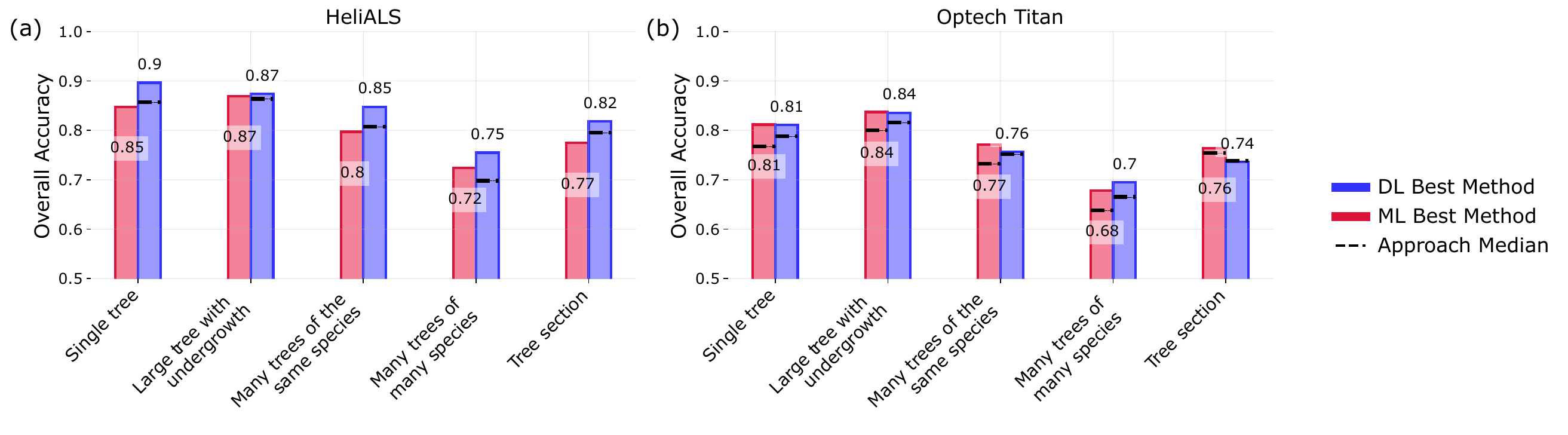}
    \caption{Overall accuracy by profile category for the best machine learning and deep learning methods for (a) HeliALS  and (b) Optech Titan datasets. Dashed lines indicate median values of overall accuracy across the methods utilizing the same dataset and approach type (DL/ML). The best method was determined by macro \(F_1\) score, which is the unweighted average of the \(F_1\) scores across all species. (a) For the HeliALS dataset, the best deep learning method is FGI-PointTransformerWeighted-DL-3D (Macro \(F_1 = 0.73\)) and the best machine learning method is FGI-RF-ML (Macro \(F_1 = 0.64\)). (b) For the Optech Titan dataset, the best deep learning method is FGI-PointTransformer-DL-3D (Macro \(F_1 = 0.58\)) and the best machine learning method is IBL-BalancedRF-ML (Macro \(F_1 = 0.57\)).}
    \label{fig: OA results for profile categories}
\end{figure*}

In Fig.~\ref{fig: classification error wrt point density}(c), we compare the classification error of the baseline DL model when using the full training set and when using one fourth of the training set. We see that the classification error reduces more rapidly as a function of the point density when using the larger training set, especially in the regime of low point density. However, classification error begins to deviate from the power law scaling around 10~$\mathrm{pts}/\mathrm{m}^2$ for both training set sizes.

In Fig.~\ref{fig: classification error wrt point density}(d), we compare the classification error of the DL model and the ML model as a function of the point density. For a low point density below 5~$\mathrm{pts}/\mathrm{m}^2$, the error of the ML model rapidly increases with decreasing point density. For point densities below 1~$\mathrm{pts}/\mathrm{m}^2$, some of the features used by the RF classifier can no longer be computed. In contrast, the DL model still reaches a decent performance in this regime of point densities. Above a point density of 10~$\mathrm{pts}/\mathrm{m}^2$, both the ML model and DL model show a slow but steady improvement of the classification error with increasing point density.

\subsection{Effect of segmentation quality on classification accuracy}
\label{sec: effect of segmentation quality}

In this section, we study the classification accuracy across different profile categories and tree crown classes introduced in Sec.~\ref{sec: segment categories}, see also Fig.~\ref{fig: profile category distributions}. We show the overall classification accuracy for the best deep learning method and the best machine learning method in the benchmarking study together with the corresponding median values across the five profile categories using the HeliALS data in Fig.~\ref{fig: OA results for profile categories}(a) and using the Optech Titan data in Fig.~\ref{fig: OA results for profile categories}(b). For both the HeliALS and Optech Titan datasets, the highest overall accuracy is reached for the categories `Single tree` and `Large tree with undergrowth`, which cover a large majority of the dataset as shown in Fig.~\ref{fig: profile category distributions}(b). This is expected since these profile categories represent ideal segments with no nearby growing trees of the same or different species. On both datasets, the worst classification accuracy is obtained for the profile category `Many trees of many species` representing a segment, in which the largest tree is accompanied with one or multiple smaller trees of different species. Nevertheless, the best deep learning model using the HeliALS dataset is able to predict the species of the largest tree correctly in such segments with an overall (macro-average) accuracy of 75.5\% (64.9\%).

\begin{figure*}[h!]
    \centering
    \includegraphics[width=\linewidth]{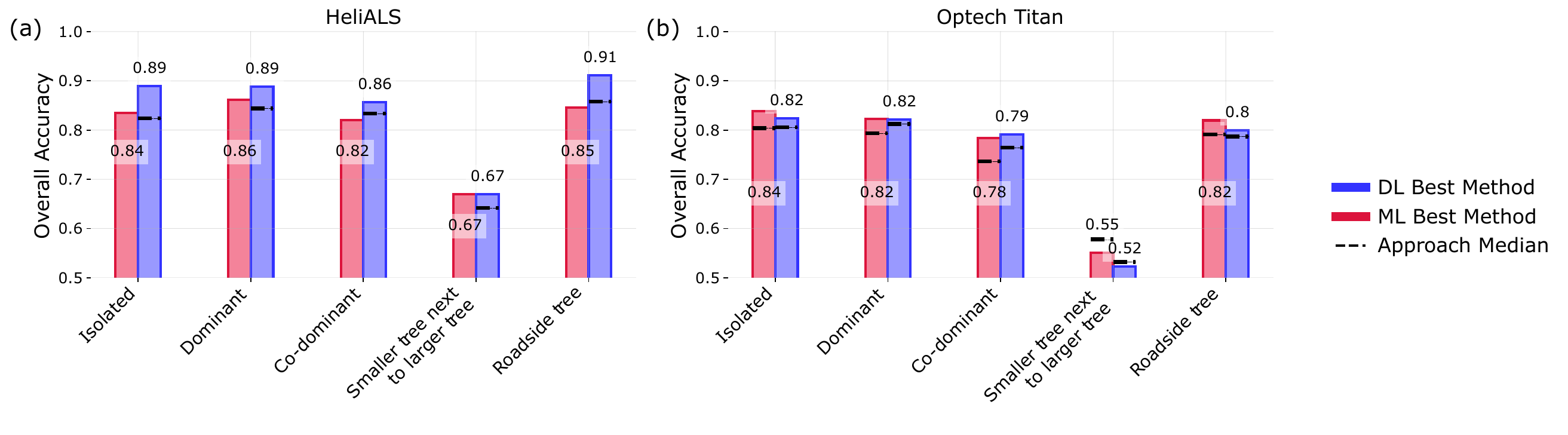}
    \caption{Overall accuracy by crown class category for the best machine learning and deep learning methods for  (a) HeliALS and  (b) Optech Titan datasets. Dashed lines indicate median values of overall accuracy across the methods utilizing the same dataset and approach type (DL/ML). The best method was determined by macro \(F_1\) score. (a) For the HeliALS dataset, the best deep learning method is FGI-PointTransformerWeighted-DL-3D (Macro \(F_1 = 0.73\)) and the best machine learning method is FGI-RF-ML (Macro \(F_1 = 0.64\)). (b) For the Optech Titan dataset, the best deep learning method is FGI-PointTransformer-DL-3D (Macro \(F_1 = 0.58\)) and the best machine learning method is IBL-BalancedRF-ML (Macro \(F_1 = 0.57\)).}
    \label{fig: OA results for tree crown categories}
\end{figure*}

For the HeliALS dataset, we observe that the best DL method consistently outperforms the best ML method across the profile categories as shown in Fig.~\ref{fig: OA results for profile categories}(a). When considering the macro-average accuracy (see~\ref{ap: macro-average recall}), the improvement provided by the DL method becomes more apparent especially for the most challenging category `Many trees of many species`. For this category, the best DL method reaches a macro-average accuracy of 64.9\%, while the corresponding quantity for the best ML method is only 45.5\%. For the Optech Titan dataset, we observe that the best DL and ML methods perform equivalently across the different profile categories as shown in Fig.~\ref{fig: OA results for profile categories}(b), though the DL model provides a small improvement for the category `Many trees of many species`, while the roles are reversed for the category `Tree section`.

When it comes to the crown class categories, we observe that the overall classification accuracy is the highest and essentially equivalent for the classes `Isolated`, `Dominant`, and `Roadside trees` across the different deep learning and machine learning models on both datasets as  shown in Figs.~\ref{fig: OA results for tree crown categories}(a) and (b). For the crown class `Co-dominant`, the overall accuracy is a few percentage points lower compared to the above mentioned crown classes. The classification accuracy is significantly lower for the category `Smaller Tree next to larger tree`. For example, the best DL method on the HeliALS data reaches an overall accuracy of only 67\% for this category, which is over 20 percentage points lower than the overall accuracy of around 90\% for the crown classes `Isolated`, `Dominant`, and `Roadside trees`. These observations are consistent with prior research \citep[e.g.,][]{hakula2023individual}, where suppressed trees, characterized by their small size and proximity to larger neighbors, had an overall accuracy 20 percentage points lower than that of dominant trees.

To the best of our knowledge, the effect of segmentation quality on species classification has not been investigated in detail in prior literature since the segmentation has often been improved manually to ensure near-ideal tree segments, such as in \citet{puliti2025benchmarking}. However, ALS-based tree species estimation across large areas will necessitate automatic segmentation, which will unavoidably lead to imperfect segmentation quality, especially for small or suppressed trees. 
In our study, the segmentation was based on a traditional watershed approach applied to the sparse Optech Titan dataset. In the future, segmentation methods based on deep learning, such as  ForAINet \citep{xiang2024automated}, SegmentAnyTree \citep{wielgosz2024segmentanytree}, and TreeLearn \citep{henrich2024treelearn}, may improve the segmentation quality especially on dense ALS data, where the high point density may contain cues for improved segmentation beyond traditional approaches. Further work should also study whether it is advisable to use the same segmentation method for both the training phase and the prediction phase in an operational setting as opposed to using a high-quality, labour-intensive manual segmentation for the training phase and an automatic segmentation for the prediction phase. According to our hypothesis, it is better to use the same segmentation method throughout the process to ensure that the training and prediction data come from the same distribution.

\subsection{Further Discussion}
\label{sec: further discussion}

Our study compares favorably against recent UAV-based studies employing state-of-the-art laser scanning systems and high-quality passive hyperspectral imaging systems.  For example, \citet{quan2023tree} used a Random Forest classifier with leaf-on data acquired from a Riegl VUX-1UAV laser scanner and a MicroCASI 1920 hyperspectral imager to classify 11 species (2 coniferous, 9 broad-leaved), achieving a macro-average accuracy of 69.2\% on a dataset with 904 trees using 90/10 train-test split.  \citet{zhong2022identification} employed a Support Vector Machine to classify 5 species (1 coniferous, 4 broad-leaved) using data from a RIEGL miniVUX-1UAV and a Resonon Pika L hyperspectral sensor. They achieved an overall accuracy of 89.2\% and a macro-average accuracy of 84.5\% on a dataset with 880 trees using a 60/40 train-test split.  In our study, the best-performing model, FGI-PointTransformer-DL, reaches an overall accuracy of 92.0\% and a macro-average accuracy of 85.1\% across the nine studied species when using the full HeliALS dataset with 6257 segments and 80/20 train-test split, see Fig.~\ref{fig: scaling laws accuracy vs training set size}(a). Thus, our best result surpasses the performance of these recent state-of-the-art studies on UAV-based species classification.

\begin{figure*}[h]
    \centering
    \includegraphics[width=0.9\textwidth]{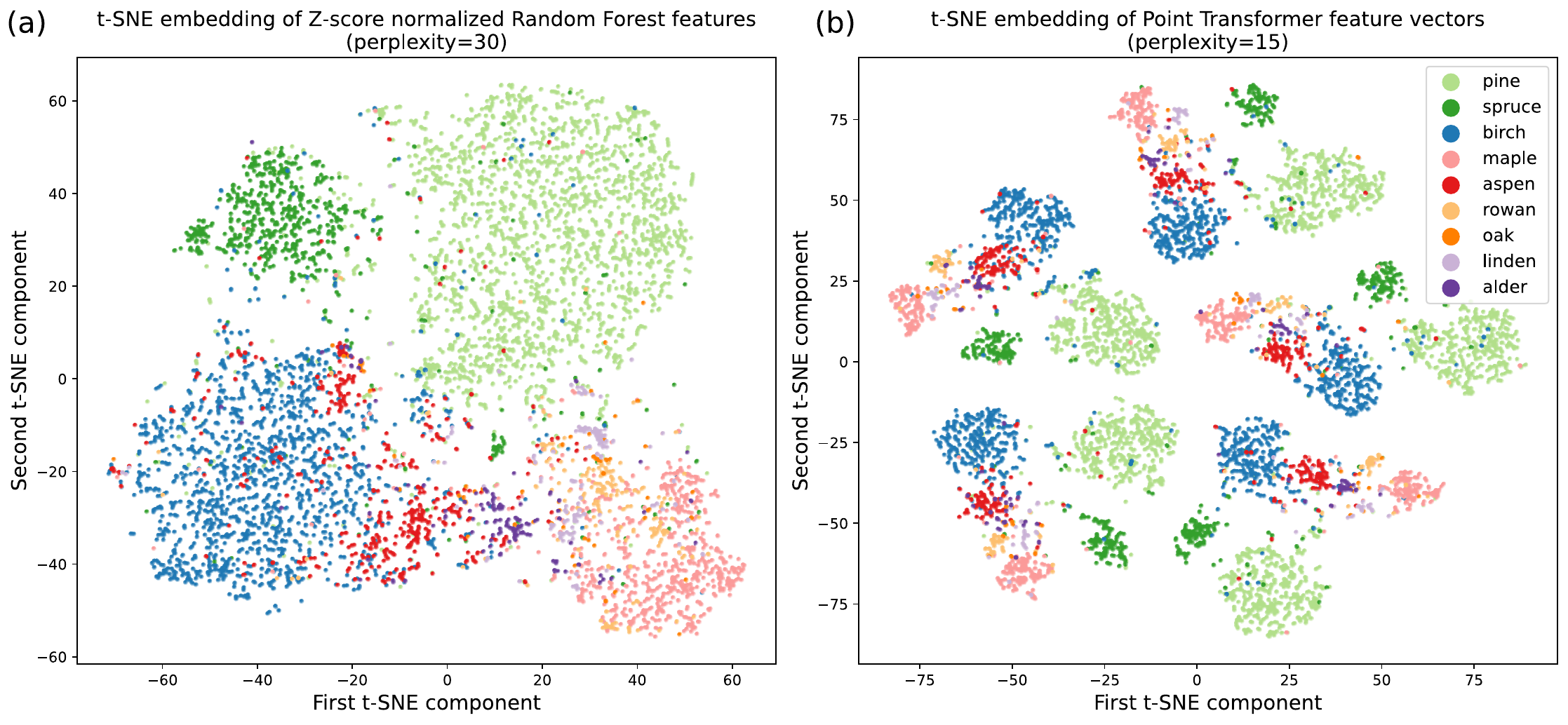}
    \caption{A comparison of t-distributed stochastic neighbor embedding (t-SNE) visualizations for Z-score normalized FGI-RF-ML input features $\mathbf{f}_{RF} \in \mathbb{R}^{30}$ in (a) and FGI-PointTransformer-DL-3D deep features $\mathbf{f}_{PT} \in \mathbb{R}^{512}$ in (b). The data samples were obtained from the full reference dataset, and the colors represent the species in the field reference.}
    \label{fig:tSNEvisualization}
\end{figure*}

To gain further insights into the classification errors discussed in Sec.~\ref{sec: species wise analysis}, we visualize the clustering of different tree species in the feature space in Figs.~\ref{fig:tSNEvisualization}(a) and (b) using input features of FGI-RF-ML and the deep features of FGI-PointTransformer-DL-3D, respectively. To project the high-dimensional features into a two-dimensional plane, we use t-distributed stochastic neighbor embedding (t-SNE) \citep{hinton2002stochastic} with perplexity values ranging from 15 to 30 to have a clear separation of the main clusters. In Fig.~\ref{fig:tSNEvisualization}(a) showing the clusters for the RF features, we observe that pines, spruces, birches and maples mostly form separate clusters in line with the high classification accuracy of these species. Visually, these species  have clearly distinct characteristics, and they are easy to recognize in the field based on their shapes and needles or leaves, see Fig.~\ref{fig: species visualization and distribution}.  However, we note that a small set of spruces is located in the middle of the feature space between pine and alder clusters, separate from other spruces. A closer inspection of individual trees in this cluster showed that these spruces were typically growing on yards, roadsides or other  places in the built-up environment, and they had a narrow shape different from large spruces growing in a forest. 

For the remaining five species, there is significant inter-species overlap in the feature space. Clusters for aspen and alder are located next to each other with a partial overlap, which seems natural due to the similar appearance of these species. Furthermore, aspens also have overlap with birches in the feature space. Rowans, oaks and lindens are overlapping and close to other deciduous species, especially maples. This might be related to similarities in the size and shape of these species. Naturally, these observations are well-aligned with our remarks based on the confusion matrices presented in Sec.~\ref{sec: species wise analysis}. The separation of deciduous species in the feature space and hence their classification accuracy may be affected by the season of data acquisition. For example, \citet{lisiewicz2025comprehensive} observed an increase in the classification accuracy of oak and aspen using ALS data collected during leaf-off season, whereas maple and linden were more accurately classified using leaf-on ALS data. A combination of features derived from leaf-off and leaf-on data has been shown to improve the species classification accuracy in several studies~\citep{kim2009tree, shi2018important, kaminska2021single, lisiewicz2025comprehensive}.

In the feature space of the point transformer model shown in Fig.~\ref{fig:tSNEvisualization}(b), each species is separated into five clusters for a reason that is not well understood. Generally, the separability of the species seems slightly better in the feature space of the point transformer model than in the RF feature space as expected from the higher classification accuracy. However, the relations between nearby species in the feature space of the point transformer model are overall similar to those of the RF model.
Importantly, we believe that feature space visualizations may also be useful for the verification of large reference datasets in future studies on species classification. Namely, reference trees located within a wrong cluster may potentially correspond to a mistake in the reference dataset due to, e.g., a positioning error during field measurements. Thus, such trees may require a further verification in the field to ensure the correctness of their species.

As briefly discussed in Sec.~\ref{sec: benchmark of species classification methods}, DetailView-DL-2D trained on our dataset without any spectral information achieved a comparable classification accuracy as the Point Transformer model augmented with single-channel (VUX) reflectance.
Hence, further studies should investigate incorporating single-channel or multi-channel spectral information in the DetailView model, which holds potential to improve the classification accuracy especially for the minority species. Importantly, retraining of the DetailView model was necessary to achieve competitive performance on the ALS data of the current study. This conclusion seems to be valid even when accounting for the different species composition between our study and that of \citet{puliti2025benchmarking}, which we tested by re-training the classification head while using frozen encoder parameters of the sensor-agnostic model from \citet{puliti2025benchmarking}. Such an approach resulted in an overall (macro-average) accuracy of 78.7\% (56.7\%), which is still significantly lower compared to the overall (macro-average) accuracy of 84.3\% (63.9\%) for the fully retrained model.

Based on our studies with varying point densities in Sec.~\ref{sec: scaling law vs point density}, the influence of single- and multi-channel spectral information is most pronounced at low point densities of around $10~\mathrm{pts}/\mathrm{m}^2$. 
However, we still observe a significant improvement in the classification accuracy when using the HeliALS data ($>1000$ pts/$\mathrm{m}^2$) instead of the Optech Titan data (35 pts/$\mathrm{m}^2$), especially when it comes to the macro-average accuracy that is 14 percentage points higher for the best-performing model of the HeliALS data. Our studies with a varying point density rely on artificial subsampling of the HeliALS point cloud. In practical implementations, the reduced point density is often a byproduct of a higher flight altitude, which may reduce the geometric accuracy of the point cloud due to an increased beam width. Thus, our results in Fig.~\ref{fig: classification error wrt point density} may not perfectly describe a scenario, in which the flight altitude controls the point density. It is also important to note the disparity in the availability of point cloud data with densities similar to HeliALS or Optech Titan systems. The Optech Titan data exhibits a density comparable to that used in municipal mapping applications in Finland, typically ranging from 40 to 60 points/$\mathrm{m}^2$. Conversely, the HeliALS data density aligns more closely with drone-based LiDAR surveys, indicating its suitability for smaller, more localized studies, e.g. for forest holding.

In principle, our results demonstrate precision and recall values suitable for operational mapping of aspen populations at the individual tree level, reaching $F_1=0.71$ for the best performing model on the HeliALS data and $F_1=0.57$ for the best-performing model on the Optech Titan data. However, the proportion of aspen trees was higher on our study area compared to typical boreal forests, and therefore, a lower $F_1$ score may be achieved in operational use for an equivalent training set size. For example, \citet{toivonen2024mapping} reached  an  $F_1$ score of 0.44  for mapping large aspens using ALS (7.2 $\mathrm{pts}/\mathrm{m}^2$) and aerial imagery on sample plots with an aspen proportion of around 6\%, whereas the $F_1$ score reduced to 0.21 on sample plots with an aspen proportion of around 0.1\%. Nevertheless, our results suggest that a higher point density and multispectral information improve the classification accuracy of aspen. 

Importantly, the point density of nation-wide ALS surveys is approaching that of Optech Titan data,  while the ALS surveys of several cities and municipalities already utilize denser point clouds than those collected with Optech Titan in this study. In most of the current nation- and city-level ALS surveys, spectral information is available only at a single wavelength. When it comes to species classification across large areas, such as entire countries, we also expect that the models using single-channel or multi-channel spectral information may not generalize as well to new scanners compared to models relying only on the point cloud geometry. Furthermore, multispectral ALS data may suffer from a reduced point density or a complete lack of points for certain channels across some parts of the mapped area due to, e.g., different fields of view of the scanners, which can be mitigated by a careful planning of the flight path. In the HeliALS dataset, a small part of the tree segments ($\sim 1\%$) suffered from a complete lack of points from miniVUX or VQ840G scanners, which led some methods to develop imputation strategies for assigning the radiometric information to each point.  Thus, the intensity information offers great advantages for improving the accuracy of species classification, but care is needed to ensure data coverage of all channels and consistent intensity calibration across large areas.

When it comes to mapping large areas, the computational cost of species classification becomes also a relevant performance metric even though species classification for forest inventory is typically a post-processing activity without strict runtime requirements. To assess the scalability of the studied classification methods, we compare the run time of model training and inference in \ref{ap: comparison of algorithm running times} for three selected methods including FGI-RF-ML representing a shallow machine learning method, FGI-PointTransformer-DL-3D representing a 3D deep learning method, and DetailView-DL-2D representing a 2D deep learning method. We found that the training time of the RF model was the shortest, but interestingly, the inference time of FGI-PointTransformer-DL-3D was 177 times shorter than that of the RF model and 52 times shorter than that of the DetailView on the HeliALS data. For the RF model, the feature computation dominated the total runtime,  whereas the training and inference times after the feature computation were quick. For more details, see Appendix \ref{ap: comparison of algorithm running times} and Fig.~\ref{fig: runtime comparison}.

Multispectral data also opens up new possibilities for the use of echo classes in species classification. After the first echo, the LiDAR signal is affected by a variety of factors, such as transmission, absorption losses, and multiple scattering, which alter the measured return intensity. The return intensities in such multiple scattering events are affected differently for each wavelength, and the effect may vary by species due to different canopy geometries, thus providing additional information for species classification. For example, features related to intermediate echos were found to contribute to the classification performance of UEF-RF-ML, though features related to the first echoes were the most important ones as they correspond to the purest return intensities. Features related to last echos provided no added value to the classifier.

More studies are also needed to investigate the transferability of deep learning methods to new datasets that may also cover large areas. In Sec.~\ref{sec: scaling law vs training set size}, we observed that the classification error followed a power law scaling as a function of training set size, which, in principle, enables us to estimate the required number of training segments for a desired classification accuracy when using a given model. However, we also noted that the DetailView model performed significantly worse in this study when trained on the FOR-species20K dataset compared to the training set of the current study. Thus, it remains to be verified whether a similar power law scaling of the classification error would be observed if the models trained on the current dataset  were transferred to new areas, though in the same forest zone. The gap between deep learning and machine learning models may narrow down if the faster  convergence of deep learning models is partially due to overfitting to the current dataset.

We also point out that it is challenging to collect accurate reference data of individual trees due to limitations in GNSS signal reception under dense canopies and the difficulty of visually correlating ground observations with a map representation of the area. Furthermore, limited coverage of trees in open tree databases of cities necessitates considerable amount of field work and verification in detailed studies. Therefore, the crowdsourcing tool introduced in Sec.~\ref{sec: reference data} was developed to this need. It allowed flexible collection of reference data by a large group of people. Improvements were made to the tool based on the first experiences in the field work, making the reference data collection  easier and more reliable as the study progressed. As a result of these improvements, the tool is easy to use. Nevertheless, the persons collecting the reference data still need some skills and practice, e.g. to recognise the tree species and to understand the correspondence between one’s position in the terrain and the map in complicated environment with several trees around and some uncertainty in GNSS positioning. Furthermore, one must understand the contents and nature of spatial datasets visible in the tool including the height of trees, segments that can include several trees or part of a tree, and possible differences between the data and the current situation due to cut and newly planted trees. Due to challenges related to the reference data collection, a separate verification phase was implemented to improve the quality of the reference data.

\section{Conclusion}
\label{sec: conclusion}

This research represents a thorough benchmark of machine and deep learning methodologies for tree species classification, leveraging multispectral ALS data. By employing both high-density HeliALS data ($>1000$ $\mathrm{pts}/\mathrm{m}^2$) and lower-density Optech Titan data (35 $\mathrm{pts}/\mathrm{m}^2$), we have provided a comprehensive comparison of various algorithmic approaches. For the benchmark study, we established a training set of 1065 tree segments and a test set of 5261 segments representing nine species in southern Finland. In our benchmark, point-based deep learning methods outperformed both image-based deep learning  and  machine learning methods on the high-density HeliALS data, while the best point-based deep learning methods performed equivalently with machine learning methods on the lower-density Optech Titan data. On the HeliALS data, a Point Transformer model topped the list with an overall (macro-average) accuracy of 87.9\% (74.5\%). On the Optech Titan dataset, a random forest classifier reached the highest overall (macro-average) accuracy of 79.9\% (57.6\%), though three other methods were within error bars in terms of the overall accuracy.

Additionally, we observed a remarkable improvement in the classification accuracy using single-channel and multi-channel intensity features. On the HeliALS data, the overall (macro-average) classification error of the Point Transformer model was reduced by 48\% (32\%) when incorporating single-channel spectral information compared to using no spectral information. By incorporating  multispectral information of three channels, the overall (macro-average) classification error was further reduced by 14\% (17\%). As explained in Sec.~\ref{sec: scaling law vs point density}, the multi-channel spectral features appear to have the largest impact at low point densities ($\sim 10$ $\mathrm{pts}/\mathrm{m}^2$) according to our studies with varying point densities  obtained by subsampling the HeliALS data. At higher point densities, it seems that deep learning approaches can effectively leverage the richer structural information, thus partially compensating for a lack of intensity data.
With high-density multispectral data, deep learning approaches also substantially improve the macro-average classification accuracy, i.e. the detection of more rare species. For example, the macro-average accuracy of the Point Transformer model is  13 percentage points higher than that of a random forest classifier on the HeliALS dataset used for the benchmarking competition. 

Furthermore, we show that the accuracy improvement provided by deep learning methods increases with training set size as discussed in Sec.~\ref{sec: scaling law vs training set size}. We observe that the classification error obeys a power law $\varepsilon(m)=Am^{-\alpha}$ with respect to the training set size both for a point transformer model and for  a random forest classifier. Importantly, the scaling exponent $\alpha$ is approximately twice higher for the deep learning model leading to a drastically faster convergence of the classification error. Based on our analysis, the studied deep learning method outperforms the random forest classifier for training set sizes exceeding a few hundred segments on both datasets. Using the fitted scaling laws on the HeliALS data, we predict that a macro-average accuracy of 90\% would require 14 000 training segments with the point transformer model and  $4.9 \times 10^6$ training segments with the random forest classifier, thus highlighting the benefit of deep learning approaches on large datasets.
The development of the crowdsourced field reference dataset, along with the innovative browser-based crowdsourcing tool, represent a significant contribution to the scientific community aiding the collection of large datasets required to scale up these methods to larger forest areas. These resources will facilitate future research and development in this domain. 

The ability to accurately identify tree species at the individual tree level is essential for sustainable municipality and multifunctional forest management practices. Future research should focus on further refining deep learning architectures for multispectral ALS data, and addressing the challenges associated with identifying rare species. More detailed studies are also needed on the effects of different preprocessing steps and data augmentation methods on the performance of deep learning methods. Integration of tree species classification to direct measurements of stem curve and volume using a high-resolution airborne laser scanning system, as depicted in \citep{hyyppa2022direct}, would be further step in developing a fully automatic reference collection technique. Additionally, investigating the transferability of these models across different geographic regions and sensor platforms is crucial for  broader adoption. From the practical point of view, it is also important to pay attention to datasets that are available nationwide or over large areas. 

\section*{Data availability}

We have made available the Espoonlahti dataset for tree species classification in Zenodo: \url{https://doi.org/10.5281/zenodo.17077255}. The dataset includes both HeliALS and Optech Titan point clouds for the studied tree segments together with ground-truth species labels. The published Zenodo dataset will be described in more detail in a forthcoming data manuscript. 

\section*{Acknowledgements}

We gratefully acknowledge the Research Council of Finland (RCF), CHIST-ERA, the NextGenerationEU instrument and Finnish Government (VN) funding for the following projects ``Mapping of forest health, species and forest fire risks using Novel ICT Data and Approaches'' (RCF 344755, CHIST-ERA-19-CES-001, main project for this study until March 2024), ``Capturing structural and functional diversity of trees and tree communities for supporting sustainable use of forests'' (RCF 348644, NextGenerationEU), ``Forest-Human-Machine Interplay -- Building Resilience, Redefining Value Networks and Enabling Meaningful Experiences'' (RCF 359175), ``Collecting accurate individual tree information for harvester operation decision making'' (RCF 359554), “High-performance computing allowing high-accuracy country-level individual tree carbon sink and biodiversity mapping” (RCF 359203), “Measuring Spatiotemporal Changes in Forest Ecosystem” (decision number 346382)'', ``Digital technologies, risk management solutions and tools for mitigating forest disturbances'' (RCF 353264, NextGenerationEU), and the Ministry of Agriculture and Forestry grant ``Future Forest Information System at Individual Tree Level'' (VN/3482/2021).
M. Myllymäki was funded by the European Union – NextGenerationEU in RCF project "Artificial intelligence, spatial statistics and Earth observation for digital twinning of forest diversity" (RCF 348154) under the Research Council of Finland’s flagship ecosystem for Forest–Human–Machine Interplay—Building Resilience, Redefining Value Networks and Enabling Meaningful Experiences (UNITE) (357909). J. Kostensalo was funded by "Digital Twin of Boreal Forest Structure with Remote Sensing and Forest Inventories" (RCF 361209).
S. Puliti and M. Astrup conducted the work as part of the Center for Research-based Innovation SmartForest: Bringing Industry 4.0 to the Norwegian forest sector (NFR SFI project no. 309671, smartforest.no).

\section*{Author contributions}

The work towards the paper represents 10+ person-year's work, Therefore, we like to acknowledge the joint-first-authorship of J. Taher, E. Hyyppä, M. Hyyppä, K. Salolahti, X. Yu and L. Matikainen. J. Hyyppä acted as a senior supervisor, including project leadership, supervision of the benchmarking, release of the benchmarking study, main contact to partner organisations and acquisition of funding, supported by E. Hyyppä in the manuscript writing and analysis phase. The first authors and the senior supervisor all contributed to writing of the paper. The detailed roles of authors are the following. J. Taher, supported by S. Thurachen, developed the FGI deep learning methods for HeliALS and performed all analysis and scaling law tests related to these techniques. E. Hyyppä planned scaling law tests and the content of the manuscript, and lead the analysis work and the write-up of the paper. M. Hyyppä developed the browser-based crowdsourcing application. M. Hyyppä and K. Salolahti jointly processed and analysed the benchmarking results, classified data into the different categories and prepared most of the result figures. X. Yu performed the segmentation with the Optech data, provided Optech data ready for partners, developed the FGI machine learning method with M. Lehtomäki for both datasets and performed scaling law analysis related to this method. 
J. Hyyppä, L. Matikainen and X. Yu planned the international benchmarking. L. Matikainen and P. Litkey preprocessed the Optech Titan data (steps preceding classification and segmentation) and provided ideas for developing the crowdsourcing tool and collecting the reference data. L. Matikainen had an active role in quality checking of the crowdsourced data and participated in the analysis of the results. A. Kukko and H. Kaartinen made the data collection using HeliALS and preprocessed the data with help of J. Taher and M. Lehtomäki. All FGI persons collected field reference using the crowdsourcing application, supported by University of Helsinki partners. Other partners provided methods and their outcomes for the study.

{
\footnotesize
\bibliography{references}
}

\clearpage
\normalsize
\appendix

\counterwithin{figure}{section}
\counterwithin{table}{section}
\renewcommand\thefigure{\thesection.\arabic{figure}}
\renewcommand\thetable{\thesection.\arabic{table}}

\begin{appendices}

\section{Distribution of mean intensity for the studied tree species }
\label{ap: distribution of mean intensity for different species}

\begin{figure*}[h!]
    \centering
    \includegraphics[width=0.7\linewidth]{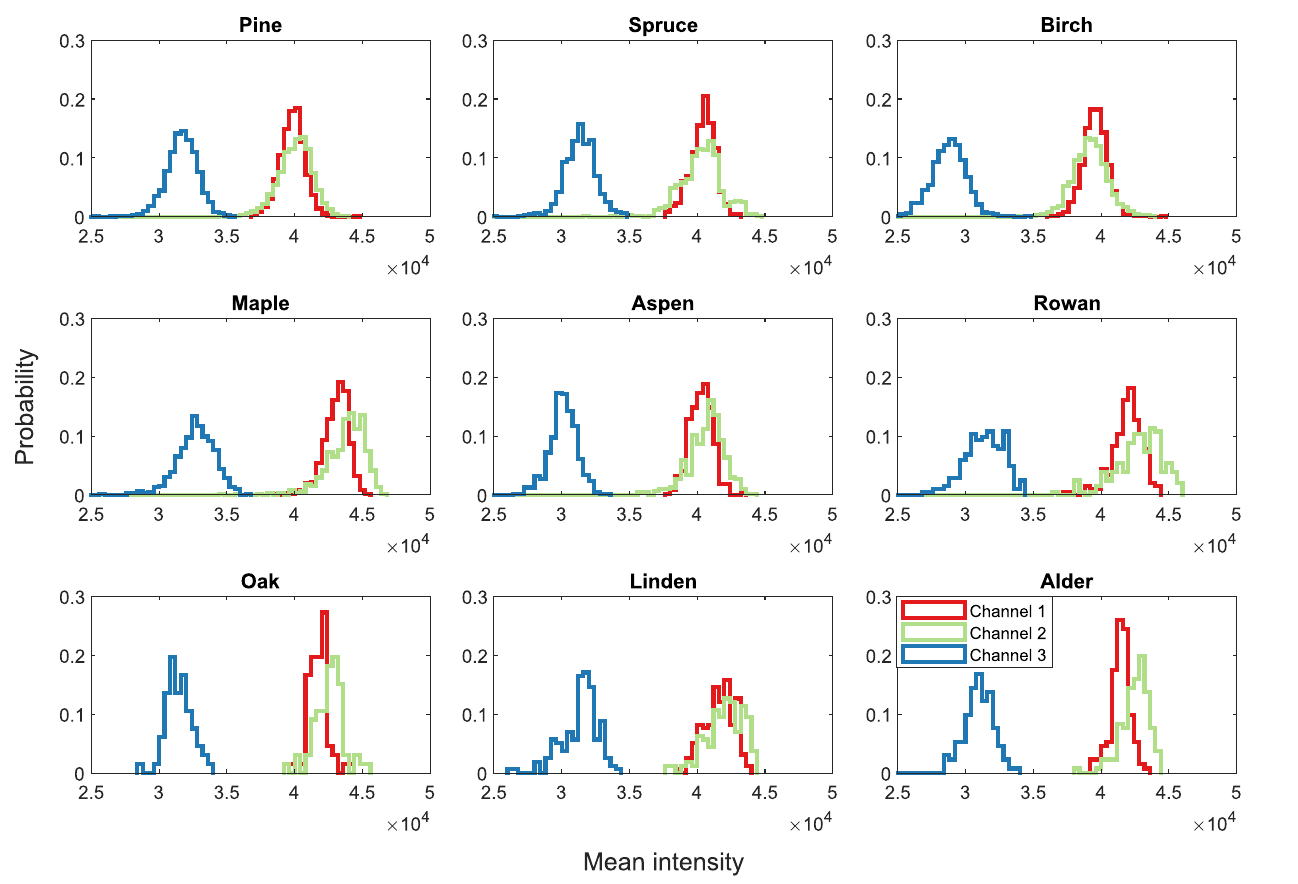}
    \caption{Distribution of mean intensity for the three channels of the HeliALS system for the studied nine species across all segments of our dataset. Here, blue corresponds to the  VQ-840-G scanner, green to the miniVUX-1DL scanner, and red to the VUX-1HA scanner.}
    \label{fig: intensity distribution helials}
\end{figure*}

\begin{figure*}[h!]
    \centering
    \includegraphics[width=0.7\linewidth]{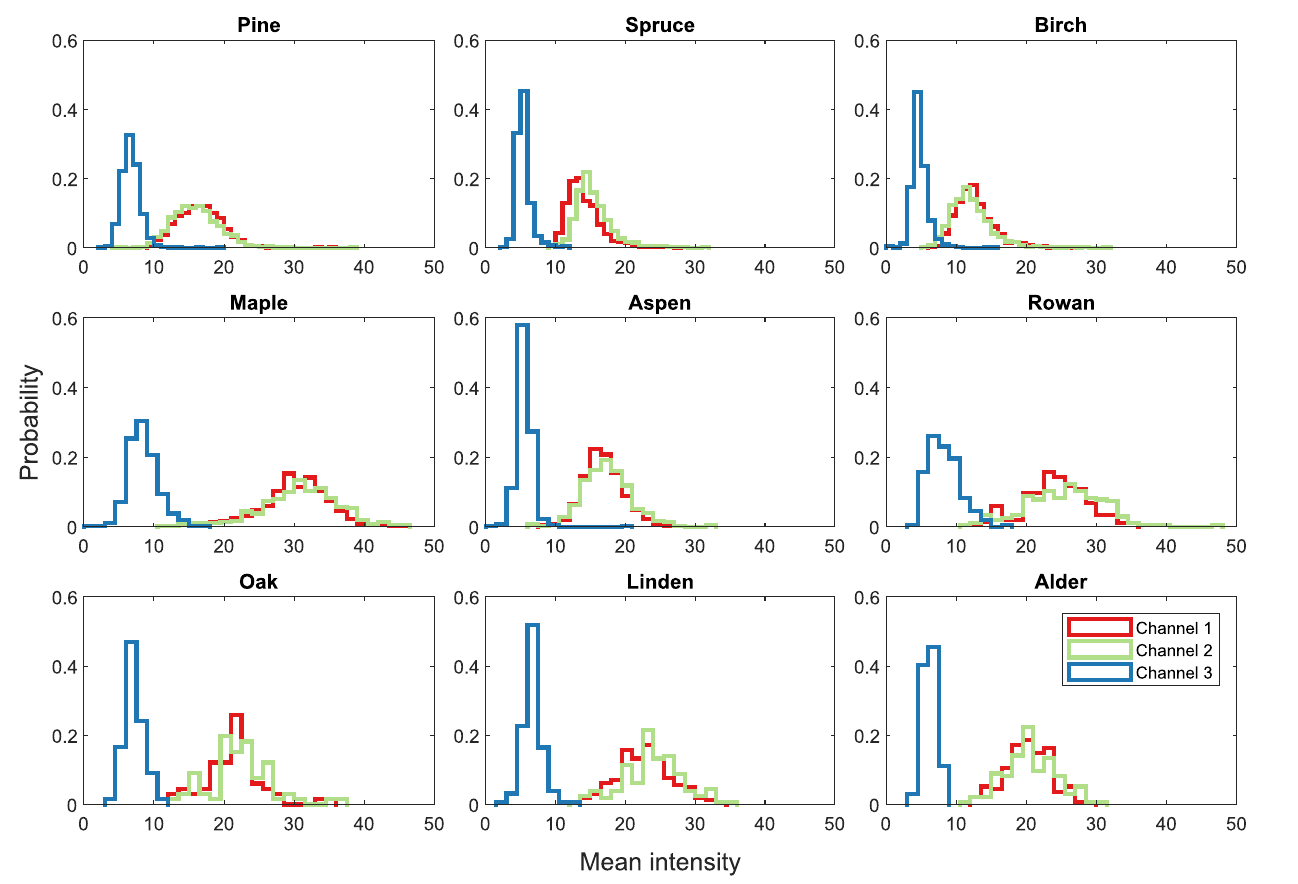}
    \caption{Distribution of mean intensity for the three channels of the Optech Titan system for the studied nine species across all segments of our dataset. Here, blue corresponds to $\lambda=532$ nm, green to $\lambda=1064$ nm, and red to $\lambda=1550$ nm.}
    \label{fig: intensity distribution titan}
\end{figure*}

Multispectral data provides a benefit for tree species classification since different tree species have a different reflectance response as a function of wavelength \citep{hovi2017spectral}. Figures~\ref{fig: intensity distribution helials} and \ref{fig: intensity distribution titan} illustrate the species-wise distribution of mean intensity for the three channels of the HeliALS system and the Optech system, respectively. For example, Fig.~\ref{fig: intensity distribution helials} shows that the mean intensity of the 3rd channel ($\lambda=532$ nm) of the HeliALS data differs significantly between pine and birch, while the difference of mean intensities is smaller for these two species at the  other  two wavelengths. Overall, the variations in the intensity distributions between different species showcase the advantage of multispectral ALS for tree species classification. 

\section{Comparison of algorithm running times}
\label{ap: comparison of algorithm running times}

The computational cost of the FGI-RF-ML, the FGI-PointTransformer-DL-3D and the DetailView-DL-2D methods was assessed to determine their scalability to larger datasets. The comparison was performed separately for model training phases and inference phases. The training set contained 1065 samples, while the test set comprised 5261 samples. In addition, the running times of the FGI-RF-ML and the FGI-PointTransformer-DL-3D were reported both for the Optech Titan dataset and for the HeliALS dataset. The hardware specifications of the experiments are listed in Table~\ref{tab: hardware specification for runtimes}. The FGI-RF-ML method was run on computer 1, and the FGI-PointTransformer-DL-3D and the DetailView-DL-2D methods on computer 2.

\begin{table}[ht!]
\large
\centering
\caption{Hardware specifications of the two computers used for runtime measurements.}
\label{tab: hardware specification for runtimes}
\resizebox{0.5\textwidth}{!}{\begin{tabular}{lcc}
    \toprule
    Specification & Computer~1  & Computer~2  \\
    \hline
    CPU & \makecell[c]{Intel Core i7-9850H \\ (6 cores, 2.6\,GHz)} 
    & \makecell[c]{Intel Xeon(R) Gold 6244 \\ (16 cores, 3.6\,GHz)} \\
    RAM & 32\,GB & 512\,GB \\
    GPU & not utilized & NVIDIA RTX A6000 (48\,GB) \\
    Operating System & Windows 10 & Ubuntu 22.04.5 LTS \\
    \bottomrule
\end{tabular}}
\end{table}

\begin{figure*}[ht]
    \centering
    \includegraphics[width=0.8\linewidth]{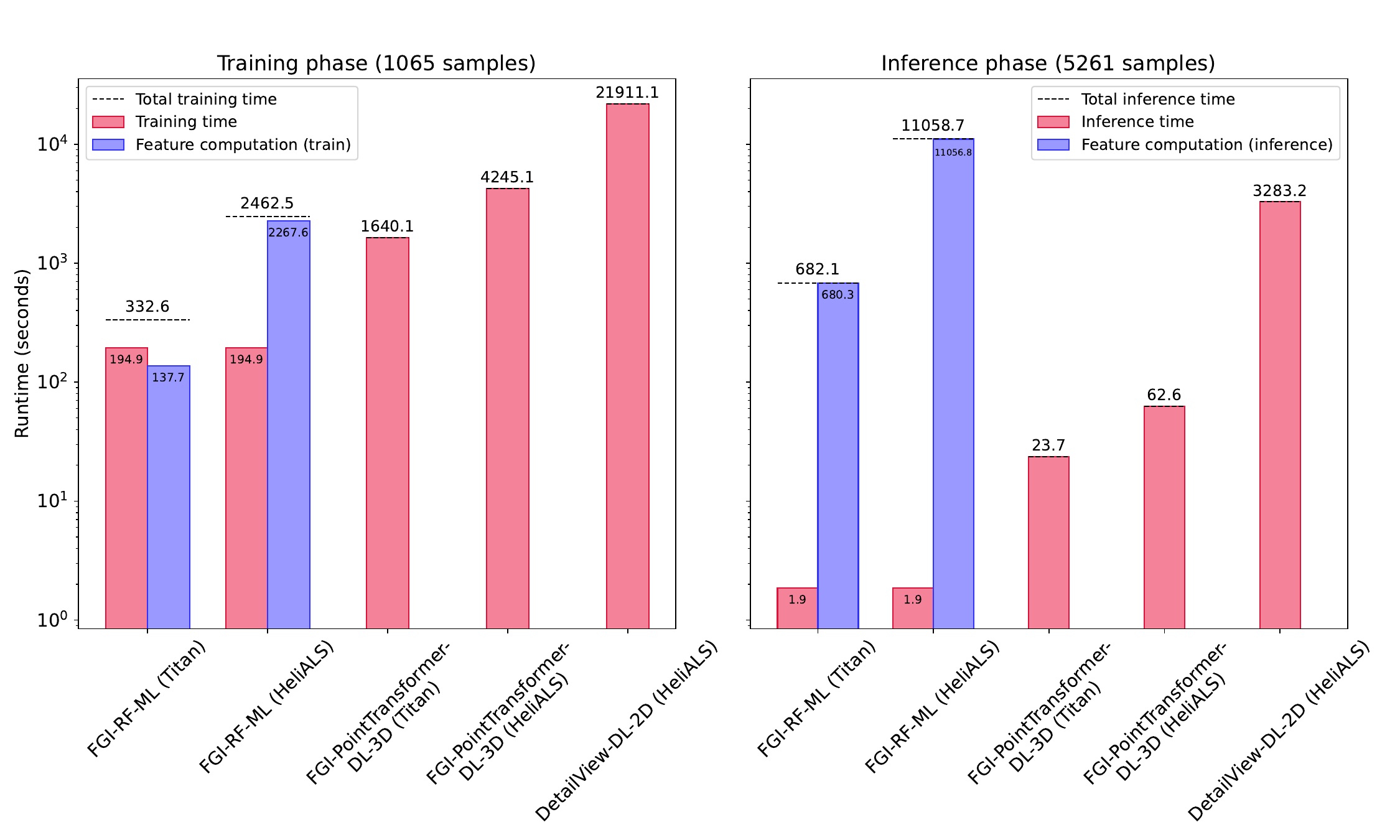}
    \caption{Comparison of running times between three tree species classification methods in different datasets. The training set size was 1065 tree segments and the test set size was 5261 tree segments. The dashed horizontal line represents the total running time including feature computation and classification. For deep learning based methods, the feature computation is inherent to the computations and has not been separately reported.}
    \label{fig: runtime comparison}
\end{figure*}

The results of the running time comparison have been visualized in Fig.~\ref{fig: runtime comparison}. In the training phase, the computational cost of the FGI-RF-ML and the FGI-PointTransformer-DL-3D methods are quite similar. However, the DetailView-DL-2D method has a substantionally longer training time than either of the previous methods. The FGI-RF-ML method is the fastest to train and can be trained in less than six minutes on our Optech Titan dataset.

During inference, the FGI-PointTransformer-DL-3D method performs the best in terms of running time. When using the HeliALS dataset, the FGI-RF-ML method is approximately 177 times slower than the FGI-PointTransformer-DL-3D method. Similarly, the inference speed of the DetailView-DL-2D method is approximately 52 times slower when compared to the FGI-PointTransformer-DL-3D method. The FGI-RF-ML method uses a majority of the computation time for the calculation of the point cloud features (99,98\% of the total time for the HeliALS dataset and 99,74\% of the total time for the Optech Titan dataset), whereas the prediction process of the RF classifier is relatively fast. It can be hypothesized that a more efficient implementation of the feature calculation process could substantially increase the inference speed of FGI-RF-ML.

The implementation details of the inference process influence the results to a certain degree. For example, the FGI-PointTransformer-DL-3D was trained separately for five times and the five model instances were used to infer the tree species on the test dataset using a majority voting scheme. Likewise, the DetailView-DL-2D method performed the inference for 50 randomly augmented versions of each test set sample obtaining the prediction by selecting the species with the maximum cumulative prediction probability. When these factors are taken into consideration, the FGI-PointTransformer-DL-3D method is approximately five times faster than the DetailView-DL-2D method.

\section{Macro-average accuracy by profile category and tree crown category}
\label{ap: macro-average recall}

\begin{figure*}[htb]
    \centering
    \includegraphics[width=\linewidth]{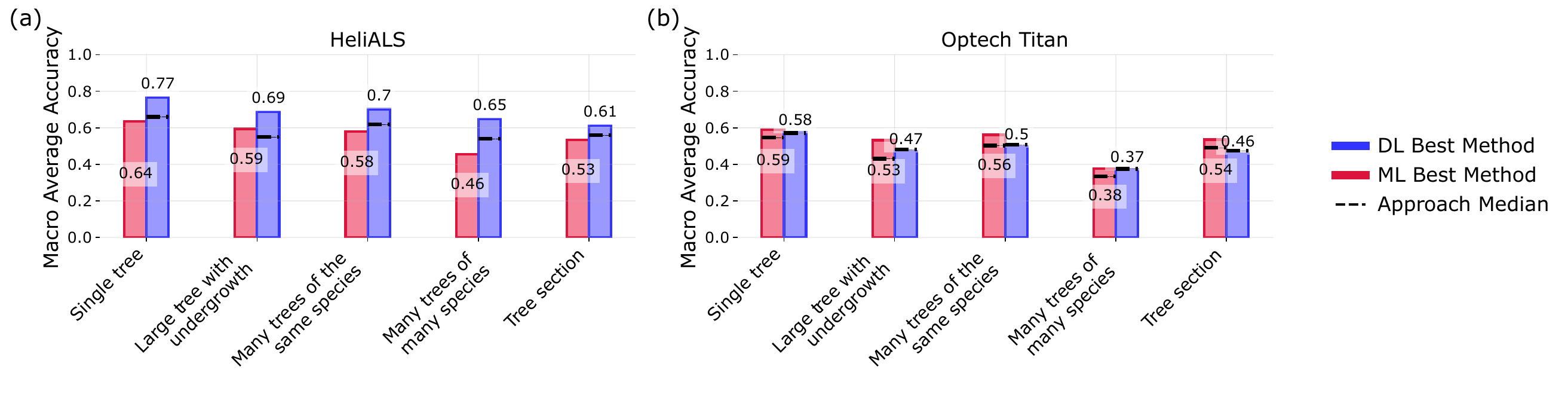}
    \caption{Macro-average accuracy by profile category for the best machine learning and deep learning methods for HeliALS (Panel A) and Optech Titan (Panel B) datasets. Median values of macro-average accuracy were calculated across the methods that utilize the same dataset and approach type (DL/ML). The best method was determined by macro \(F_1\) score. (a) For the HeliALS dataset, the best deep learning method is FGI-PointTransformerWeighted-DL-3D (Macro \(F_1 = 0.73\)) and the best machine learning method is FGI-RF-ML (Macro \(F_1 = 0.64\)). (b) For the Optech Titan dataset, the best deep learning method is FGI-PointTransformer-DL-3D (Macro \(F_1 = 0.58\)) and the best machine learning method is IBL-BalancedRF-ML (Macro \(F_1 = 0.57\)).}
    \label{fig: recall_profile_categories}
\end{figure*}

\begin{figure*}[htb]
    \centering
    \includegraphics[width=\linewidth]{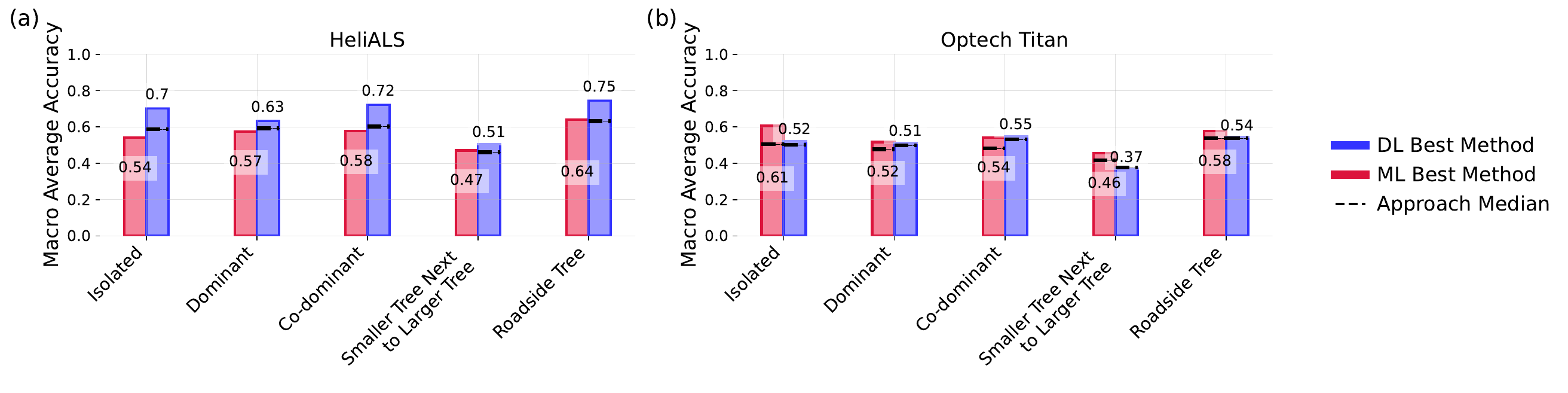}
    \caption{Macro-average accuracy by crown class category for the best machine learning and deep learning methods for  HeliALS (Panel A) and Optech Titan (Panel B) datasets. Median values of macro-average accuracy were calculated across the methods that utilize the same dataset and approach type (DL/ML). The best method was determined by macro \(F_1\) score. (a) For the HeliALS dataset, the best deep learning method is FGI-PointTransformerWeighted-DL-3D (Macro \(F_1 = 0.73\)) and the best machine learning method is FGI-RF-ML (Macro \(F_1 = 0.64\)). (b) For the Optech Titan dataset, the best deep learning method is FGI-PointTransformer-DL-3D (Macro \(F_1 = 0.58\)) and the best machine learning method is IBL-BalancedRF-ML (Macro \(F_1 = 0.57\)).}
    \label{fig: recall_isolation_categories}
\end{figure*}

Figures~\ref{fig: recall_profile_categories} and~\ref{fig: recall_isolation_categories} present the macro-average accuracies by profile category and by crown class category, respectively, for the best methods according to the macro $F_1$ score. The figures have been made for comparison against Fig.~\ref{fig: OA results for profile categories} and Fig.~\ref{fig: OA results for tree crown categories}, which present the overall accuracies for the same categories. Notice the different scaling of the y-axis compared to the Fig.~\ref{fig: OA results for profile categories} and Fig.~\ref{fig: OA results for tree crown categories}.

\pagebreak

\section{Confusion matrices for all methods and confusion matrices for best methods with raw counts}
\label{ap: confusion matrices}

\begin{figure*}[h!]
    \centering
    \includegraphics[width=\linewidth]{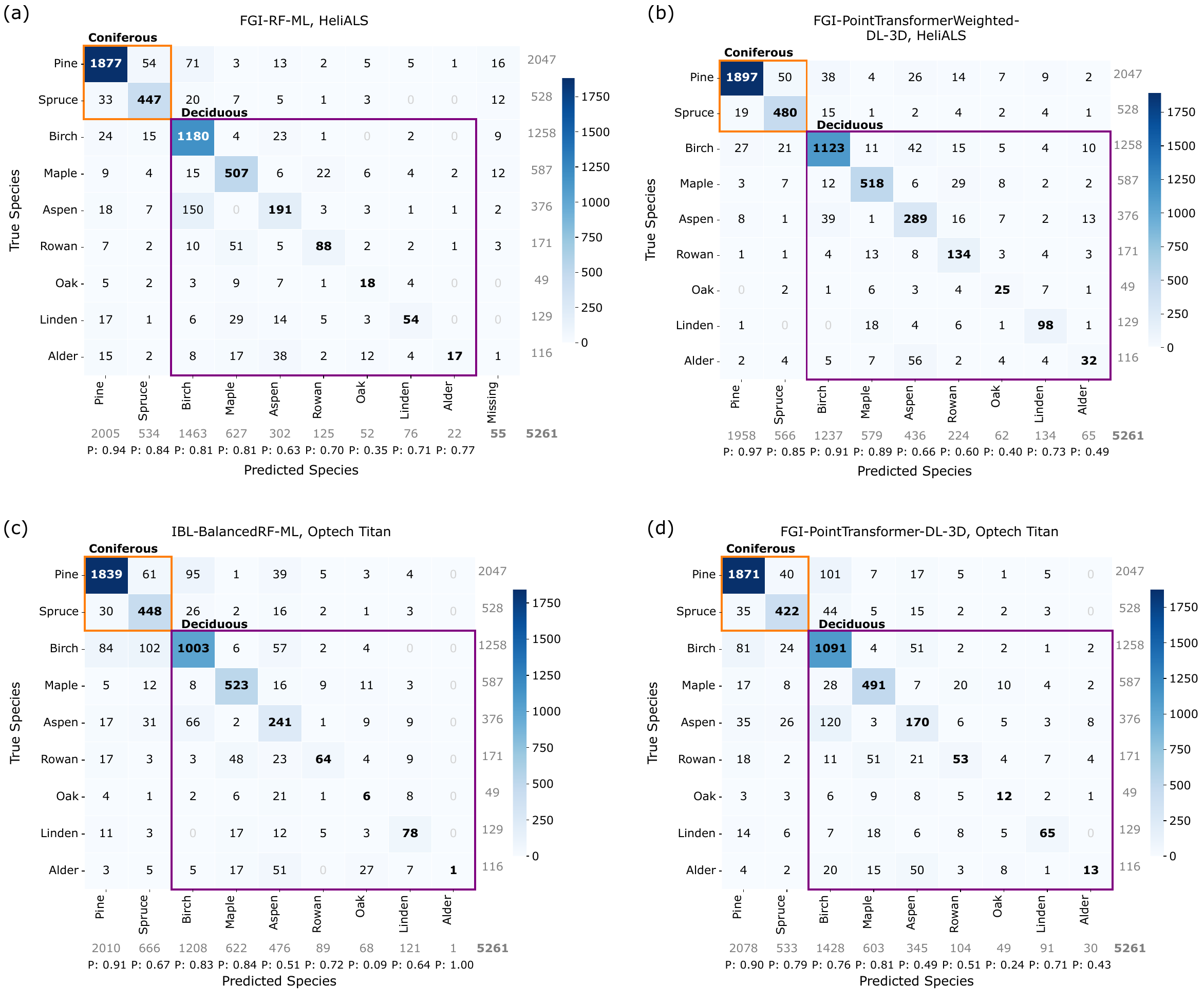}
    \caption{(a) Unnormalized confusion matrix of FGI-RF-ML obtaining the highest macro $F_1$ score (0.64) among the machine learning models using HeliALS data. FGI-RF-ML produced 55 missing predictions. (b) Unnormalized confusion matrix of FGI-PointTransformerWeighted-DL-3D obtaining the highest macro $F_1$ score (0.73) among the deep learning models using HeliALS data. (c) Unnormalized confusion matrix of IBL-BalancedRF-ML obtaining the highest macro $F_1$ score (0.57) among the machine learning models using Optech Titan data. (d) Unnormalized confusion matrix of FGI-PointTransformer-DL-3D obtaining the highest macro $F_1$ score (0.58) among the deep learning models using Optech Titan data. All confusion matrices show raw counts. Precisions are displayed below the matrix. Row and column sums are shown to the right and bottom of the matrix in gray.}
    \label{fig: confusion_matrices_raw}
\end{figure*}

Figure~\ref{fig: confusion_matrices_raw} shows the raw confusion matrices of counts for the best performing machine learning and deep learning methods on the HeliALS and Optech Titan datasets. The corresponding normalized confusion matrices are shown in Fig.~\ref{fig:confusion-matrices}.

Tables~\ref{tab:confusion_matrix_FBK-PointNet++-DL-3D,_HeliALS}--\ref{tab:confusion_matrix_TUW-PointNet++-DL-3D,_HeliALS} show the row-normalized confusion matrices of the deep learning methods using the HeliALS dataset, with the bolded diagonal entries representing the per-class recall. Furthermore, Tables~\ref{tab:confusion_matrix_FBK-PointNet++-DL-3D,_Optech_Titan}--\ref{tab:confusion_matrix_NTNU-ConvNeXt-T-DL-2D,_Optech_Titan} show the normalized confusion matrices of the deep learning methods using the Optech Titan dataset, and Tables~\ref{tab:confusion_matrix_Aalto-RF-ML,_Optech_Titan}--\ref{tab:confusion_matrix_UPV-GB-ML,_Optech_Titan} show the normalized confusion matrices of the machine learning methods using the Optech Titan dataset.

\begin{table}[htbp]
\centering
\caption{Confusion matrix: FBK-PointNet++-DL-3D, HeliALS}
\label{tab:confusion_matrix_FBK-PointNet++-DL-3D,_HeliALS}
\footnotesize
\resizebox{\columnwidth}{!}{%
\begin{tabular}{lcccccccccc}
\toprule
 & \multicolumn{10}{c}{\textbf{Predicted species}} \\
\textbf{Reference} & Pine & Spruce & Birch & Maple & Aspen & Rowan & Oak & Linden & Alder & Count \\
\midrule
Pine & \textbf{0.70} & 0.06 & 0.06 & 0.13 & 0.03 & 0.00 & 0.01 & 0.00 & 0.00 & 2047 \\
Spruce & 0.07 & \textbf{0.70} & 0.09 & 0.09 & 0.03 & 0.00 & 0.01 & 0.00 & 0.00 & 528 \\
Birch & 0.03 & 0.03 & \textbf{0.79} & 0.09 & 0.05 & 0.00 & 0.00 & 0.00 & 0.00 & 1258 \\
Maple & 0.10 & 0.02 & 0.09 & \textbf{0.50} & 0.14 & 0.08 & 0.01 & 0.01 & 0.04 & 587 \\
Aspen & 0.08 & 0.03 & 0.28 & 0.14 & \textbf{0.42} & 0.01 & 0.01 & 0.01 & 0.02 & 376 \\
Rowan & 0.08 & 0.00 & 0.09 & 0.37 & 0.16 & \textbf{0.22} & 0.01 & 0.04 & 0.03 & 171 \\
Oak & 0.02 & 0.02 & 0.31 & 0.29 & 0.20 & 0.08 & \textbf{0.00} & 0.02 & 0.06 & 49 \\
Linden & 0.13 & 0.01 & 0.02 & 0.25 & 0.20 & 0.05 & 0.00 & \textbf{0.33} & 0.01 & 129 \\
Alder & 0.08 & 0.03 & 0.30 & 0.08 & 0.35 & 0.00 & 0.02 & 0.03 & \textbf{0.11} & 116 \\
Count & 1635 & 554 & 1397 & 906 & 488 & 114 & 29 & 76 & 62 & 5261 \\
\bottomrule
\end{tabular}
}
\end{table}

\begin{table}[htbp]
\centering
\caption{Confusion matrix: FGI-Point2Vec-DL-3D, HeliALS}
\label{tab:confusion_matrix_FGI-Point2Vec-DL-3D,_HeliALS}
\footnotesize
\resizebox{\columnwidth}{!}{%
\begin{tabular}{lcccccccccc}
\toprule
 & \multicolumn{10}{c}{\textbf{Predicted species}} \\
\textbf{Reference} & Pine & Spruce & Birch & Maple & Aspen & Rowan & Oak & Linden & Alder & Count \\
\midrule
Pine & \textbf{0.92} & 0.02 & 0.02 & 0.00 & 0.01 & 0.01 & 0.01 & 0.00 & 0.01 & 2047 \\
Spruce & 0.05 & \textbf{0.90} & 0.03 & 0.01 & 0.01 & 0.00 & 0.01 & 0.00 & 0.00 & 528 \\
Birch & 0.03 & 0.02 & \textbf{0.81} & 0.02 & 0.07 & 0.02 & 0.01 & 0.01 & 0.02 & 1258 \\
Maple & 0.01 & 0.01 & 0.01 & \textbf{0.85} & 0.02 & 0.03 & 0.04 & 0.02 & 0.01 & 587 \\
Aspen & 0.02 & 0.01 & 0.14 & 0.01 & \textbf{0.61} & 0.05 & 0.03 & 0.04 & 0.09 & 376 \\
Rowan & 0.02 & 0.01 & 0.01 & 0.05 & 0.08 & \textbf{0.69} & 0.04 & 0.01 & 0.09 & 171 \\
Oak & 0.02 & 0.00 & 0.00 & 0.10 & 0.02 & 0.08 & \textbf{0.71} & 0.02 & 0.04 & 49 \\
Linden & 0.00 & 0.00 & 0.01 & 0.08 & 0.04 & 0.11 & 0.05 & \textbf{0.66} & 0.05 & 129 \\
Alder & 0.02 & 0.01 & 0.03 & 0.08 & 0.34 & 0.04 & 0.22 & 0.00 & \textbf{0.25} & 116 \\
Count & 1961 & 541 & 1131 & 573 & 415 & 217 & 150 & 136 & 137 & 5261 \\
\bottomrule
\end{tabular}
}
\end{table}

\begin{table}[htbp]
\centering
\caption{Confusion matrix: FGI-DGCNN-DL-3D, HeliALS}
\label{tab:confusion_matrix_FGI-DGCNN-DL-3D,_HeliALS}
\footnotesize
\resizebox{\columnwidth}{!}{%
\begin{tabular}{lcccccccccc}
\toprule
 & \multicolumn{10}{c}{\textbf{Predicted species}} \\
\textbf{Reference} & Pine & Spruce & Birch & Maple & Aspen & Rowan & Oak & Linden & Alder & Count \\
\midrule
Pine & \textbf{0.94} & 0.02 & 0.02 & 0.00 & 0.01 & 0.00 & 0.00 & 0.00 & 0.00 & 2047 \\
Spruce & 0.05 & \textbf{0.89} & 0.04 & 0.01 & 0.00 & 0.01 & 0.00 & 0.01 & 0.00 & 528 \\
Birch & 0.02 & 0.01 & \textbf{0.93} & 0.01 & 0.02 & 0.00 & 0.00 & 0.00 & 0.00 & 1258 \\
Maple & 0.01 & 0.01 & 0.02 & \textbf{0.90} & 0.01 & 0.03 & 0.01 & 0.01 & 0.00 & 587 \\
Aspen & 0.03 & 0.01 & 0.20 & 0.03 & \textbf{0.66} & 0.02 & 0.01 & 0.03 & 0.01 & 376 \\
Rowan & 0.02 & 0.01 & 0.04 & 0.18 & 0.05 & \textbf{0.63} & 0.02 & 0.05 & 0.00 & 171 \\
Oak & 0.04 & 0.06 & 0.02 & 0.29 & 0.08 & 0.08 & \textbf{0.33} & 0.10 & 0.00 & 49 \\
Linden & 0.02 & 0.00 & 0.03 & 0.20 & 0.05 & 0.01 & 0.02 & \textbf{0.67} & 0.00 & 129 \\
Alder & 0.02 & 0.02 & 0.16 & 0.20 & 0.36 & 0.03 & 0.08 & 0.04 & \textbf{0.09} & 116 \\
Count & 2014 & 524 & 1364 & 650 & 358 & 155 & 47 & 132 & 17 & 5261 \\
\bottomrule
\end{tabular}
}
\end{table}

\begin{table}[htbp]
\centering
\caption{Confusion matrix: FGI-PointNet-DL-3D, HeliALS}
\label{tab:confusion_matrix_FGI-PointNet-DL-3D,_HeliALS}
\footnotesize
\resizebox{\columnwidth}{!}{%
\begin{tabular}{lcccccccccc}
\toprule
 & \multicolumn{10}{c}{\textbf{Predicted species}} \\
\textbf{Reference} & Pine & Spruce & Birch & Maple & Aspen & Rowan & Oak & Linden & Alder & Count \\
\midrule
Pine & \textbf{0.94} & 0.02 & 0.02 & 0.00 & 0.01 & 0.01 & 0.00 & 0.00 & 0.00 & 2047 \\
Spruce & 0.06 & \textbf{0.87} & 0.05 & 0.02 & 0.00 & 0.00 & 0.00 & 0.00 & 0.00 & 528 \\
Birch & 0.02 & 0.01 & \textbf{0.93} & 0.00 & 0.02 & 0.00 & 0.00 & 0.00 & 0.00 & 1258 \\
Maple & 0.01 & 0.01 & 0.04 & \textbf{0.89} & 0.02 & 0.02 & 0.01 & 0.01 & 0.00 & 587 \\
Aspen & 0.06 & 0.01 & 0.24 & 0.03 & \textbf{0.59} & 0.02 & 0.01 & 0.04 & 0.00 & 376 \\
Rowan & 0.04 & 0.01 & 0.05 & 0.16 & 0.06 & \textbf{0.63} & 0.02 & 0.03 & 0.01 & 171 \\
Oak & 0.04 & 0.06 & 0.08 & 0.35 & 0.08 & 0.08 & \textbf{0.20} & 0.10 & 0.00 & 49 \\
Linden & 0.04 & 0.00 & 0.02 & 0.24 & 0.05 & 0.02 & 0.00 & \textbf{0.64} & 0.00 & 129 \\
Alder & 0.03 & 0.02 & 0.10 & 0.33 & 0.33 & 0.04 & 0.02 & 0.05 & \textbf{0.08} & 116 \\
Count & 2037 & 519 & 1386 & 669 & 331 & 161 & 25 & 121 & 12 & 5261 \\
\bottomrule
\end{tabular}
}
\end{table}

\begin{table}[htbp]
\centering
\caption{Confusion matrix: FGI-PointTransformer-DL-3D, HeliALS}
\label{tab:confusion_matrix_FGI-PointTransformer-DL-3D,_HeliALS}
\footnotesize
\resizebox{\columnwidth}{!}{%
\begin{tabular}{lcccccccccc}
\toprule
 & \multicolumn{10}{c}{\textbf{Predicted species}} \\
\textbf{Reference} & Pine & Spruce & Birch & Maple & Aspen & Rowan & Oak & Linden & Alder & Count \\
\midrule
Pine & \textbf{0.94} & 0.02 & 0.02 & 0.00 & 0.01 & 0.01 & 0.00 & 0.00 & 0.00 & 2047 \\
Spruce & 0.05 & \textbf{0.88} & 0.04 & 0.01 & 0.01 & 0.01 & 0.00 & 0.01 & 0.00 & 528 \\
Birch & 0.03 & 0.01 & \textbf{0.92} & 0.01 & 0.02 & 0.01 & 0.00 & 0.00 & 0.00 & 1258 \\
Maple & 0.01 & 0.01 & 0.03 & \textbf{0.89} & 0.02 & 0.03 & 0.01 & 0.00 & 0.00 & 587 \\
Aspen & 0.02 & 0.01 & 0.15 & 0.01 & \textbf{0.78} & 0.02 & 0.01 & 0.00 & 0.01 & 376 \\
Rowan & 0.01 & 0.01 & 0.02 & 0.10 & 0.05 & \textbf{0.76} & 0.02 & 0.02 & 0.01 & 171 \\
Oak & 0.04 & 0.04 & 0.08 & 0.14 & 0.06 & 0.12 & \textbf{0.39} & 0.12 & 0.00 & 49 \\
Linden & 0.01 & 0.01 & 0.01 & 0.19 & 0.03 & 0.03 & 0.01 & \textbf{0.71} & 0.02 & 129 \\
Alder & 0.03 & 0.01 & 0.17 & 0.11 & 0.47 & 0.02 & 0.03 & 0.03 & \textbf{0.14} & 116 \\
Count & 2014 & 524 & 1324 & 600 & 428 & 192 & 39 & 113 & 27 & 5261 \\
\bottomrule
\end{tabular}
}
\end{table}

\begin{table}[htbp]
\centering
\caption{Confusion matrix: SA-Detailview-DL-2D, HeliALS}
\label{tab:confusion_matrix_NIBIO-DetailView-DL-2D*,_HeliALS}
\footnotesize
\resizebox{\columnwidth}{!}{%
\begin{tabular}{lccccccccccc}
\toprule
 & \multicolumn{11}{c}{\textbf{Predicted species}} \\
\textbf{Reference} & Pine & Spruce & Birch & Maple & Aspen & Rowan & Oak & Linden & Alder & Missing & Count \\
\midrule
Pine & \textbf{0.50} & 0.13 & 0.23 & 0.05 & 0.00 & 0.00 & 0.08 & 0.00 & 0.00 & 0.00 & 2047 \\
Spruce & 0.05 & \textbf{0.87} & 0.07 & 0.00 & 0.00 & 0.00 & 0.01 & 0.00 & 0.00 & 0.01 & 528 \\
Birch & 0.01 & 0.11 & \textbf{0.83} & 0.02 & 0.00 & 0.00 & 0.02 & 0.00 & 0.00 & 0.00 & 1258 \\
Maple & 0.01 & 0.07 & 0.37 & \textbf{0.28} & 0.00 & 0.00 & 0.26 & 0.00 & 0.00 & 0.00 & 587 \\
Aspen & 0.03 & 0.08 & 0.73 & 0.03 & \textbf{0.00} & 0.00 & 0.11 & 0.00 & 0.00 & 0.01 & 376 \\
Rowan & 0.01 & 0.24 & 0.54 & 0.15 & 0.00 & \textbf{0.00} & 0.05 & 0.00 & 0.00 & 0.00 & 171 \\
Oak & 0.04 & 0.04 & 0.65 & 0.02 & 0.00 & 0.00 & \textbf{0.24} & 0.00 & 0.00 & 0.00 & 49 \\
Linden & 0.26 & 0.20 & 0.19 & 0.10 & 0.00 & 0.00 & 0.24 & \textbf{0.00} & 0.00 & 0.01 & 129 \\
Alder & 0.03 & 0.03 & 0.46 & 0.20 & 0.00 & 0.00 & 0.28 & 0.00 & \textbf{0.00} & 0.01 & 116 \\
Count & 1122 & 1008 & 2251 & 375 & 2 & 0 & 484 & 0 & 0 & 19 & 5261 \\
\bottomrule
\end{tabular}
}
\end{table}

\begin{table}[htbp]
\centering
\caption{Confusion matrix: SLU-YOLOv8-DL-2D, HeliALS}
\label{tab:confusion_matrix_SLU-YOLOv8-DL-2D,_HeliALS}
\footnotesize
\resizebox{\columnwidth}{!}{%
\begin{tabular}{lcccccccccc}
\toprule
 & \multicolumn{10}{c}{\textbf{Predicted species}} \\
\textbf{Reference} & Pine & Spruce & Birch & Maple & Aspen & Rowan & Oak & Linden & Alder & Count \\
\midrule
Pine & \textbf{0.90} & 0.01 & 0.03 & 0.01 & 0.02 & 0.00 & 0.01 & 0.01 & 0.00 & 2047 \\
Spruce & 0.05 & \textbf{0.89} & 0.03 & 0.00 & 0.02 & 0.00 & 0.00 & 0.01 & 0.00 & 528 \\
Birch & 0.02 & 0.01 & \textbf{0.93} & 0.01 & 0.03 & 0.00 & 0.01 & 0.00 & 0.00 & 1258 \\
Maple & 0.03 & 0.01 & 0.03 & \textbf{0.83} & 0.02 & 0.04 & 0.03 & 0.01 & 0.00 & 587 \\
Aspen & 0.06 & 0.03 & 0.23 & 0.02 & \textbf{0.59} & 0.01 & 0.06 & 0.01 & 0.00 & 376 \\
Rowan & 0.07 & 0.00 & 0.04 & 0.20 & 0.03 & \textbf{0.60} & 0.02 & 0.03 & 0.00 & 171 \\
Oak & 0.20 & 0.02 & 0.06 & 0.24 & 0.04 & 0.06 & \textbf{0.31} & 0.06 & 0.00 & 49 \\
Linden & 0.04 & 0.01 & 0.02 & 0.30 & 0.07 & 0.05 & 0.02 & \textbf{0.48} & 0.00 & 129 \\
Alder & 0.35 & 0.03 & 0.08 & 0.23 & 0.11 & 0.06 & 0.03 & 0.02 & \textbf{0.09} & 116 \\
Count & 2008 & 527 & 1383 & 643 & 340 & 161 & 91 & 96 & 12 & 5261 \\
\bottomrule
\end{tabular}
}
\end{table}

\begin{table}[htbp]
\centering
\caption{Confusion matrix: TUW-PointNet++-DL-3D, HeliALS}
\label{tab:confusion_matrix_TUW-PointNet++-DL-3D,_HeliALS}
\footnotesize
\resizebox{\columnwidth}{!}{%
\begin{tabular}{lcccccccccc}
\toprule
 & \multicolumn{10}{c}{\textbf{Predicted species}} \\
\textbf{Reference} & Pine & Spruce & Birch & Maple & Aspen & Rowan & Oak & Linden & Alder & Count \\
\midrule
Pine & \textbf{0.85} & 0.03 & 0.05 & 0.02 & 0.03 & 0.01 & 0.00 & 0.01 & 0.00 & 2047 \\
Spruce & 0.04 & \textbf{0.76} & 0.10 & 0.02 & 0.07 & 0.00 & 0.00 & 0.01 & 0.00 & 528 \\
Birch & 0.04 & 0.02 & \textbf{0.87} & 0.00 & 0.05 & 0.00 & 0.01 & 0.00 & 0.00 & 1258 \\
Maple & 0.02 & 0.00 & 0.02 & \textbf{0.76} & 0.02 & 0.06 & 0.07 & 0.03 & 0.01 & 587 \\
Aspen & 0.06 & 0.03 & 0.24 & 0.01 & \textbf{0.57} & 0.01 & 0.03 & 0.05 & 0.01 & 376 \\
Rowan & 0.06 & 0.00 & 0.03 & 0.41 & 0.03 & \textbf{0.38} & 0.02 & 0.06 & 0.01 & 171 \\
Oak & 0.06 & 0.02 & 0.10 & 0.20 & 0.18 & 0.12 & \textbf{0.18} & 0.12 & 0.00 & 49 \\
Linden & 0.04 & 0.00 & 0.01 & 0.16 & 0.03 & 0.12 & 0.09 & \textbf{0.53} & 0.02 & 129 \\
Alder & 0.04 & 0.03 & 0.09 & 0.11 & 0.48 & 0.01 & 0.09 & 0.07 & \textbf{0.09} & 116 \\
Count & 1862 & 499 & 1382 & 612 & 474 & 149 & 112 & 147 & 24 & 5261 \\
\bottomrule
\end{tabular}
}
\end{table}

\begin{table}[htbp]
\centering
\caption{Confusion matrix: FBK-PointNet++-DL-3D, Optech Titan}
\label{tab:confusion_matrix_FBK-PointNet++-DL-3D,_Optech_Titan}
\footnotesize
\resizebox{\columnwidth}{!}{%
\begin{tabular}{lcccccccccc}
\toprule
 & \multicolumn{10}{c}{\textbf{Predicted species}} \\
\textbf{Reference} & Pine & Spruce & Birch & Maple & Aspen & Rowan & Oak & Linden & Alder & Count \\
\midrule
Pine & \textbf{0.87} & 0.03 & 0.09 & 0.00 & 0.01 & 0.00 & 0.00 & 0.00 & 0.00 & 2047 \\
Spruce & 0.06 & \textbf{0.81} & 0.10 & 0.01 & 0.01 & 0.00 & 0.00 & 0.00 & 0.01 & 528 \\
Birch & 0.07 & 0.04 & \textbf{0.84} & 0.00 & 0.03 & 0.00 & 0.00 & 0.00 & 0.00 & 1258 \\
Maple & 0.04 & 0.02 & 0.05 & \textbf{0.77} & 0.03 & 0.06 & 0.02 & 0.01 & 0.02 & 587 \\
Aspen & 0.10 & 0.14 & 0.28 & 0.02 & \textbf{0.38} & 0.03 & 0.02 & 0.01 & 0.04 & 376 \\
Rowan & 0.13 & 0.01 & 0.07 & 0.27 & 0.13 & \textbf{0.22} & 0.03 & 0.12 & 0.02 & 171 \\
Oak & 0.04 & 0.10 & 0.12 & 0.27 & 0.24 & 0.04 & \textbf{0.14} & 0.02 & 0.02 & 49 \\
Linden & 0.12 & 0.06 & 0.09 & 0.15 & 0.17 & 0.15 & 0.00 & \textbf{0.26} & 0.02 & 129 \\
Alder & 0.05 & 0.08 & 0.28 & 0.07 & 0.33 & 0.07 & 0.03 & 0.03 & \textbf{0.06} & 116 \\
Count & 1999 & 626 & 1487 & 563 & 311 & 119 & 37 & 74 & 45 & 5261 \\
\bottomrule
\end{tabular}
}
\end{table}

\begin{table}[htbp]
\centering
\caption{Confusion matrix: FGI-DGCNN-DL-3D, Optech Titan}
\label{tab:confusion_matrix_FGI-DGCNN-DL-3D,_Optech_Titan}
\footnotesize
\resizebox{\columnwidth}{!}{%
\begin{tabular}{lcccccccccc}
\toprule
 & \multicolumn{10}{c}{\textbf{Predicted species}} \\
\textbf{Reference} & Pine & Spruce & Birch & Maple & Aspen & Rowan & Oak & Linden & Alder & Count \\
\midrule
Pine & \textbf{0.93} & 0.02 & 0.03 & 0.00 & 0.01 & 0.00 & 0.00 & 0.01 & 0.00 & 2047 \\
Spruce & 0.11 & \textbf{0.79} & 0.06 & 0.01 & 0.01 & 0.00 & 0.00 & 0.02 & 0.00 & 528 \\
Birch & 0.12 & 0.03 & \textbf{0.80} & 0.01 & 0.04 & 0.00 & 0.00 & 0.00 & 0.00 & 1258 \\
Maple & 0.09 & 0.02 & 0.03 & \textbf{0.66} & 0.02 & 0.02 & 0.02 & 0.14 & 0.00 & 587 \\
Aspen & 0.16 & 0.08 & 0.25 & 0.01 & \textbf{0.44} & 0.00 & 0.02 & 0.03 & 0.00 & 376 \\
Rowan & 0.29 & 0.02 & 0.03 & 0.16 & 0.05 & \textbf{0.09} & 0.06 & 0.30 & 0.00 & 171 \\
Oak & 0.24 & 0.04 & 0.04 & 0.16 & 0.20 & 0.00 & \textbf{0.12} & 0.18 & 0.00 & 49 \\
Linden & 0.15 & 0.03 & 0.01 & 0.12 & 0.02 & 0.02 & 0.01 & \textbf{0.64} & 0.00 & 129 \\
Alder & 0.07 & 0.05 & 0.07 & 0.15 & 0.57 & 0.02 & 0.03 & 0.04 & \textbf{0.00} & 116 \\
Count & 2320 & 537 & 1240 & 478 & 335 & 33 & 43 & 275 & 0 & 5261 \\
\bottomrule
\end{tabular}
}
\end{table}

\begin{table}[htbp]
\centering
\caption{Confusion matrix: FGI-PointNet-DL-3D, Optech Titan}
\label{tab:confusion_matrix_FGI-PointNet-DL-3D,_Optech_Titan}
\footnotesize
\resizebox{\columnwidth}{!}{%
\begin{tabular}{lcccccccccc}
\toprule
 & \multicolumn{10}{c}{\textbf{Predicted species}} \\
\textbf{Reference} & Pine & Spruce & Birch & Maple & Aspen & Rowan & Oak & Linden & Alder & Count \\
\midrule
Pine & \textbf{0.90} & 0.02 & 0.05 & 0.01 & 0.01 & 0.01 & 0.00 & 0.00 & 0.00 & 2047 \\
Spruce & 0.05 & \textbf{0.84} & 0.07 & 0.01 & 0.01 & 0.01 & 0.00 & 0.01 & 0.00 & 528 \\
Birch & 0.06 & 0.04 & \textbf{0.84} & 0.01 & 0.04 & 0.00 & 0.00 & 0.00 & 0.00 & 1258 \\
Maple & 0.02 & 0.02 & 0.04 & \textbf{0.82} & 0.02 & 0.07 & 0.00 & 0.01 & 0.00 & 587 \\
Aspen & 0.06 & 0.09 & 0.28 & 0.02 & \textbf{0.49} & 0.02 & 0.02 & 0.02 & 0.02 & 376 \\
Rowan & 0.08 & 0.01 & 0.06 & 0.29 & 0.10 & \textbf{0.38} & 0.01 & 0.08 & 0.00 & 171 \\
Oak & 0.02 & 0.04 & 0.12 & 0.18 & 0.20 & 0.12 & \textbf{0.20} & 0.10 & 0.00 & 49 \\
Linden & 0.05 & 0.02 & 0.02 & 0.19 & 0.12 & 0.12 & 0.05 & \textbf{0.43} & 0.00 & 129 \\
Alder & 0.02 & 0.03 & 0.09 & 0.13 & 0.58 & 0.03 & 0.03 & 0.03 & \textbf{0.08} & 116 \\
Count & 1999 & 592 & 1360 & 616 & 384 & 153 & 35 & 103 & 19 & 5261 \\
\bottomrule
\end{tabular}
}
\end{table}

\begin{table}[htbp]
\centering
\caption{Confusion matrix: FGI-PointTransformerWeighted-
DL-3D, Optech Titan}
\label{tab:confusion_matrix_FGI-PointTransformerWeighted-
DL-3D,_Optech_Titan}
\footnotesize
\resizebox{\columnwidth}{!}{%
\begin{tabular}{lcccccccccc}
\toprule
 & \multicolumn{10}{c}{\textbf{Predicted species}} \\
\textbf{Reference} & Pine & Spruce & Birch & Maple & Aspen & Rowan & Oak & Linden & Alder & Count \\
\midrule
Pine & \textbf{0.88} & 0.03 & 0.06 & 0.00 & 0.02 & 0.00 & 0.00 & 0.00 & 0.00 & 2047 \\
Spruce & 0.04 & \textbf{0.80} & 0.07 & 0.01 & 0.04 & 0.00 & 0.01 & 0.02 & 0.00 & 528 \\
Birch & 0.06 & 0.05 & \textbf{0.82} & 0.01 & 0.06 & 0.00 & 0.00 & 0.00 & 0.00 & 1258 \\
Maple & 0.02 & 0.01 & 0.03 & \textbf{0.82} & 0.05 & 0.04 & 0.02 & 0.01 & 0.00 & 587 \\
Aspen & 0.09 & 0.08 & 0.18 & 0.02 & \textbf{0.58} & 0.02 & 0.02 & 0.01 & 0.01 & 376 \\
Rowan & 0.09 & 0.01 & 0.04 & 0.31 & 0.19 & \textbf{0.28} & 0.01 & 0.05 & 0.02 & 171 \\
Oak & 0.06 & 0.02 & 0.06 & 0.18 & 0.31 & 0.04 & \textbf{0.24} & 0.08 & 0.00 & 49 \\
Linden & 0.09 & 0.03 & 0.01 & 0.13 & 0.12 & 0.02 & 0.02 & \textbf{0.57} & 0.01 & 129 \\
Alder & 0.00 & 0.03 & 0.05 & 0.14 & 0.57 & 0.01 & 0.09 & 0.01 & \textbf{0.10} & 116 \\
Count & 1971 & 599 & 1297 & 607 & 508 & 88 & 53 & 114 & 24 & 5261 \\
\bottomrule
\end{tabular}
}
\end{table}

\begin{table}[htbp]
\centering
\caption{Confusion matrix: NTNU-ConvNeXt-T-DL-2D, Optech Titan}
\label{tab:confusion_matrix_NTNU-ConvNeXt-T-DL-2D,_Optech_Titan}
\footnotesize
\resizebox{\columnwidth}{!}{%
\begin{tabular}{lcccccccccc}
\toprule
 & \multicolumn{10}{c}{\textbf{Predicted species}} \\
\textbf{Reference} & Pine & Spruce & Birch & Maple & Aspen & Rowan & Oak & Linden & Alder & Count \\
\midrule
Pine & \textbf{0.84} & 0.02 & 0.09 & 0.02 & 0.02 & 0.00 & 0.00 & 0.01 & 0.00 & 2047 \\
Spruce & 0.05 & \textbf{0.73} & 0.15 & 0.01 & 0.03 & 0.00 & 0.00 & 0.02 & 0.00 & 528 \\
Birch & 0.06 & 0.04 & \textbf{0.81} & 0.01 & 0.05 & 0.00 & 0.01 & 0.01 & 0.00 & 1258 \\
Maple & 0.01 & 0.00 & 0.02 & \textbf{0.91} & 0.03 & 0.02 & 0.01 & 0.01 & 0.00 & 587 \\
Aspen & 0.06 & 0.06 & 0.18 & 0.05 & \textbf{0.48} & 0.01 & 0.07 & 0.06 & 0.02 & 376 \\
Rowan & 0.03 & 0.01 & 0.02 & 0.43 & 0.08 & \textbf{0.36} & 0.01 & 0.06 & 0.00 & 171 \\
Oak & 0.08 & 0.02 & 0.08 & 0.24 & 0.20 & 0.02 & \textbf{0.20} & 0.14 & 0.00 & 49 \\
Linden & 0.04 & 0.02 & 0.01 & 0.22 & 0.05 & 0.05 & 0.01 & \textbf{0.60} & 0.00 & 129 \\
Alder & 0.02 & 0.02 & 0.08 & 0.21 & 0.42 & 0.02 & 0.10 & 0.09 & \textbf{0.05} & 116 \\
Count & 1855 & 505 & 1384 & 743 & 411 & 95 & 70 & 179 & 19 & 5261 \\
\bottomrule
\end{tabular}
}
\end{table}

\begin{table}[htbp]
\centering
\caption{Confusion matrix: Aalto-RF-ML, Optech Titan}
\label{tab:confusion_matrix_Aalto-RF-ML,_Optech_Titan}
\footnotesize
\resizebox{\columnwidth}{!}{%
\begin{tabular}{lccccccccccc}
\toprule
 & \multicolumn{11}{c}{\textbf{Predicted species}} \\
\textbf{Reference} & Pine & Spruce & Birch & Maple & Aspen & Rowan & Oak & Linden & Alder & Missing & Count \\
\midrule
Pine & \textbf{0.83} & 0.02 & 0.13 & 0.00 & 0.00 & 0.00 & 0.00 & 0.01 & 0.00 & 0.00 & 2047 \\
Spruce & 0.09 & \textbf{0.67} & 0.22 & 0.01 & 0.00 & 0.01 & 0.00 & 0.00 & 0.00 & 0.00 & 528 \\
Birch & 0.08 & 0.02 & \textbf{0.88} & 0.00 & 0.00 & 0.00 & 0.00 & 0.00 & 0.01 & 0.00 & 1258 \\
Maple & 0.02 & 0.00 & 0.04 & \textbf{0.84} & 0.00 & 0.06 & 0.03 & 0.00 & 0.01 & 0.00 & 587 \\
Aspen & 0.13 & 0.08 & 0.48 & 0.01 & \textbf{0.00} & 0.03 & 0.10 & 0.07 & 0.12 & 0.00 & 376 \\
Rowan & 0.08 & 0.02 & 0.05 & 0.32 & 0.00 & \textbf{0.40} & 0.07 & 0.05 & 0.02 & 0.00 & 171 \\
Oak & 0.02 & 0.08 & 0.16 & 0.08 & 0.00 & 0.29 & \textbf{0.16} & 0.08 & 0.12 & 0.00 & 49 \\
Linden & 0.14 & 0.05 & 0.18 & 0.09 & 0.00 & 0.11 & 0.09 & \textbf{0.33} & 0.01 & 0.00 & 129 \\
Alder & 0.06 & 0.03 & 0.10 & 0.17 & 0.00 & 0.16 & 0.26 & 0.09 & \textbf{0.14} & 0.00 & 116 \\
Count & 1957 & 467 & 1744 & 594 & 0 & 177 & 121 & 113 & 87 & 1 & 5261 \\
\bottomrule
\end{tabular}
}
\end{table}

\begin{table}[htbp]
\centering
\caption{Confusion matrix: FGI-RF-ML, Optech Titan}
\label{tab:confusion_matrix_FGI-RF-ML,_Optech_Titan}
\footnotesize
\resizebox{\columnwidth}{!}{%
\begin{tabular}{lccccccccccc}
\toprule
 & \multicolumn{11}{c}{\textbf{Predicted species}} \\
\textbf{Reference} & Pine & Spruce & Birch & Maple & Aspen & Rowan & Oak & Linden & Alder & Missing & Count \\
\midrule
Pine & \textbf{0.86} & 0.02 & 0.11 & 0.00 & 0.01 & 0.00 & 0.00 & 0.00 & 0.00 & 0.00 & 2047 \\
Spruce & 0.05 & \textbf{0.74} & 0.16 & 0.01 & 0.02 & 0.00 & 0.00 & 0.01 & 0.00 & 0.00 & 528 \\
Birch & 0.09 & 0.05 & \textbf{0.79} & 0.00 & 0.04 & 0.00 & 0.00 & 0.00 & 0.00 & 0.02 & 1258 \\
Maple & 0.01 & 0.01 & 0.02 & \textbf{0.85} & 0.03 & 0.03 & 0.03 & 0.00 & 0.00 & 0.01 & 587 \\
Aspen & 0.10 & 0.09 & 0.15 & 0.01 & \textbf{0.61} & 0.01 & 0.01 & 0.01 & 0.00 & 0.00 & 376 \\
Rowan & 0.10 & 0.01 & 0.02 & 0.25 & 0.11 & \textbf{0.44} & 0.02 & 0.04 & 0.00 & 0.02 & 171 \\
Oak & 0.08 & 0.08 & 0.04 & 0.12 & 0.37 & 0.16 & \textbf{0.12} & 0.02 & 0.00 & 0.00 & 49 \\
Linden & 0.10 & 0.02 & 0.00 & 0.15 & 0.13 & 0.07 & 0.03 & \textbf{0.48} & 0.00 & 0.02 & 129 \\
Alder & 0.03 & 0.05 & 0.08 & 0.09 & 0.53 & 0.03 & 0.14 & 0.04 & \textbf{0.00} & 0.00 & 116 \\
Count & 1976 & 544 & 1382 & 593 & 446 & 129 & 61 & 89 & 0 & 41 & 5261 \\
\bottomrule
\end{tabular}
}
\end{table}

\begin{table}[htbp]
\centering
\caption{Confusion matrix: LUKE-MultiRF-ML, Optech Titan}
\label{tab:confusion_matrix_LUKE-MultiRF-ML,_Optech_Titan}
\footnotesize
\resizebox{\columnwidth}{!}{%
\begin{tabular}{lcccccccccc}
\toprule
 & \multicolumn{10}{c}{\textbf{Predicted species}} \\
\textbf{Reference} & Pine & Spruce & Birch & Maple & Aspen & Rowan & Oak & Linden & Alder & Count \\
\midrule
Pine & \textbf{0.83} & 0.06 & 0.06 & 0.01 & 0.01 & 0.00 & 0.02 & 0.00 & 0.00 & 2047 \\
Spruce & 0.09 & \textbf{0.72} & 0.14 & 0.01 & 0.02 & 0.00 & 0.01 & 0.01 & 0.00 & 528 \\
Birch & 0.06 & 0.04 & \textbf{0.82} & 0.00 & 0.06 & 0.00 & 0.01 & 0.00 & 0.01 & 1258 \\
Maple & 0.04 & 0.02 & 0.01 & \textbf{0.85} & 0.03 & 0.02 & 0.02 & 0.01 & 0.01 & 587 \\
Aspen & 0.10 & 0.06 & 0.23 & 0.01 & \textbf{0.51} & 0.01 & 0.02 & 0.02 & 0.05 & 376 \\
Rowan & 0.09 & 0.02 & 0.03 & 0.51 & 0.13 & \textbf{0.17} & 0.01 & 0.05 & 0.00 & 171 \\
Oak & 0.10 & 0.18 & 0.10 & 0.24 & 0.16 & 0.00 & \textbf{0.12} & 0.04 & 0.04 & 49 \\
Linden & 0.12 & 0.01 & 0.02 & 0.33 & 0.13 & 0.08 & 0.01 & \textbf{0.32} & 0.00 & 129 \\
Alder & 0.17 & 0.03 & 0.08 & 0.12 & 0.40 & 0.02 & 0.09 & 0.02 & \textbf{0.08} & 116 \\
Count & 1935 & 587 & 1346 & 691 & 411 & 62 & 87 & 81 & 61 & 5261 \\
\bottomrule
\end{tabular}
}
\end{table}

\begin{table}[htbp]
\centering
\caption{Confusion matrix: IDEAS-RF-ML, Optech Titan}
\label{tab:confusion_matrix_NCBR-RF-ML,_Optech_Titan}
\footnotesize
\resizebox{\columnwidth}{!}{%
\begin{tabular}{lcccccccccc}
\toprule
 & \multicolumn{10}{c}{\textbf{Predicted species}} \\
\textbf{Reference} & Pine & Spruce & Birch & Maple & Aspen & Rowan & Oak & Linden & Alder & Count \\
\midrule
Pine & \textbf{0.89} & 0.01 & 0.08 & 0.00 & 0.02 & 0.00 & 0.00 & 0.00 & 0.00 & 2047 \\
Spruce & 0.11 & \textbf{0.64} & 0.23 & 0.01 & 0.02 & 0.01 & 0.00 & 0.00 & 0.00 & 528 \\
Birch & 0.08 & 0.02 & \textbf{0.84} & 0.01 & 0.04 & 0.00 & 0.00 & 0.00 & 0.00 & 1258 \\
Maple & 0.03 & 0.00 & 0.03 & \textbf{0.87} & 0.03 & 0.03 & 0.01 & 0.00 & 0.00 & 587 \\
Aspen & 0.13 & 0.05 & 0.26 & 0.01 & \textbf{0.53} & 0.01 & 0.01 & 0.01 & 0.00 & 376 \\
Rowan & 0.19 & 0.01 & 0.02 & 0.24 & 0.09 & \textbf{0.40} & 0.01 & 0.05 & 0.00 & 171 \\
Oak & 0.16 & 0.06 & 0.08 & 0.16 & 0.29 & 0.04 & \textbf{0.08} & 0.12 & 0.00 & 49 \\
Linden & 0.16 & 0.02 & 0.00 & 0.16 & 0.19 & 0.07 & 0.01 & \textbf{0.40} & 0.00 & 129 \\
Alder & 0.06 & 0.00 & 0.16 & 0.14 & 0.52 & 0.04 & 0.05 & 0.03 & \textbf{0.00} & 116 \\
Count & 2108 & 416 & 1471 & 618 & 429 & 116 & 25 & 78 & 0 & 5261 \\
\bottomrule
\end{tabular}
}
\end{table}

\begin{table}[htbp]
\centering
\caption{Confusion matrix: UEF-LGBM-ML, Optech Titan}
\label{tab:confusion_matrix_UEF-LGBM-ML,_Optech_Titan}
\footnotesize
\resizebox{\columnwidth}{!}{%
\begin{tabular}{lcccccccccc}
\toprule
 & \multicolumn{10}{c}{\textbf{Predicted species}} \\
\textbf{Reference} & Pine & Spruce & Birch & Maple & Aspen & Rowan & Oak & Linden & Alder & Count \\
\midrule
Pine & \textbf{0.86} & 0.03 & 0.09 & 0.00 & 0.01 & 0.00 & 0.00 & 0.01 & 0.00 & 2047 \\
Spruce & 0.05 & \textbf{0.81} & 0.10 & 0.01 & 0.01 & 0.00 & 0.00 & 0.01 & 0.00 & 528 \\
Birch & 0.09 & 0.04 & \textbf{0.82} & 0.00 & 0.04 & 0.00 & 0.00 & 0.00 & 0.00 & 1258 \\
Maple & 0.03 & 0.02 & 0.02 & \textbf{0.87} & 0.02 & 0.03 & 0.02 & 0.01 & 0.00 & 587 \\
Aspen & 0.14 & 0.10 & 0.21 & 0.01 & \textbf{0.49} & 0.02 & 0.01 & 0.01 & 0.02 & 376 \\
Rowan & 0.11 & 0.01 & 0.04 & 0.29 & 0.09 & \textbf{0.37} & 0.04 & 0.06 & 0.00 & 171 \\
Oak & 0.20 & 0.00 & 0.10 & 0.24 & 0.22 & 0.08 & \textbf{0.10} & 0.04 & 0.00 & 49 \\
Linden & 0.09 & 0.02 & 0.02 & 0.19 & 0.07 & 0.02 & 0.02 & \textbf{0.56} & 0.00 & 129 \\
Alder & 0.13 & 0.08 & 0.09 & 0.14 & 0.40 & 0.05 & 0.10 & 0.01 & \textbf{0.01} & 116 \\
Count & 2014 & 595 & 1389 & 629 & 355 & 105 & 46 & 113 & 15 & 5261 \\
\bottomrule
\end{tabular}
}
\end{table}

\begin{table}[htbp]
\centering
\caption{Confusion matrix: UEF-RF-ML, Optech Titan}
\label{tab:confusion_matrix_UEF-RF-ML,_Optech_Titan}
\footnotesize
\resizebox{\columnwidth}{!}{%
\begin{tabular}{lcccccccccc}
\toprule
 & \multicolumn{10}{c}{\textbf{Predicted species}} \\
\textbf{Reference} & Pine & Spruce & Birch & Maple & Aspen & Rowan & Oak & Linden & Alder & Count \\
\midrule
Pine & \textbf{0.89} & 0.01 & 0.09 & 0.00 & 0.01 & 0.00 & 0.00 & 0.00 & 0.00 & 2047 \\
Spruce & 0.06 & \textbf{0.81} & 0.09 & 0.01 & 0.02 & 0.00 & 0.00 & 0.01 & 0.00 & 528 \\
Birch & 0.11 & 0.03 & \textbf{0.82} & 0.00 & 0.03 & 0.00 & 0.00 & 0.00 & 0.00 & 1258 \\
Maple & 0.02 & 0.00 & 0.02 & \textbf{0.89} & 0.02 & 0.03 & 0.01 & 0.00 & 0.00 & 587 \\
Aspen & 0.20 & 0.07 & 0.14 & 0.00 & \textbf{0.57} & 0.01 & 0.02 & 0.01 & 0.00 & 376 \\
Rowan & 0.12 & 0.01 & 0.02 & 0.32 & 0.06 & \textbf{0.42} & 0.01 & 0.05 & 0.00 & 171 \\
Oak & 0.14 & 0.00 & 0.10 & 0.20 & 0.20 & 0.16 & \textbf{0.12} & 0.06 & 0.00 & 49 \\
Linden & 0.14 & 0.02 & 0.00 & 0.22 & 0.10 & 0.08 & 0.00 & \textbf{0.45} & 0.00 & 129 \\
Alder & 0.09 & 0.01 & 0.07 & 0.10 & 0.53 & 0.04 & 0.15 & 0.00 & \textbf{0.00} & 116 \\
Count & 2131 & 529 & 1334 & 640 & 385 & 119 & 40 & 83 & 0 & 5261 \\
\bottomrule
\end{tabular}
}
\end{table}

\begin{table}[htbp]
\centering
\caption{Confusion matrix: UEF-SVM-ML, Optech Titan}
\label{tab:confusion_matrix_UEF-SVM-ML,_Optech_Titan}
\footnotesize
\resizebox{\columnwidth}{!}{%
\begin{tabular}{lcccccccccc}
\toprule
 & \multicolumn{10}{c}{\textbf{Predicted species}} \\
\textbf{Reference} & Pine & Spruce & Birch & Maple & Aspen & Rowan & Oak & Linden & Alder & Count \\
\midrule
Pine & \textbf{0.84} & 0.02 & 0.11 & 0.00 & 0.02 & 0.00 & 0.01 & 0.01 & 0.00 & 2047 \\
Spruce & 0.06 & \textbf{0.71} & 0.18 & 0.00 & 0.03 & 0.00 & 0.00 & 0.01 & 0.00 & 528 \\
Birch & 0.09 & 0.04 & \textbf{0.82} & 0.00 & 0.04 & 0.00 & 0.00 & 0.00 & 0.00 & 1258 \\
Maple & 0.01 & 0.01 & 0.04 & \textbf{0.83} & 0.04 & 0.05 & 0.03 & 0.00 & 0.00 & 587 \\
Aspen & 0.12 & 0.07 & 0.23 & 0.01 & \textbf{0.55} & 0.02 & 0.01 & 0.00 & 0.00 & 376 \\
Rowan & 0.04 & 0.01 & 0.01 & 0.35 & 0.10 & \textbf{0.48} & 0.00 & 0.01 & 0.00 & 171 \\
Oak & 0.04 & 0.02 & 0.08 & 0.29 & 0.31 & 0.16 & \textbf{0.08} & 0.02 & 0.00 & 49 \\
Linden & 0.06 & 0.01 & 0.03 & 0.19 & 0.12 & 0.08 & 0.02 & \textbf{0.49} & 0.00 & 129 \\
Alder & 0.03 & 0.02 & 0.10 & 0.07 & 0.64 & 0.03 & 0.09 & 0.02 & \textbf{0.00} & 116 \\
Count & 1924 & 504 & 1478 & 607 & 460 & 147 & 55 & 86 & 0 & 5261 \\
\bottomrule
\end{tabular}
}
\end{table}

\begin{table}[htbp]
\centering
\caption{Confusion matrix: UPV-GB-ML, Optech Titan}
\label{tab:confusion_matrix_UPV-GB-ML,_Optech_Titan}
\footnotesize
\resizebox{\columnwidth}{!}{%
\begin{tabular}{lcccccccccc}
\toprule
 & \multicolumn{10}{c}{\textbf{Predicted species}} \\
\textbf{Reference} & Pine & Spruce & Birch & Maple & Aspen & Rowan & Oak & Linden & Alder & Count \\
\midrule
Pine & \textbf{0.84} & 0.04 & 0.09 & 0.01 & 0.02 & 0.00 & 0.00 & 0.00 & 0.00 & 2047 \\
Spruce & 0.05 & \textbf{0.70} & 0.17 & 0.01 & 0.04 & 0.00 & 0.01 & 0.01 & 0.01 & 528 \\
Birch & 0.09 & 0.07 & \textbf{0.78} & 0.01 & 0.04 & 0.00 & 0.00 & 0.00 & 0.00 & 1258 \\
Maple & 0.03 & 0.02 & 0.03 & \textbf{0.83} & 0.01 & 0.06 & 0.01 & 0.01 & 0.00 & 587 \\
Aspen & 0.10 & 0.14 & 0.23 & 0.00 & \textbf{0.45} & 0.02 & 0.03 & 0.01 & 0.02 & 376 \\
Rowan & 0.11 & 0.04 & 0.05 & 0.25 & 0.13 & \textbf{0.39} & 0.02 & 0.02 & 0.00 & 171 \\
Oak & 0.20 & 0.12 & 0.12 & 0.14 & 0.14 & 0.12 & \textbf{0.14} & 0.00 & 0.00 & 49 \\
Linden & 0.19 & 0.02 & 0.07 & 0.14 & 0.08 & 0.10 & 0.05 & \textbf{0.36} & 0.00 & 129 \\
Alder & 0.13 & 0.16 & 0.19 & 0.07 & 0.23 & 0.11 & 0.10 & 0.00 & \textbf{0.01} & 116 \\
Count & 1964 & 644 & 1401 & 597 & 360 & 145 & 66 & 66 & 18 & 5261 \\
\bottomrule
\end{tabular}
}
\end{table}

\end{appendices}

\end{document}